\documentclass[10pt]{article} 
\usepackage[preprint]{tmlr}

\usepackage{hyperref}
\usepackage{url}

\usepackage{bbm}
\usepackage{color}
\usepackage{algorithm}
\usepackage{algpseudocode}
\usepackage{bigfoot}  
\usepackage{graphicx}
\usepackage{subcaption}
\usepackage{amssymb}
\usepackage{amsmath}
\usepackage{titlesec}  
\usepackage{enumitem}
\usepackage{ulem}

\usepackage{booktabs}

\usepackage{tikz}
\usetikzlibrary{shapes,arrows,arrows.meta}
\usepackage{adjustbox}

\newcommand{\btheta}{\boldsymbol{\theta}}
\newcommand{\by}{\boldsymbol{y}}
\newcommand{\bphi}{\boldsymbol{\phi}}

\usepackage{bm}

\newcommand{\vect}[1]{\mathbf{#1}}

\newcommand{\br}[1]{\mathopen{}\left(#1\right)\mathclose{}}
\newcommand{\set}[1]{\left\{#1\right\}}

\newcommand{\prob}[1]{p\br{#1}}
\newcommand{\hatprob}[1]{\hat{p}\br{#1}}
\newcommand{\g}{\,|\,}

\title{Fast, accurate and lightweight sequential simulation-based inference using Gaussian locally linear mappings}

\author{\name Henrik H\"{a}ggstr\"{o}m \email henhagg@chalmers.se \\
      \addr Dept. Mathematical Sciences\\
      Chalmers and University of Gothenburg
      \AND
      \name Pedro L. C. Rodrigues \email pedro.rodrigues@inria.fr \\
      \addr Univ. Grenoble Alpes, Inria, CNRS, Grenoble INP, Inserm U1216, CHU Grenoble Alpes,\\ Grenoble Institut des
Neurosciences, LJK
      \AND
      \name Geoffroy Oudoumanessah \email Geoffroy.Oudoumanessah@inria.fr\\
      \addr Univ. Grenoble Alpes, Inria, CNRS, Grenoble INP, Inserm U1216, CHU Grenoble Alpes,\\ Grenoble Institut des
Neurosciences, LJK\\
Univ. Lyon, CNRS, Inserm, INSA Lyon, UCBL, CREATIS
\AND
\name Florence Forbes \email florence.forbes@inria.fr\\
\addr Univ. Grenoble Alpes, Inria, CNRS, Grenoble INP, Inserm U1216, CHU Grenoble Alpes,\\ Grenoble Institut des
Neurosciences, LJK
\AND
\name Umberto Picchini \email picchini@chalmers.se\\
\addr Dept. Mathematical Sciences,\\ Chalmers and University of Gothenburg
}

\def\month{MM}  
\def\year{YYYY} 
\def\openreview{\url{https://openreview.net/forum?id=XXXX}} 

\begin{document}

\maketitle

\begin{abstract}
  Bayesian inference for complex models with an intractable likelihood can be tackled using algorithms 
  performing many calls to computer simulators. These approaches are collectively known as \textit{simulation-based inference} (SBI). Recent SBI methods have made use of neural networks (NN) to provide approximate, yet expressive constructs for the unavailable likelihood function and the posterior distribution. \textcolor{black}{However, the trade-off between accuracy and computational demand leaves much space for improvement.} 
   In this work, we propose an alternative that provides both approximations to the likelihood and the posterior distribution, using  structured mixtures of probability distributions. 
   Our approach produces accurate posterior inference when compared to state-of-the-art NN-based SBI methods, even for multimodal posteriors, while exhibiting a much smaller computational footprint. We illustrate our results on several benchmark models from the SBI literature and on a biological model of the translation kinetics after mRNA transfection.
\end{abstract}

\section{Introduction}\label{sec:intro}

We propose a novel methodology achieving state-of-the-art Bayesian inference for models with an intractable likelihood function. We illustrate, via several benchmark models, how our methodology returns accurate inference, even in presence of multimodal posteriors, with a small runtime.
Moreover, for numerous simulation studies and applications, machine and statistical learning solutions  show an ever growing performance, but are not generally evaluated in terms of resource and energy consumption. Although their environmental impact is more and more documented \citep{Schwartz2020,Strubell2019,thompson2022computational}, 
the question of resource efficiency and energy consumption when learning model parameters is not as much studied as the quality of the inference itself, due to the multiple aspects to be taken into account and to the lack of clear metrics that limit the algorithms evaluation. When comparing with widely employed inference engines based on neural-networks, our method considerably lower the energy and memory required for the inference task, while showing competitive inference quality.
  
We consider simulation-based inference (SBI) for model parameters $\btheta$ in stochastic modelling.
  SBI refers to approaches that bypass the need to have a likelihood function $p(\by|\btheta)$ (for data $\by$) that is available in closed form, or one that is cheap to approximate, by instead considering a generative model $f(\btheta)$ or {\it computer simulator}. The simulator depends on $\btheta$ and possibly  on other inputs such as covariates and a stream of pseudorandom numbers. Generating a dataset $\by^*$ by running the simulator $\by^*=f(\btheta^*)$ at $\btheta=\btheta^*$ is then equivalent to simulating  $\by^*$ from the unavailable likelihood function. In the following this simulation is written as $\by^*\sim p(\by|\btheta^*)$.
Established SBI methods include approximate Bayesian computation (ABC, reviewed in \citealp{sisson2018handbook}), synthetic likelihoods \citep{wood2010statistical,price2018bayesian}, particle Markov chain Monte Carlo \citep{andrieu2009pseudo,andrieu2010particle}, and several more that we discuss in  Section \ref{sec:related-work}. 
More recently, methods exploiting neural networks (NN) have gained considerable attention. Neural-based estimators have been used to produce approximated likelihoods or approximated posterior distributions, typically using (discrete) normalizing flows \citep{rezende2015variational,papamakarios2021normalizing}.  NN have been used to approximate the likelihood function in implicit models (\citealp{papamakarios2019,chen2021neural}), the posterior distribution \citep{papamakarios2016,greenberg2019,durkan2020contrastive,chen2021neural} or the likelihood and the posterior distribution simultaneously \citep{wiqvist2021sequential,radev2023jana}.

\begin{figure}[h]
    \centering
    \includegraphics[width=10cm]{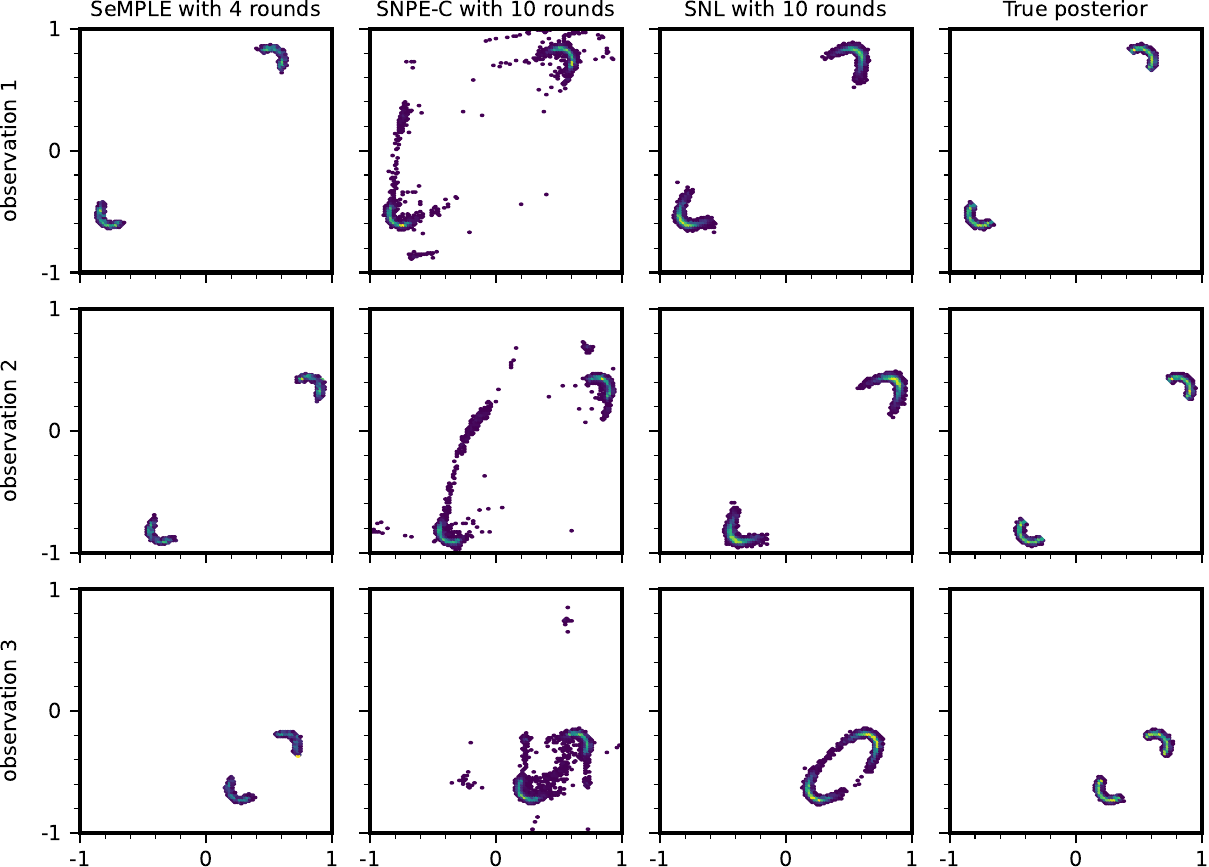}
    \caption{Two Moons model: inference using datasets \#1 (top), \#2 (middle) and \#3 (bottom) from \texttt{SBIBM}.  Posterior samples after 4 rounds of SeMPLE (column 1), 10 rounds of SNPE-C and SNL (columns 2 and 3). All methods used a total budget of 10K model simulations. The true posteriors are in column 4. }
\label{fig:mainpaper_two_moons_density_scatter_sbibmoptimal}
\end{figure}

\begin{figure}[h]
     \centering
\includegraphics[scale=0.35]{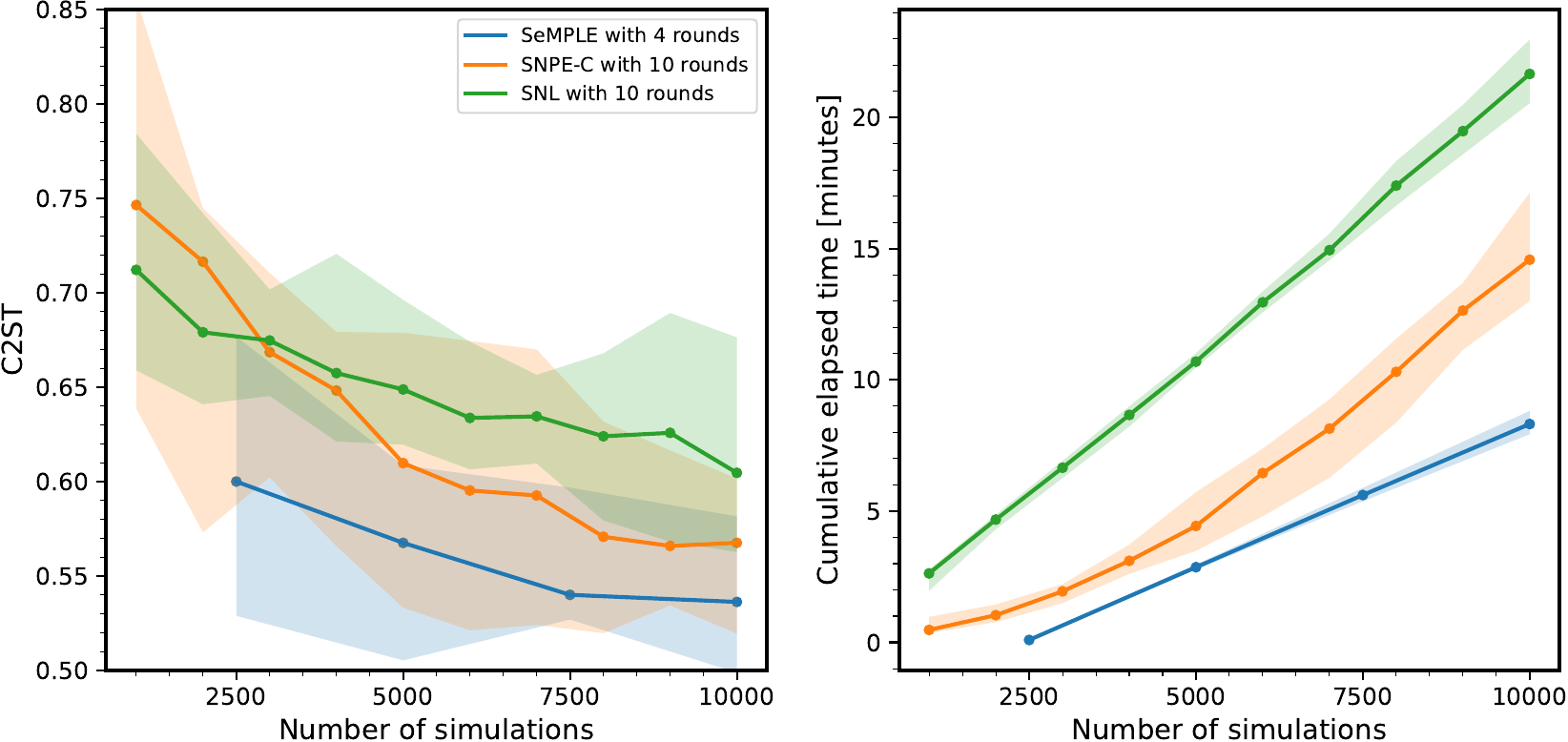}
     \caption{Two Moons. (left) Median C2ST (the lower the better) and median cumulative runtime in minutes (right) for 10 runs with different data sets vs the number of model simulations. Shaded bands enclose the min and max values. \textcolor{black}{C2ST is the two-samples classifier test, taking values in [0.5,1], see Section \ref{sec:examples}.}}
     \label{fig:TM_settings_metric_vs_sims}
\end{figure}

These neural-based posterior and likelihood estimation methods have produced exceptionally flexible inference strategies  for models with intractable likelihoods.
In this work, we propose to investigate more {\it frugal} strategies \citep{Evchenko2021} that can run with limited 
resources and a much smaller computational footprint, while returning inference of similar quality as neural-based approaches. 
For SBI, we propose to investigate structured mixtures of probability distributions whose versatility has been widely recognized for a variety of data and tasks, while not requiring excessive design effort or  tuning. Their expressivity makes them good candidates to account for complex multivariate models both for the likelihood and posterior distribution. Moreover, mixture models have a much smaller number of parameters $\bphi$ (see Table \ref{tab1}), compared to NN, and this makes them more amenable to interpretation and efficient learning. 
More specifically, we design a new simulation-based inference method named ``Sequential Mixture Posterior
and Likelihood Estimation'' (SeMPLE). SeMPLE learns approximations for both the posterior $p(\btheta|\by)$ and the likelihood $p(\by|\btheta)$ using a structured mixture model obtained via the  Gaussian Locally Linear Mapping (GLLiM, \citealp{deleforge}) method. Interestingly, both approximations are jointly returned, simultaneously, following a run of an Expectation-Maximization (EM) algorithm within GLLiM. We then use the GLLiM approximate posterior as a well-informed proposal distribution in a tuning-free Metropolis-Hastings sampler. The procedure is embedded in a sequential strategy, which we compare with state-of-the-art NN-based sequential algorithms for posterior inference, namely SNL \citep{papamakarios2019} and SNPE-C \citep{greenberg2019}. Other algorithms could be considered for comparison, however we stick with SNL and SNPE-C and a motivation for this is given in Section \ref{sec:related-work}. Our comparisons are in terms of inference quality as well as resource and energy requirements. We did not compare against ABC samplers such as SMC-ABC, except for the Lotka-Volterra example where such comparison was more suitable, as in \citealp{sbibm} it was shown that standard implementations of ABC approaches usually require a larger number of model simulations, compared to NN-based approaches, although recent research improved the SMC-ABC performance \citep{picchini2022guided}.  We found that, for a given computational budget specified by the total number of model simulations, our method is cheaper to run and produces  accurate inference thanks to (i) the informed proposal sampler created by SeMPLE and (ii) the fast model training. 
As an illustration, Figure \ref{fig:mainpaper_two_moons_density_scatter_sbibmoptimal} shows inference for three datasets of the {\it Two Moons} model, a model with bimodal posterior that we further analyse in Section \ref{sec:examples}. 
The runtime, energy \textcolor{black}{(consumed by the CPU and the DRAM)} and memory requirements for SeMPLE are considerably lower, see  Table \ref{tab1}, Figure \ref{fig:TM_settings_metric_vs_sims} (right) \textcolor{black}{and section \ref{sec:resource-requirements} for details on how energy and memory requirements were measured}. In other examples, ({\it e.g.} Figure \ref{fig:hyperb_settings_metric_vs_sims}), the speedup provided by SeMPLE is even more dramatic (for the model in Supplementary section \ref{sec:bernoulli-results_10rounds} we show that SeMPLE is 74 times faster then SNL and 26 times faster than SNPE-C).
Overall SeMPLE shows low resources costs that are always of the same magnitude  ($<10$ kJ and $<1$ GB), while the requirements for SNL and SNPE-C vary much more with the model dimension or the inference complexity. These results highlight the advantage brought by SeMPLE, which can be run on a minimal configuration while maintaining good performance.

\begin{table}[t]
  \centering
  \caption{Median resource requirements (across 10 independent repetitions) and number of parameters $\phi$ for posterior inference with a budget of 10K  (two moons and Bernoulli GLM) and 40K (hyperboloid and Ornstein-Uhlenbeck) model simulations. 
 Best (lowest)  values are in bold. SNPE-C and SNL used the default \texttt{SBIBM} setup \textcolor{black}{described in Appendix \ref{sec:sbibm}}.} 
    \begin{tabular}{c|cc|c}
    Method  & \textcolor{black}{Energy Consumption} &  Peak Memory Usage&  \#parameters $\bphi$\\
    {} & [kJ] & [GB] & {}\\
    \midrule
    \multicolumn{4}{c}{ Two moons} \\
    \midrule
    SNL &   24.5  &  19   & 28,020 \\
    SNPE-C &  15  &  9  & 60,925 \\
    SeMPLE  & {\bf 7.1 } &  {\bf 0.71 }   &  {\bf 449} \\
    \midrule
    \multicolumn{4}{c}{ Hyperboloid} \\
    \midrule
    SNL &  54  & 30   & 30,020 \\ 
    SNPE-C &    120  &  15  & 66,925 \\
    SeMPLE &  {\bf 3.2 } &  {\bf 0.88 } &  {\bf  1,479}  \\
    \midrule
     \multicolumn{4}{c}{ Bernoulli GLM} \\
    \midrule
    SNL &   108  & 75  & 36,100\\
    SNPE-C &    41  &  10  & 98,025\\
    SeMPLE &    {\bf 4.8 } &  {\bf 0.93 } & {\bf 2,309}\\
    \midrule
    \multicolumn{4}{c}{ Ornstein-Uhlenbeck} \\
    \midrule
    SNL &   66  & 39  & 41,030 \\
    SNPE-C &    134  &  15  & 102,242 \\
    SeMPLE &    {\bf 6.1 } &  {\bf 0.93 }   & {\bf 30,799} 
    \end{tabular}
  \label{tab1}
\end{table}

Section \ref{sec:related-work} gives a brief overview of the main SBI approaches. 
Section \ref{sec:gllim} describes how GLLiM can be used to construct surrogate posteriors and surrogate likelihoods, while the complete SeMPLE methodology is explained in Section \ref{sec:semple}. Results from SeMPLE, SNL and SNPE-C are then illustrated in Section \ref{sec:examples}. Test models, partly reported in the Appendices, include simulation studies comprising multimodal surfaces (two moons, multiple hyperboloid), stochastic differential equations (Ornstein-Uhlenbeck), generalised linear models (Bernoulli GLM), a Markov jump process with additional measurement error (Lotka-Volterra), highly correlated posteriors with 20 parameters (twisted-prior), and a challenging and partially-observed biological model of the translation kinetics after mRNA transfection. 
 Code is available at 
 \url{https://github.com/henhagg/semple}.

\section{Related work}\label{sec:related-work}
Simulation-based inference, also referred to as {\it likelihood-free inference}, is reviewed {\it e.g.} in \cite{cranmer} and \cite{pesonen2023abc}. In a Bayesian framework the focus is on posterior estimation, which in  SBI  directly relates to the quality of the approximations made to the unavailable likelihood, the posterior distribution or both. This aspect raises the issue of how to evaluate  the quality of posterior inference when a reference ground truth posterior is absent by definition. This  is still an open question and is further discussed in Section \ref{sec:conclusions}. A second important ingredient is  sampling efficiency, as simulation-based methods are typically  relying on exploration, and on Monte Carlo principles that can be prohibitively expensive. 
Active or sequential learning is a way to address sample efficiency by gradually discovering and guiding where simulations should be made for a specific given observation.
Our proposed method is sequential and 
we focus on comparisons with NN-based sequential methods for posterior inference, namely SNL \citep{papamakarios2019} and SNPE-C \citep{greenberg2019}.
Many other approaches exist for Bayesian inference with intractable likelihoods, such as sequential neural ratio estimation (SNRE, \citealp{hermans2020likelihood}), the several variants of ABC such as SMC-ABC \citep{toni2009approximate,del2012adaptive}, synthetic likelihoods (\citealp{wood2010statistical,price2018bayesian}) and particle MCMC (\citealp{andrieu2009pseudo,andrieu2010particle}). 
Comparisons between several of the above mentioned methods are available in previous works.
For example, \cite{greenberg2019} show that SNPE-C is computationally more efficient than the  predecessors\footnote{The acronyms SNPE-A and SNPE-B follow the taxonomy in \cite{papamakarios2019}.} SNPE-A \citep{papamakarios2016} and SNPE-B \citep{lueckmann2017flexible}, where {\it efficient} means that, for a given computational budget, SNPE-C approaches the true posterior using a smaller number of model simulations. \cite{sbibm} show that, for nontrivial case-studies, both SNPE-C and SNRE are more efficient than SNL and SMC-ABC. Moreover, SNPE-C and SNRE perform similarly. The above motivates why in this work we focus on the comparison with SNPE-C using the \texttt{SBIBM}  package implementation \citep{sbibm} (an exception is the Lotka-Volterra Markov jump process, where we compare against SMC-ABC, as an objective comparison with NN-based methods is difficult here, see section \ref{sec:main-lotka-volterra} for more details). Since SeMPLE includes a Markov chain Monte Carlo (MCMC) step, same as SNL, it is also relevant to compare against SNL, using its implementation in  \texttt{SBIBM}, which is an ameliorated one, compared to the original version of \cite{papamakarios2019}.

Note that there are crucial differences between our SeMPLE and SNPE-A, as detailed in \textcolor{black}{Appendix \ref{sec:snpeA}}. Briefly, in SNPE-A a $K$ component mixture density network (MDN) is used as a proposal sampler. As such, the MDN means and covariances are parametrised via neural networks. However SNPE-A does not have a built-in strategy to learn each of the $K$ components. Instead, the GLLiM procedure embedded into SeMPLE automatically determines the means and covariance matrices for each of the $K$ components, thus increasing flexibility to the shapes of the posteriors that can be targeted \textcolor{black}{(see also \citealp{wang2018meta}, where MDNs are employed as proposal functions within a Gibbs sampler). The fact that SeMPLE does not use MDNs, 
and more generally no NN architectures are employed in SeMPLE\footnote{While NN could potentially be incorporated into SeMPLE for the purpose of obtaining summary statistics of the data, as done in eg ABC, this is a separate problem that is not considered in the present work.}, it results in an easy to train framework, with fewer parameters to learn (see Table \ref{tab1}). Other architectures using NN within proposal samplers are e.g. the \textit{neural proposal} of \cite{stites2021learning} and the mixture of Gaussian copulas of \cite{chen2019adaptive}.}

Recent work that is relevant for SBI, but is less comparable with our work, includes amortized strategies. 
Amortization is a property that refers to the ability of algorithms to make use of previous computational efforts, not to repeat  computationally expensive steps for each new observation considered. 
Examples in this class are the approaches in \cite{radev2023jana}, \cite{radev2023bayesflow}, \cite{dyer2022amortised}, \cite{gao2023generalized}.
Although GLLiM is by design an amortized procedure, this property is partly lost when used in a sequential approach such as SeMPLE, as amortization and sequential learning are somewhat opposite objectives.
The next section gives a brief introduction to the Gaussian local linear mapping (GLLiM) procedure, which allows to obtain amortized surrogates of the posterior and of the likelihood in closed form.

\section{Amortized surrogates for likelihoods and posteriors via GLLiM}\label{sec:gllim}

The {\it inverse regression} approach via Gaussian local linear mapping 
(GLLiM) was first introduced in \cite{deleforge}.  GLLiM is a parametric mixture model on the joint distribution of $(\by,\btheta)$, denoted by $q_{\tilde{\bphi}}$ and depending on a set of parameters $\tilde{\bphi}$. For interpretation or visualization, the model can also be presented as an invertible regression model as follows. In GLLiM the relationship between observations $\by\in\mathbb{R}^d$ and $\btheta\in\mathbb{R}^l$ is assumed to be locally-linear with $K$ components, defined through an additional latent variable $z \in \{1,...,K\}$ as, 
\begin{equation}
    \by = \sum_{k=1}^K \mathbbm{1}_{\{z = k\}} (\tilde{\boldsymbol A}_k \btheta + \tilde{\boldsymbol b}_k + \tilde{\boldsymbol \epsilon}_k),\label{eq:affine}
\end{equation}
where  $\mathbbm{1}$ denotes the indicator function, and $\tilde{\boldsymbol A}_k\in\mathbb{R}^{d\times l}$  and $\tilde{\boldsymbol b}_k\in\mathbb{R}^d$  are matrices and vectors, respectively, defining the affine transformations, while $\tilde{\boldsymbol \epsilon}_k\in\mathbb{R}^d$ corresponds to an error term capturing both the observational
noise and the modelling error due to assuming an affine approximation for the data. In the following we consider  Gaussian noise, 
$\tilde{\boldsymbol \epsilon}_k \sim \mathcal{N}_d(\boldsymbol 0, \tilde{\boldsymbol \Sigma}_k$), and assume $\tilde{\boldsymbol \epsilon}_k$ to not depend on $\btheta$, $\by$ nor $z$.
Therefore, conditionally on component $z=k$, an approximate generative model (while the assumed generative model is denoted $p(\by|\btheta)$) is given by
\begin{equation}
    q_{\tilde{\bphi}}(\by \g \btheta, z=k) = \mathcal{N}_d(\by; \tilde{\boldsymbol A}_k \btheta + \tilde{\boldsymbol b}_k, \tilde{\boldsymbol \Sigma}_k).
\end{equation}
To complete the hierarchical model, $\btheta$ is assumed to follow a mixture of Gaussian distributions specified by 
\begin{subequations}
\begin{alignat}{2}
     q_{\tilde{\bphi}}(\btheta \g z=k) &= \mathcal{N}_l(\btheta; \tilde{\boldsymbol c}_k, \tilde{\boldsymbol \Gamma}_k), \\
     q_{\tilde{\bphi}}(z=k) &= \pi_k. \label{eq:pik}
\end{alignat}
\end{subequations}
The GLLiM model hierarchical construction above  (eq.(\ref{eq:affine}) to (\ref{eq:pik})) defines then a joint  Gaussian mixture model on $(\by,\btheta)$:
\begin{equation}
    \label{eq:gllim_decomp}
     q_{\tilde{\bphi}}(\by, \btheta)\! = \!\sum_{k=1}^K  q_{\tilde{\bphi}}(\by \!\g\! \btheta, z\!=\!k)  q_{\tilde{\bphi}}(\btheta \!\g\! z\!=\!k)  q_{\tilde{\bphi}}(z\!=\!k),
\end{equation}
where the full vector of mixture model parameters is
\begin{equation}
    \label{eq:mixture_param}
    \tilde{\boldsymbol \phi} = \{\pi_k, \tilde{c}_k, \tilde{\boldsymbol \Gamma}_k, \tilde{\boldsymbol A}_k, \tilde{\boldsymbol b}_k, \tilde{\boldsymbol \Sigma}_k\}_{k=1}^K.
\end{equation}
From this joint model, we can deduce straightforwardly both its conditional distributions $q_{\tilde{\bphi}}(\by\g\btheta)$ and $q_{{\bphi}}(\btheta\g\by)$ in closed-form.
First, it comes that,
\begin{equation}
    \label{eq:surrLik}
    q_{\tilde{\bphi}}(\by \g \btheta) = \sum_{k=1}^{K} \tilde{\eta}_k(\btheta) \mathcal{N}_d(\by; \tilde{\boldsymbol{A_k}} \btheta + \tilde{\boldsymbol{b}}_k, \tilde{\boldsymbol{\Sigma}}_k),
\end{equation}
with
\begin{equation}
    \label{eq:surrLikProb}
    \tilde{\eta}_k(\btheta) = \frac{\pi_k \mathcal{N}_l(\btheta; \tilde{\boldsymbol{c}}_k, \tilde{\boldsymbol{\Gamma}}_k)}{\sum_{j=1}^{K} \pi_j \mathcal{N}_l(\btheta; \tilde{\boldsymbol{c}}_j, \tilde{\boldsymbol{\Gamma}}_j)},
\end{equation}
and similarly, 
\begin{equation}
    \label{eq:surrPost}
    q_{{\bphi}}(\btheta \g \by) = \sum_{k=1}^{K} \eta_k(\by) \mathcal{N}_l(\btheta; \boldsymbol A_k \by + \boldsymbol b_k, \boldsymbol \Sigma_k),
\end{equation}
with
\begin{equation}
    \label{eq:surrPostProb}
    \eta_k(\by) = \frac{\pi_k \mathcal{N}_d(\by; \boldsymbol c_k, \boldsymbol \Gamma_k)}{\sum_{j=1}^{K} \pi_j \mathcal{N}_d(\by; \boldsymbol c_j, \boldsymbol \Gamma_j)},
\end{equation}
where 
$\boldsymbol{\phi} = \{\pi_k, \boldsymbol c_k, \boldsymbol \Gamma_k, \boldsymbol A_k, \boldsymbol b_k, \boldsymbol \Sigma_k\}_{k=1}^K$ is just  a convenient  equivalent reparameterization that can easily be deduced from $\tilde{\bphi}$ \citep{deleforge} as
\begin{subequations} \label{eq:forwardParam}
\begin{alignat}{3}
    \boldsymbol{c}_k &= \tilde{\boldsymbol A_k} \tilde{\boldsymbol c}_k + \tilde{\boldsymbol b}_k \label{eq:forwardParamC} \\
    \boldsymbol \Gamma_k &= \tilde{\boldsymbol \Sigma}_k + \tilde{\boldsymbol A_k} \tilde{\boldsymbol \Gamma}_k \tilde{\boldsymbol A_k}^\top \\
    \boldsymbol \Sigma_k &= \big(\tilde{\boldsymbol \Gamma}_k^{-1} + \tilde{\boldsymbol A_k}^\top \tilde{\boldsymbol \Sigma}_k^{-1} \tilde{\boldsymbol A_k} \big)^{-1} \\
    \boldsymbol A_k &= \boldsymbol \Sigma_k \tilde{\boldsymbol A_k}^\top \tilde{\boldsymbol \Sigma}_k^{-1} \\
    \boldsymbol b_k &= \boldsymbol \Sigma_k \big(\tilde{\boldsymbol \Gamma}_k^{-1} \tilde{\boldsymbol{c}}_k - \tilde{\boldsymbol A_k}^\top \tilde{\boldsymbol \Sigma}_k^{-1} \tilde{\boldsymbol b}_k \big).
\end{alignat}
\end{subequations}
The modelling power of GLLiM comes essentially from the fact that both conditional distributions above are mixtures of Gaussian affine experts (MoE), which are known for their modelling and approximation capabilities \citep{nguyen2019approximation,nguyen2021approximations,nguyen_demystifying_2023}. As such, they are good candidates to provide surrogate likelihoods and posteriors. All the more so as, given
a training dataset made of many pairs of $\{\btheta_n,\by_n\}_{n=1}^N$,
surrogate models specified via GLLiM can be  estimated via an EM algorithm, see \cite{deleforge} and \cite{forbes} for details. In fact a surrogate likelihood and posterior can be obtained with $\{\btheta_n,\by_n\}_{n=1}^N$ 
prior-predictive realizations  simulated from $p(\btheta)p(\by\g\btheta)$, but more generally, as exploited in SeMPLE,  GLLiM can be used in a sequential way, by changing the training set to include  posterior draws.

Once an estimate for the parameter vector $\tilde{\boldsymbol{\phi}}$ (or equivalently $\bphi$) is available, we have an \textit{amortized} procedure providing a surrogate likelihood function $q_{\tilde{\bphi}}(\by \g \btheta)$,
which can be evaluated for any data $\by$.
In particular, if we specify $\by$ to be a given {\it observed dataset} $\by_o$, and plug $\by_o$ into \eqref{eq:surrLik}, we obtain an approximate (surrogate) likelihood function $q_{\tilde{\bphi}}(\by_o \g \btheta)$ for observation $\by_o$. Similarly, and using the same training data, 
an amortized surrogate posterior $q_{\bphi}(\btheta \g \by)$, for a generic $\by$, can also be derived as a mixture of Gaussian experts distributions from (\ref{eq:surrPost}).
If we plug $\by_o$ into \eqref{eq:surrPost}, then $q_{\bphi}(\btheta \g \by_o)$ is an approximate posterior based on observed data. In summary, once $\tilde{\bphi}$ is estimated via EM,  we obtain surrogates of both the likelihood and the posterior in the same GLLiM run, simultaneously and using the same training data.
The way this is performed and integrated with the rest of the SeMPLE method is explained in Section \ref{sec:semple}. Note that it is possible to assume a different distribution for $\tilde{\boldsymbol \epsilon}_k$ than the Gaussian one. For example \eqref{eq:affine} can be a mixture of generalized Student distributions, resulting in the so-called SLLiM method \citep{Perthame2018}. Both GLLiM and SLLiM are implemented in the R package \texttt{xLLiM} \citep{xllim} which we employ in our experiments, see Section \ref{sec:examples} for details. 
\textcolor{black}{Notice that GLLiM has already been used in SBI, see \cite{forbes}, namely they produced an amortized sampler whose posterior draws were then ``refined'' via a single ABC step. However, in \cite{forbes} only surrogate posteriors were investigated and, in addition, they did not sequentially refine the surrogate posterior and surrogate likelihood. In contrast, our SeMPLE approach  makes use of both posterior and likelihood surrogates, to then provide an automatically ``tuned'' proposal function for MCMC (as described in Section \ref{sec:semple}) but also to provide a surrogate likelihood, which finds excellent use in the MCMC we construct in Section \ref{sec:semple}. Then, SeMPLE   sequentially refines, in multiple rounds, both the surrogate likelihood and the surrogate posterior (and hence the proposal function), without requiring an ABC step, and hence without the critical tuning/setup decisions that ABC inference entails (eg the choice of distances, threshold values, the determination of suitable proposal functions), and without the large number of simulations that ABC requires.}

\textcolor{black}{Another approach that, similarly to GLLiM, could be considered for closed-form approximations of densities without involving neural networks, is the ``elliptical process'' of \cite{baankestad2020elliptical}, in particular when using their piecewise constant mixing distribution. }

\begin{algorithm}[ht]
\caption{SeMPLE}\label{alg:semple}
\begin{algorithmic}[1]
\State $\tilde{p}_0(\btheta) \gets p(\boldsymbol\theta)$, set starting $K$ and number of rounds $R$. 
\For{$r = 0:R$}
        \If{$r=0$ or $r=1$} 
        \State sample $\{\btheta_n\}_{n=1}^N$ iid $\btheta_n\sim \tilde{p}_r(\btheta)$ (without MCMC), $n=1,...,N$. 
        \Else 
        \State sample $\{\btheta_n\}_{n=1}^N$ via  $\boldsymbol\theta_n \sim \tilde{p}_r(\btheta)$ using Algorithm \ref{alg:MH_gllim}. \label{alg:jass:mcmc}
        \EndIf
        \State Conditionally on every $\btheta_n$, obtain $\{\by_n\}_{n=1}^N$ by sampling (implicitly) $\by_n \sim p(\by\g\boldsymbol\theta_n)$ ($n=1,...,N$).
    \If{r = 0}
        \State Collect $\mathcal{D}_0 = \{\boldsymbol\theta_n,\by_n\}_{n=1}^{N}$.
        \State Train $q_{\tilde{\bphi}_0}(\by\g\btheta)$ on $\mathcal{D}_0$. Update $K$. 
        \State Obtain $q_{\bphi_0}(\boldsymbol\theta\g\by)$. 
        \State Set $\tilde{p}_{1}(\boldsymbol\theta) \gets q_{\bphi_0}(\boldsymbol\theta \g \by=\by_o)$.
    \Else
        \State Collect $\mathcal{D}_r = \mathcal{D}_{r-1}\cup\{\btheta_n,\by_n\}_{n=1}^N \setminus \mathcal{D}_0$.
        \State Train $q_{\tilde{\bphi}_r}(\by\g\btheta)$ on $\mathcal{D}_r$. Update $K$. 
        \State Obtain $q_{\bphi_r}(\btheta\g\by)$. 
        \State Set $\tilde{p}_{r+1}(\btheta) \gets q_{\tilde{\bphi}_r}(\by_o \g \btheta) p(\btheta)$ \label{alg:jass:likorpost}
        
        \State\quad ((\textit{optional}) $\tilde{p}_{r+1}(\btheta) \gets \frac{p(\btheta)}{\tilde{p}_{r}(\boldsymbol\theta)} q_{\bphi_r}(\btheta\g \by_o)$)
    \EndIf
\EndFor
\end{algorithmic}
\end{algorithm}
\begin{algorithm}[ht]
\caption{Independence MH with GLLiM surrogate posterior as proposal distribution}\label{alg:MH_gllim}
\begin{algorithmic}[1]
\State For the initial value $\btheta_1$, pick the last accepted value from the currently available Markov chain.
\For{$j = 2:N$}
    \State Propose $\btheta^* \sim q_{\bphi_{r-1}}(\btheta\g \by_o)$,
    \State $\alpha = \min\{1, \frac{\tilde{p}_r(\btheta^*)}{\tilde{p}_r(\btheta_{j-1})} \frac{q_{\bphi_{r-1}}(\btheta_{j-1}\g\by_o)}{q_{\bphi_{r-1}}(\btheta^*\g\by_o)}\}.$
    \State Sample $u \sim \mathcal{U}[0,1]$
    \If{$u \leq \alpha$}
        \State $\btheta_j = \btheta^*$, 
    \Else
        \State $\btheta_j = \btheta_{j-1}$.
    \EndIf
\EndFor
\end{algorithmic}
\end{algorithm}

\section{SeMPLE: sequential learning of likelihoods and posteriors}\label{sec:semple}

We have shown how GLLiM can provide amortized estimations of both the likelihood and the posterior distribution at no additional cost, given training data. This GLLiM feature is exploited in our proposed Sequential Mixture Posterior and Likelihood Estimation (SeMPLE), to obtain a sequence of increasingly improved samplers, to more accurately and more efficiently sample from the targeted posterior of observed data $\by_o$. 
Similarly to SNL \citep{papamakarios2019}, given a likelihood approximation $q_{\tilde{\bphi}}(\by_o \g \btheta)$, SeMPLE targets an approximate posterior $\tilde{p}(\btheta)\propto q_{\tilde{\bphi}}(\by_o \g \btheta) p(\btheta)$ with posterior sampling being carried out using a MCMC procedure. 
Typically, the MCMC scheme used in SNL is slice sampling \citep{neal2003slice}, which has the advantage to be tuning-free but is likely to have difficulties in exploring high dimensional posteriors with complex multimodalities. As a matter of fact,  \cite{papamakarios2019} recommend SNL ``at most for tens of parameters that do not have pathologically strong correlations''.
In SeMPLE, since GLLiM is able to learn a closed-form approximation of the posterior distribution (in addition to a likelihood function), such posterior is used as a proposal distribution within a Metropolis-Hastings (MH) algorithm.
SeMPLE, by using an efficient EM procedure to fit a flexible proposal distribution, is expected to benefit from a MH step that better guides simulations towards more likely parameter values. To ease the comparison with SNL, the latter is recalled in \textcolor{black}{Appendix \ref{sec:snl}}, while SeMPLE is summarized in Algorithms \ref{alg:semple}-\ref{alg:MH_gllim}. 

Algorithm \ref{alg:semple} shows that the targeted  approximate posterior for $\btheta$ is a function denoted with $\hat{p}_r(\btheta)\propto q_{\tilde{\bphi}_r}(\by_o \g \btheta) p(\btheta)$ which is sequentially obtained, where $r=0,...,R$ denotes the SeMPLE {\it rounds} for the current approximation\footnote{We use the term {\it round} instead of the most common {\it iteration}, as the latter could be confused with the MCMC iterations that are run within a given SeMPLE round.}. \textcolor{black}{In the following we denote by $\btheta^{(r)}$ and $\by^{(r)}$ a parameter and an observation simulated in round $r$, respectively}. At first ($r=0$) $\hat{p}_0(\btheta)$ is initialised at the prior $p(\btheta)$. An initial batch of $N$ pairs $\mathcal{D}_0=\{\btheta_n^{(0)},\by_n^{(0)}\}_{n=1}^N$ produced from the prior-predictive distribution (\textcolor{black}{step 4}) constitutes the starting training data for GLLiM, and the parameters $\tilde{\phi}_0$ and ${\phi}_0$ are obtained via EM within GLLiM to produce the initial surrogate likelihood and posterior $q_{\tilde{\bphi}_0}(\by\g\btheta)$ and $q_{\bphi_0}(\boldsymbol\theta\g\by)$ respectively (\textcolor{black}{steps 11-12}). The initial surrogate posterior becomes the current target (\textcolor{black}{step 13}) $\tilde{p}_1(\btheta)  \equiv q_{\bphi_0}(\boldsymbol\theta\g\by_o)$. In the next \textcolor{black}{round} ($r=1$), the latter can be sampled from without MH (\textcolor{black}{step 4}) since  $\tilde{p}_1(\btheta)$ is defined as a Gaussian mixture, from which it is trivial to sample $N$ parameters $\{\btheta_n^{(1)}\}_{n=1}^N$. These $N$ draws  are then plugged into the model simulator to produce $\by_n^{(1)} \sim p(\by\g\boldsymbol\theta_n^{(1)})$ and define $\mathcal{D}_1=\{\btheta_n^{(1)},\by_n^{(1)}\}_{n=1}^N$ as the new training data (we discard $\mathcal{D}_0$ for future \textcolor{black}{rounds} as it is deemed too uninformative for training). GLLiM is then fitted to $\mathcal{D}_1$ and the corresponding surrogate likelihood and surrogate posterior are obtained. From this step onwards, the surrogate posterior is employed as a proposal distribution within a tuning-free MH algorithm (Algorithm \ref{alg:MH_gllim}) since, for an arbitrary prior $p(\btheta)$, when $r\geq 2$ we do not explicitly know which distribution is proportional to the product $p(\btheta)q_{\tilde{\phi}_r}(\by_o\g \btheta)$ (for $r=1$ we know by construction that the targeted posterior is a Gaussian mixture), except for the special case of Gaussian and Uniform priors (see \citealp{papamakarios2016}).  
In Algorithm \ref{alg:MH_gllim}, the acceptance probability of a proposed $\btheta^*$ is (for $r> 1$) 
\[\alpha = \min\biggl\{1, \frac{\tilde{p}_r(\btheta^*)}{\tilde{p}_r(\btheta_{j-1})} \frac{q_{\bphi_{r-1}}(\btheta_{j-1}\g\by_o)}{q_{\bphi_{r-1}}(\btheta^*\g\by_o)}\biggr\}.\]
Samples from the posterior $\tilde{p}_r$ are obtained by first proposing parameters $\btheta^*$ from a Gaussian mixture model. From the current proposal function 
$q_{\phi_{r-1}}(\btheta\g \by_o)$, which is the mixture $$q_{\phi_{r-1}}(\btheta\g \by_o)=\sum_{k=1}^{K} \eta_k^{(r-1)}(\by_o)\mathcal{N}_l(\btheta; \boldsymbol A_k^{(r-1)} \by_o~\!\!+\!\!~\boldsymbol b_k^{(r-1)}, \boldsymbol \Sigma_k^{(r-1)}),$$ a component $k^*$ is randomly sampled from the set $\{1,...,K\}$ with associated probabilities $\{\eta_1^{(r-1)}(\by_0),...,\eta_K^{(r-1)}(\by_0)\}$. Then  $\btheta^*$ is sampled from the $k^*$-th  component $\mathcal{N}_l(\btheta; \boldsymbol A_{k^*}^{(r-1)} \by + \boldsymbol b_{k^*}^{(r-1)}, \gamma \boldsymbol\Sigma_{k^*}^{(r-1)})$ by (optionally) slightly inflating its covariance matrix with a factor $\gamma\geq1$, which we set to $\gamma=1$ (no inflation) or $\gamma\in[1.1,1.2]$ in our experiments. The inflation factor is to ensure that the proposal is sampled from a broad-enough distribution covering the support of the target (no inflation factor is used when $r=1$). Also, note that the $N$ retained draws via MH are $N$ post-burnin draws, where the burnin consists of 100 iterations, and the last accepted draw from round $r$ of SeMPLE is used to initialize the chain at round $r+1$.

Importantly, our MCMC procedure is tuning-free. We use an \textit{independence sampler} \citep{robert2004monte}, relying on GLLiM to provide an informative proposal function. This proposal function is trained on draws accepted in previous rounds which are, reasonably, more spread than the currently targeted posterior. In addition, being a mixture model, it is suitable to deal with multimodalities in the target. This greatly simplifies the construction of SeMPLE as no complex tuning is necessary for the MCMC sampler. 
While $K$ could in principle be set arbitrarily large, in practice this may slow the EM algorithm convergence. In SeMPLE, the value of $K$ is allowed to decrease between \textcolor{black}{rounds}, as expressed with {``\it update $K$''} in lines \textcolor{black}{11 and 16} of Algorithm \ref{alg:semple}. In fact, mixture components having probabilities $\tilde{\eta}_k$  smaller than a threshold are discarded (ie these gets removed from the mixture model).
In most cases we find that setting a starting value of $K=20$ or 30 produces good results, with  a threshold for $\tilde{\eta}_k$ typically set to 0.005 (multiple hyperboloid, Ornstein-Uhlenbeck, Lotka-Volterra, biological SDE model), 0.03 (Bernoulli GLM model) or simply 0 for no deletion (Two Moons model). This is to prevent having too many irrelevant components, that could be problematic when fitting the mixture to training data produced via MCMC, as the latter may not manage to visit tiny modes. 
It is also possible to choose the starting value of $K$ in a principled way.
In \cite{deleforge}, the Bayesian Information
Criterion (BIC) is used to select $K$, as we do in some of our examples, see \textcolor{black}{Appendix \ref{sec:bic}}.
An appropriate value for $K$ can be learned by repeatedly computing the BIC on realizations of the prior-predictive distribution, before SeMPLE is started. This use of the BIC is an \textit{amortized} procedure, as  the selected values can be reused regardless the specific observed data $\by_o$. The impact on the overall computational budget is negligible.
We have also experimented with a version of SeMPLE that learns a {\it corrected posterior} (see the {\it optional} part in step 19 of Algorithm \ref{alg:semple}) using the correction formula from \cite{papamakarios2016}. However, we found that learning a surrogate likelihood was much more beneficial, providing better inference coupled with a higher acceptance rate. Notice that none of our reported results use the optional corrected-posterior approach.

In summary, a major difference with the several variants of sequential neural posterior estimation (where the most used one is SNPE-C, \citealp{greenberg2019}), and with 
SNL, is that SeMPLE uses Gaussian mixtures learned via EM to approximate both the likelihood function and the posterior distribution of $\btheta$, as opposed
to neural density estimation. Thus no  NN architecture has to be designed, nor  special stochastic gradient descent (SGD) setup has to be considered to fit NNs. Only the number of components $K$ and the structure of the covariance matrices $\tilde{\bm{\Sigma}}_k$ (as the $\bm{\Sigma}_k$ are obtained as an algebraic by-product), according to the options offered by the \texttt{xLLiM} package\footnote{The $\tilde{\bm{\Sigma}}_k$'s can be isotropic, diagonal or full matrices, and set all equal or varying with $k$.}, have to be set.  As we show in our examples, inference based on procedures using SGD can be much more computationally demanding and time consuming. 

\section{Numerical illustrations}\label{sec:examples}

We illustrate SeMPLE on several simulation studies, see also the 
\textcolor{black}{Appendices} for extra results and additional models. 
SeMPLE is written in R and uses the GLLiM implementation from the \texttt{xLLiM} package
\citep{xllim}. 
Comparisons with neural-based likelihood and posterior estimation methods (SNL and SNPE-C)  use the Python implementations in the \texttt{SBIBM} package for SBI benchmarking \citep{sbibm}, and architecture specifications for \texttt{SBIBM} are in \textcolor{black}{Appendix \ref{sec:sbibm}}. 
To compare approximate posteriors with a suitable reference posterior, we consider the classifier two-samples test (C2ST), as implemented in \texttt{SBIBM}, and Wasserstein distances via the \texttt{POT} Python package \citep{flamary2021pot}. C2ST varies in the interval [0.5,1], the lower the better. It equals 0.5 when samples from two distributions appear statistically indistinguishable  and  equals 1 when the classifier is able to perfectly separate samples as belonging to two different distributions. 
We also report runtimes in figures and resource requirements in Table \ref{tab1}. 
For SeMPLE, the time required to select $K$ via BIC (when the BIC procedure is performed, which is not the case for all studies) is reported separately and not taken into account in the figures below. In fact, for some case studies, such as Lotka-Volterra, we merely picked a starting value of $K$ without running the BIC procedure. When we used the BIC, adding this overhead time does not change the overall conclusion that SeMPLE is much more time-efficient than the other methods. Moreover, the selection of $K$ is an amortized procedure, hence it does not need to be repeated for  different observations of the same model. 

In the next two sections \ref{sec:main-twomoons} and \ref{sec:main-hyperboloid}, we illustrate results from two models having multimodal posteriors, in section \ref{sec:main-lotka-volterra} we summarise the Lotka-Volterra experiment, and in section \ref{sec:results-summary} and Table \ref{tab:results-summary} we summarise results from other experiments detailed in \textcolor{black}{the Appendices}.
For comparison, all methods are given the same  budget of model simulations. For SNL and SNPE-C the \texttt{SBIBM} default of distributing these simulations uniformly across $R=10$ rounds is used, but to ease comparison we also run SNL and SNPE-C with the same number of rounds used for SeMPLE, which is $R=4$, except for the Bernoulli GLM model where $R=2$ suffices. 
In fact, we observe that SeMPLE does not need as many sequential steps as the other methods. 
It may be argued that SNL could  perform better if a larger number of MCMC iterations was used, at the cost of a smaller number of  rounds $R$. Results in \textcolor{black}{the Appendices} show that both SNL and SNPE-C perform  worse under this alternative configuration. 
Regarding SeMPLE, when the computational budget limits the number of model simulations, we found it beneficial to keep the number of rounds $R$ between 2 and 3, where $R=2$ implies that we have exactly one round where MCMC is used, and this is typically already enough for satisfying inference, though usually an additional round with MCMC (hence $R=3$) produces further refined inference. In our work we show results up to $R=4$, but in general, four rounds are not recommended. The algorithm performance relies first on the GLLiM ability to learn reasonably accurate likelihood and posterior approximations and then on the MCMC sampler to explore the parameter space. Both steps require a large enough number of  model simulations, which depends mainly on the model dimension.

\subsection{Two Moons}\label{sec:main-twomoons}

The Two Moons model is a standard benchmark example in SBI \citep{greenberg2019,sbibm,picchini2022guided} to illustrate how algorithms deal
with multimodality and the local crescent shapes of the posterior distribution. The  model itself is defined in \textcolor{black}{Appendix \ref{sec:twomoons}}. Our experiments are run on the same 10 datasets provided in \texttt{SBIBM}, each of them being a vector $\by_0$ of length two. We use $K=30$, a value we determine as described in \textcolor{black}{Appendix \ref{sec:bic}}  (this determination only required 90 seconds). We run SeMPLE, SNL and SNPE-C algorithms for a total budget of $10^4$ model simulations, uniformly distributed across inference rounds, and we used both $R=4$ and $R=10$ rounds. The obtained samples are then  compared  with the corresponding posterior reference samples (obtained via MCMC, since the likelihood is tractable) provided in \texttt{SBIBM}. The performance metric results are averaged over the 10 different data sets. 
Figure \ref{fig:TM_settings_metric_vs_sims} shows C2ST and runtime values. SeMPLE provides the best approximation while SNL performs the worst. The comparison with SNL is particularly interesting as while both SNL and SeMPLE incorporate a tuning-free MCMC step, SeMPLE is able to visit both modes with an acceptance rate of 60-70\%, while SNL often struggles in efficiently exploring a multimodal posterior as demonstrated by the higher C2ST values. Moreover, in this simple example the runtime for SeMPLE is about half the SNPE-C runtime, and is about three times lower than the SNL runtime. In other examples, ({\it e.g.} Figure \ref{fig:hyperb_settings_metric_vs_sims}), the speedup provided by SeMPLE is even more dramatic. All these benefits from SeMPLE come with a very small impact on the memory usage, see Section \ref{sec:resource-requirements} for details.

\subsection{Multiple hyperboloid model}\label{sec:main-hyperboloid}

\begin{figure}[h]
     \centering
     \includegraphics[scale=0.35]{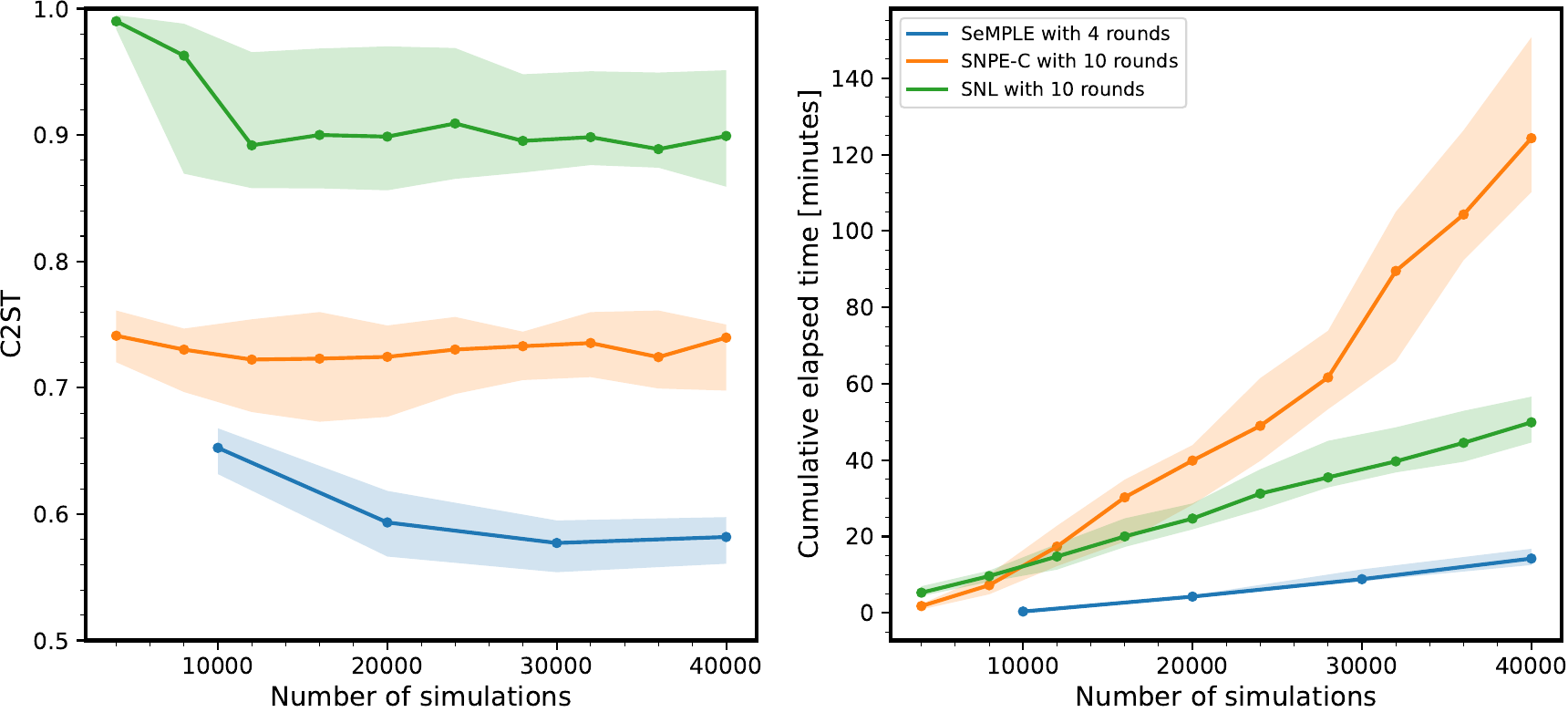}
     \caption{Hyperboloid. Median C2ST (left) and median cumulative runtime in minutes (right) for 10 runs with the same data vs the number of model simulations. Shaded bands enclose the min and max values.}
     \label{fig:hyperb_settings_metric_vs_sims}
\end{figure}

\begin{figure}[h]
    \centering
    \includegraphics[scale=0.45]{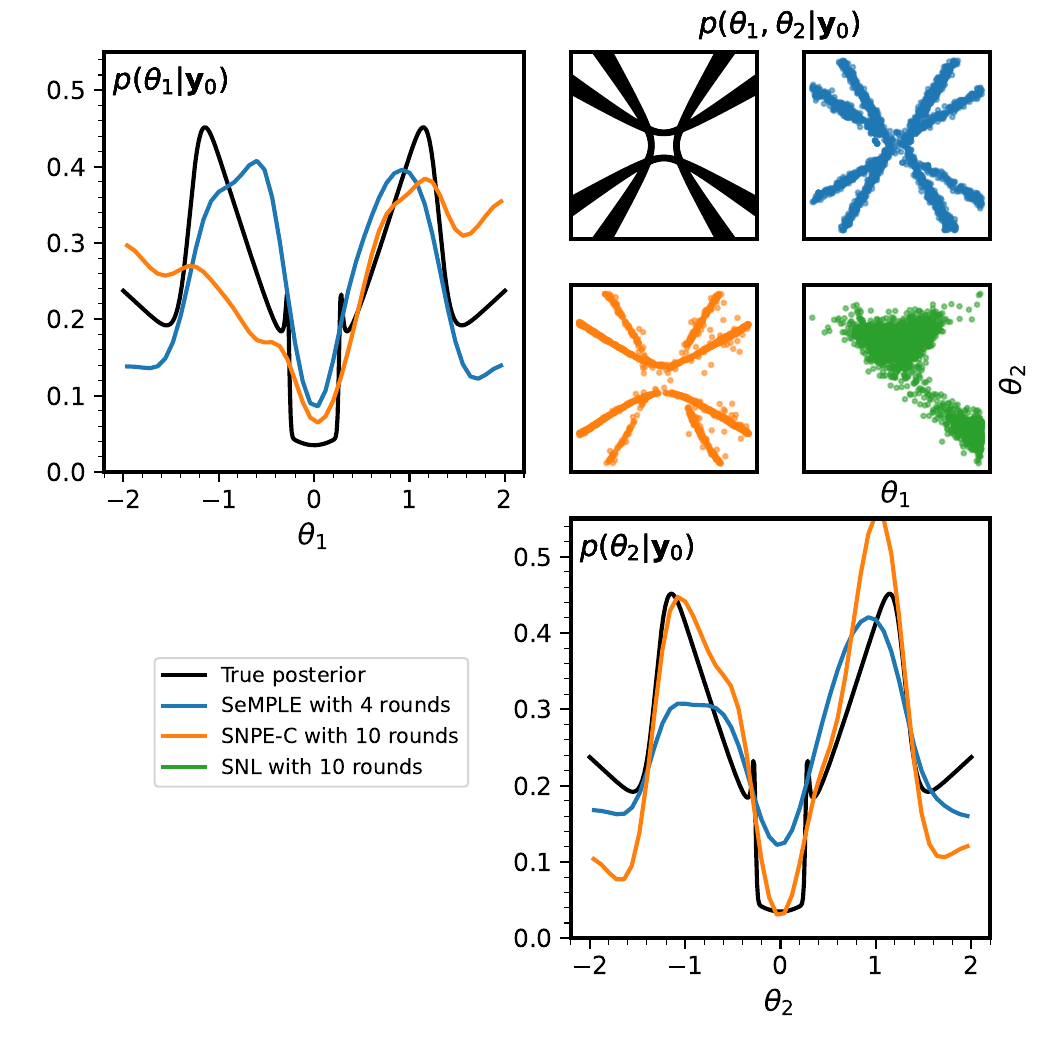}
    \caption{Hyperboloid. An example of marginal posteriors and samples from the last \textcolor{black}{round} of each algorithm.
    The SNL marginal posteriors are not reported as their inference is completely off.}
    \label{fig:hyperb_run5}
\end{figure}

The multiple hyperboloid model \citep{forbes} is constructed from a sound source localization problem in audio processing. This is a challenging inference task with a complex symmetrical and multimodal posterior distribution. The model is provided in \textcolor{black}{Appendix  \ref{sec:hyperboloid}}. The observation is a vector of length 10 and $\btheta$ has dimension 2. SeMPLE is run with $K=40$, a value that is selected via BIC in 4.5 minutes.  The total number of model simulations is set to $4\times 10^4$.
C2ST and runtime results are reported in Figure \ref{fig:hyperb_settings_metric_vs_sims}. An exemplary posterior is shown in Figure \ref{fig:hyperb_run5}. SeMPLE shows the best performance according to C2ST, and its runtime is about half that of SNL, and is about 10 times more efficient than SNPE-C. Adding the BIC overhead does not change the conclusion that SeMPLE is much more efficient. 
The runtime comparison with SNL is not particularly relevant, since the latter does not generally manage to explore the  multimodal posterior (see Figure \ref{fig:hyperb_run5} and the posteriors in \textcolor{black}{Appendix \ref{sec:hyperboloid}}).

\subsection{Lotka-Volterra}\label{sec:main-lotka-volterra}

\begin{figure}[h]
    \centering
    \includegraphics[scale=0.47]{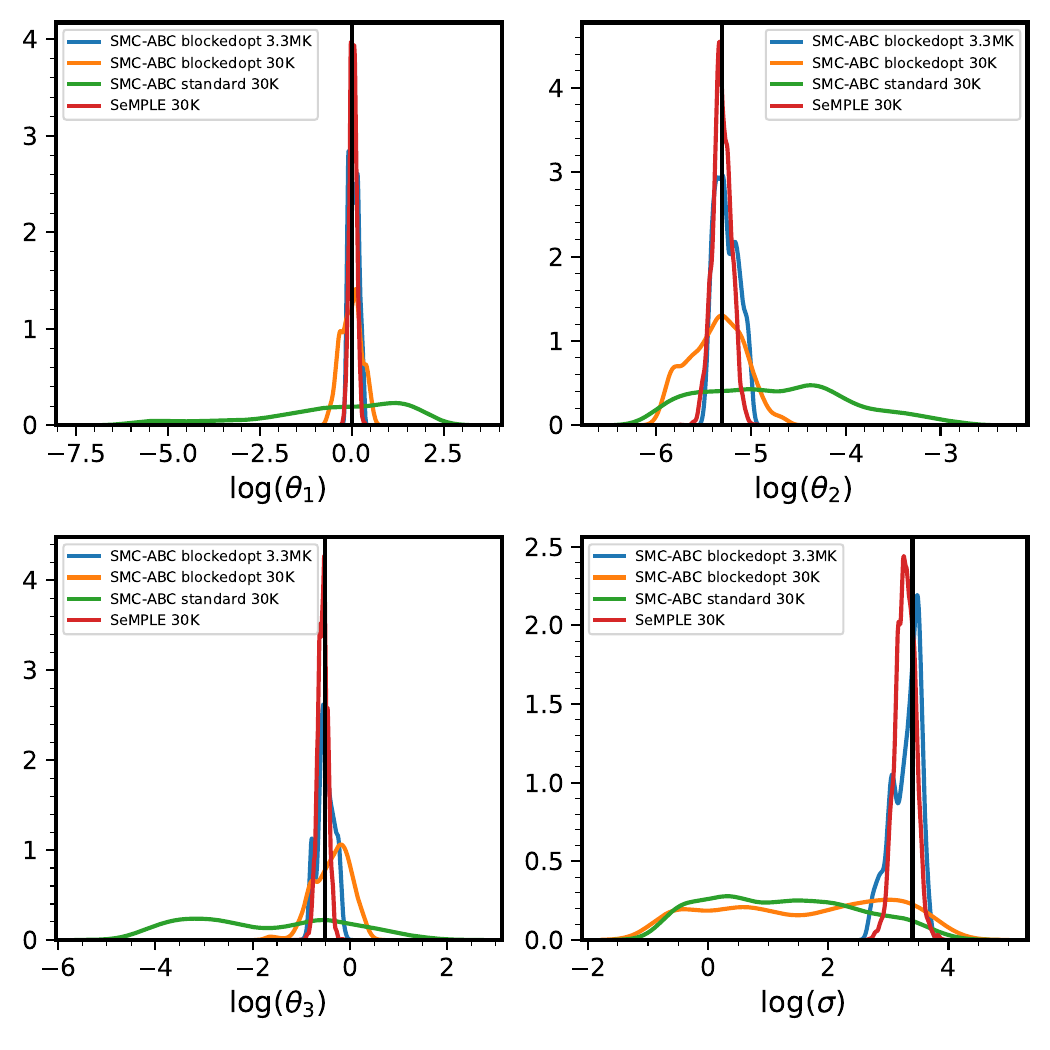}
    \caption{Lotka-Volterra. Marginal posteriors from SMC-ABC and SeMPLE. SeMPLE is run for a total of 30,000 model simulations. We report posterior samples from \texttt{blockedopt} SMC-ABC after a total of 32,865 and 3,337,640 model simulations. We also report the posterior samples from \texttt{standard} SMC-ABC after a total of 30,436 model simulations. True parameter values are indicated with black vertical lines.}
    \label{fig:lv-kde-obs2}
\end{figure}

The Lotka-Volterra predator prey model is a classical example of a population dynamics model. We consider the stochastic Markov jump process version of the model, which is defined in \textcolor{black}{Appendix \ref{sec:lotka-volterra}}. This model, although being relatively simple and characterized by a set of three interactions on the two species, each with a corresponding rate $\theta_j$ ($j=1,2,3$),  encompasses many of the difficulties associated with larger, more complex systems. The observations consider the addition of non-negligible measurement error to the simulated dynamics, and the standard deviation $\sigma$ of this noise term is inferred as an additional parameter, resulting in a total of four model parameters to infer $(\theta_1,\theta_2,\theta_3,\sigma)$. For this model, we are not primarily focusing on the performance comparison to other neural-based methods, as this version of the model, as a Markov jump process, is not included in \texttt{SBIBM}. Moreover, running times for a single model call are random even for repeated calls using the same parameter value. We use the Gillespie algorithm, as explained in \textcolor{black}{Appendix \ref{sec:lotka-volterra}}, and the number of ``reactions'' at any given $\btheta$ is stochastic. Due to the reasons above, the primary focus is to show that high quality inference can be obtained with SeMPLE using a comparably small computational effort, and compare our results against two versions of sequential Monte Carlo ABC (SMC-ABC). A ``reference posterior'' is obtained with an efficient version of SMC-ABC called \texttt{blockedopt} in \cite{picchini2022guided}, which is  run for very many model simulations, around 3.3 million, see \textcolor{black}{Appendix \ref{sec:lotka-volterra}}  for details. We also compare it with a more standard off-the-shelf SMC-ABC named \texttt{standard}. SeMPLE inference is obtained with only 30,000 model simulations. In Figure \ref{fig:lv-kde-obs2} we also include the corresponding inference obtained with around 30,000 model simulations for both \texttt{blockedopt} and \texttt{standard} SMC-ABC.  With only 30,000 model simulations, SeMPLE manages to produce inference for all parameters that is similar to the inference returned after 3.3 million model simulations from \texttt{blockedopt} SMC-ABC, a 100-fold computational saving against the most efficient version of SMC-ABC. We observe that, unlike SMC-ABC, SeMPLE produces accurate posterior inference for all model parameters, including the noise parameter $\sigma$, but with a much smaller number of model simulations compared to SMC-ABC. The latter fact, namely the SMC-ABC inefficiency in terms of model simulations, was not unexpected and known from \cite{greenberg2019,sbibm}. However it is reassuring to notice the excellent inference quality brought by SeMPLE in this challenging case study.

\subsection{Biological model of the translation kinetics after mRNA transfection}

\begin{figure}[htbp]
    \centering
    \includegraphics[width=13cm,height=7cm]{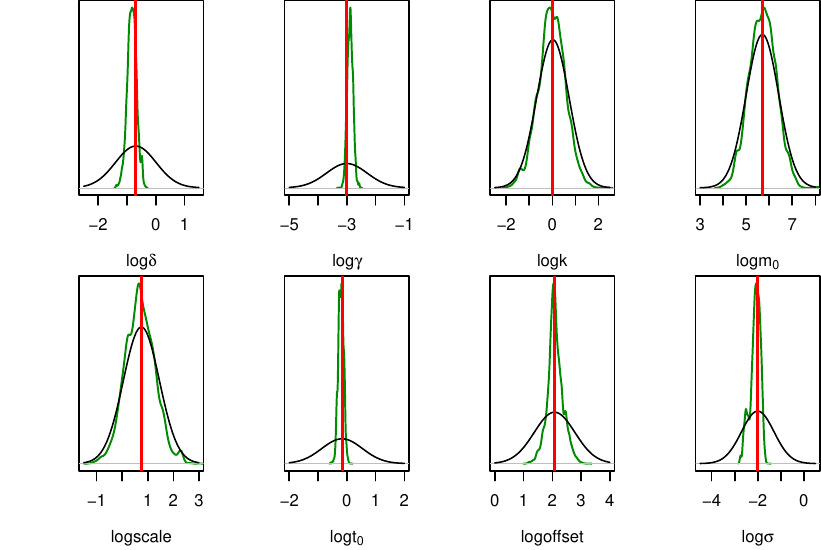}
    \caption{\textcolor{black}{Translation kinetics} model: in green are the marginal posteriors from SeMPLE (at round $r=3$), priors (solid black lines) and ground truth parameters (vertical red lines).}
    \label{fig:mrna_semple}
\end{figure}

We consider a two-dimensional stochastic differential equation (SDE) model of the translation kinetics after mRNA transfection, as studied in \cite{pieschner2022identifiability}. Only one of the two coordinates of the process is observed (with measurement noise). Details about the model are in Appendix \ref{sec:mRNA}. \cite{pieschner2022identifiability} perform several experiments, where data are either exact observations of the system or are perturbed with measurement error, and inferences for both an ODE and an SDE model are compared. However, here we only study the SDE model with data perturbed with measurement error.
The parameter to infer is (we consider parameters on log-scale, same as for their priors) $\theta=(\log\delta,\log\gamma,\log k,\log m_0,\log\textrm{scale},\log t_0,\log\textrm{offset},\log\sigma)$. We ran inference for five datasets, each consisting of a time-series of length 60, each generated with different parameter values.
Details about data-generating parameters, and other settings, are in Appendix \ref{sec:mRNA}. 
For one of the five datasets, in Figure \ref{fig:mrna_semple} we show the marginal posteriors from the third round of SeMPLE, in comparison to the parameter priors, while in Appendix \ref{sec:mRNA} we also show comparisons with SNL and SNPE-C. For all datasets, SeMPLE accurately infers $(\delta,\gamma,t_0,\textrm{offset},\sigma)$, and we confirm the findings in \cite{pieschner2022identifiability} (which they obtain using exact Bayesian inference via the Hamiltonian Monte Carlo method implemented in \texttt{Stan}), namely that $k$, $m_0$ and $\textrm{scale}$ suffer from identifiability issues (see the supplementary section A.4.1 in \citealp{pieschner2022identifiability} where their parameter $\theta_2$ is our $k$). In Appendix \ref{sec:mRNA} we show that only SeMPLE can identify $t_0$ accurately, while SNL and SNPE-C provide a biased estimation of $t_0$. Moreover, SNPE-C is often unable to infer the measurement error variability $\sigma$, while this is not a problem for SeMPLE.

\subsection{Resource requirement metrics}\label{sec:resource-requirements}

The metrics reported in Table \ref{tab1} correspond to four models and have been measured in kiloJoules (kJ) \textcolor{black}{using the \texttt{PyJoules}\footnote{\url{https://pyjoules.readthedocs.io/en/latest/}} module found in the \texttt{PowerAPI} package \citep{powerapi} in Python, which makes use of the \textit{Running Average Power Limit} (RAPL) technology available on Intel\textsuperscript{\tiny\textregistered} CPUs. It is important to note that these energy measurements are based on hardware readings, and not on estimations depending on eg running time.} The experiments were run on a machine with a CPU Intel\textsuperscript{\tiny\textregistered} Core\textsuperscript{TM} i7-4790 CPU @ 3.60GHz and 16GB of DRAM.  Memory cost and peak memory usage results have been measured by the \texttt{tracemalloc} Python library. These results highlight the advantage brought by SeMPLE, which can be run on a minimal configuration while maintaining good performance. 
For instance, in the Two Moons case,  the memory requirement (Peak Memory Usage) for SeMPLE is 13 times lower than SNPE-C and 27 times lower than SNL. The gain is even larger for the other models in Table \ref{tab1}. Similarly, for the Ornstein-Uhlenbeck model the energy consumption (CPU+DRAM) is more than  10 times smaller with SeMPLE.

\subsection{Results summary}\label{sec:results-summary}

\begin{table}[t]
  \centering
  \caption{Median metric results, across 10 independent runs (brackets include minimum and maximum values), and median runtime for posterior inference. 
  Best (lowest) values are in bold. Notice, ``Metric (final)'' reports the metric at the final inference round, instead ``\#Model sims'' gives the total number of model simulations across all rounds for a single run, and similarly for ``Runtime''.
  SNPE-C and SNL used the default \texttt{SBIBM} setup except when the number of algorithm \textcolor{black}{rounds} is reduced from 10. The C2ST metric is reported for multimodal posterior distributions and the Wasserstein distance for unimodal posterior distributions according to the discussion in Section \ref{sec:results-summary}.}

    \begin{tabular}{c|c|c|c|c}
    Method &  Metric (final) & \#Model sims & Runtime (minutes) & \#Rounds \\
    \midrule
    \multicolumn{5}{c}{ Two moons (multimodal)} \\
    \midrule
    SNL & C2ST=0.62 [0.56, 0.92] & 10K & 19 [17, 31] & 4 \\
    SNL & C2ST=0.60 [0.56, 0.68] & 10K & 22 [21, 23] & 10 \\
    SNPE-C &  C2ST=0.59 [0.54, 0.66] & 10K & {\bf 8.1} [7.1, 12] & 4 \\
    SNPE-C &  C2ST=0.57 [0.52, 0.60] & 10K & 15 [13, 17] & 10 \\
    SeMPLE  &  C2ST={\bf 0.54} [0.50, 0.58] & 10K &  8.3 [7.9, 8.8] & 4 \\
    \midrule
    \multicolumn{5}{c}{ Hyperboloid (multimodal)} \\
    \midrule
    SNL &   C2ST=0.90 [0.84, 1.0] & 40K & 35 [30, 39] & 4 \\
    SNL &   C2ST=0.90 [0.86, 0.95] & 40K & 50 [45, 57] & 10 \\
    SNPE-C &  C2ST=0.76 [0.74, 0.81] & 40K & 100 [76, 150] & 4 \\
    SNPE-C &  C2ST=0.74 [0.70, 0.75] & 40K & 124 [110, 151] & 10 \\
    SeMPLE  &  C2ST={\bf 0.58} [0.56, 0.60] & 40K & {\bf 14} [13, 17] & 4 \\
    \midrule
     \multicolumn{5}{c}{ Bernoulli GLM (unimodal)} \\
    \midrule
    SNL &  Wasserstein={\bf 0.58} [0.44, 0.71] & 10K & 22 [22, 24] & 2\\
    SNL &  Wasserstein=0.69 [0.46, 0.89] & 10K & 91 [91, 92] & 10\\
    SNPE-C &  Wasserstein=0.94 [0.65, 1.6] & 10K & 9.5 [4.9, 15] & 2\\
    SNPE-C &  Wasserstein=0.71 [0.49, 1.78] & 10K & 31 [25, 40] & 10\\
    SeMPLE  & Wasserstein=0.77 [0.43, 1.0] & 10K & {\bf 1.2} [0.98, 1.2] & 2\\
    \midrule
    \multicolumn{5}{c}{ Ornstein-Uhlenbeck (unimodal)} \\
    \midrule
    SNL &  Wasserstein={\bf0.067} [0.057, 0.23] & 40K & 56 [54, 60] & 4 \\
    SNL &  Wasserstein=0.12 [0.079, 0.20] & 40K & 62 [58, 71] & 10 \\    
    SNPE-C &   Wasserstein=0.76 [0.27, 1.0] & 40K & 100 [83, 150] & 4 \\
    SNPE-C &  Wasserstein=0.34 [0.19, 0.48] & 40K & 130 [100, 150] & 10 \\
    SeMPLE  &  Wasserstein=0.15 [0.11, 0.28] & 40K & {\bf5.9} [3.7, 14]  & 4 \\
    \end{tabular}
  \label{tab:results-summary}
\end{table}

In Table \ref{tab:results-summary} we give a summary of the main results from all experiments where an objective comparison with NN-based methods was possible, including those in \textcolor{black}{the Appendices}. Instead, for the Lotka-Volterra model we did not run NN-based methods and we reported comparisons with SMC-ABC both in section \ref{sec:main-lotka-volterra} and in \textcolor{black}{Appendix \ref{sec:lotka-volterra}}. Generally, SeMPLE returns accurate inference, either by producing the smallest metric value (the smaller the better), as for Two Moons and Hyperboloid, or being second best after SNL, as for Bernoulli GLM and Ornstein-Uhlenbeck, also considering the larger variation of the SNPE-C results. However, even when SNL produces a smaller Wasserstein distance, it has to be balanced with respect to the running time. For Bernoulli GLM, SNL is between 18 and 75 times slower than SeMPLE, depending on the number of rounds, as  SNL has to perform an expensive  neural network training at each round. For the same reason, when looking at Bernoulli GLM, SNPE-C is between 8 and 26 times slower than SeMPLE, again depending on how many times it needs to be retrained. When considering the Ornstein-Uhlenbeck model, the Wasserstein distances have a good degree of overlap between SNL and SeMPLE, but SNL is between 9.5 and 10.5 times slower than SeMPLE, while SNPE-C is between 17 and 22 times slower than SeMPLE. 
In  Table \ref{tab:results-summary} we highlight whether an experiment has a unimodal or a multimodal target, as this determines the type of metric we report. Namely we use the C2ST metric for multimodal targets and the Wasserstein distance for unimodal targets. The reasons for this are now discussed, however notice that in the Appendices  we report results for both metrics in each experiment. In our experiments, we found the C2ST metric (advocated in \citealp{sbibm}) useful to evaluate the inference quality with multimodal targets, when draws from a reference posterior are available. However, we also found that at times C2ST produced contradicting results with unimodal targets. An example is the Bernoulli GLM model considered in \textcolor{black}{Appendix \ref{sec:bernoulli}}. There, it is obvious that the marginal posteriors returned by SeMPLE are accurate and that inference improved between the first two rounds, however C2ST showed that inference was apparently deteriorating for SeMPLE from round $r=1$ to $r=2$, which was not the case in reality. We also computed Wasserstein distances, which could confirm that the inference quality via SeMPLE did improve from round 1 to round 2. However, for multimodal targets, it is true that Wasserstein distances may not be a suitable tool, see {\it e.g.} \cite{balaji2019normalized}. We therefore warn  the reader from blindly relying on a single performance metric. It is also important to examine  marginal and joint posterior densities. For example, in \cite{bischoff2024practical} it is mentioned that interpreting results for C2ST requires knowledge of the
classifier used and its appropriateness for the data at hand, and that its effectiveness is critically dependent on the
selection and training of a suitable classifier.

\section{Conclusion and perspectives}\label{sec:conclusions}

We have addressed the challenging question of Bayesian inference  in the absence of explicit likelihoods. We have investigated mixture models through the GLLiM method, and shown that their expressivity could account for multivariate distributions, while their much simpler structure made them more amenable to efficient learning than NN solutions. The resulting new 
 inference strategy, named SeMPLE, yielded inference that was on par with inference from the state-of-art neural-network (NN)-based methods that we examined (SNL and SNPE-C), while being more energy and memory efficient. For the considered examples, \textcolor{black}{and the considered simulation setup,} SeMPLE produced inference of superior or similar quality than NN-based methods, and when it did 
  occasionally under-perform, it was by a negligible amount that can largely be compensated by the much smaller runtime.
  In fact, in all tested cases, SeMPLE displayed 
   much lower running times and memory requirements.
   The energy gains appear to increase with the complexity of the model simulator, and we expect to obtain even larger gains with real-life simulators. For example, for the Ornstein-Uhlenbeck model, SeMPLE achieved a 17-fold runtime speedup compared to SNPE-C. 
Although very preliminary, we hope with this attempt to meet the constraints of ``Green AI''  \citep{Schwartz2020}, and open the way to the development of more methods that can achieve accurate posterior inference, balancing the environmental impact of growing energy costs  with the obtained results quality. 

 Of course, SNL and SNPE-C could certainly  be improved for each specific case study, but we did not examine or fine-tune SNL and SNPE-C on the many other possible configurations, as this can be influenced by many factors. 
 While examining several  case studies implemented in \texttt{SBIBM}, we restricted to simulation settings provided in the \texttt{SBIBM} package, including reusing  the same NN architectures and training setups, see \textcolor{black}{Appendix \ref{sec:sbibm}} for details.

Some open questions, that can be addressed in future research, follow. SeMPLE can benefit from the summarization of data when the dimension of $\by$  is large, just like other SBI methods. 
In addition to reducing the dimension, using summary statistics (as we did with the Lotka-Volterra model) can also improve the GLLiM performance. 
Indeed the performance of SeMPLE depends on the availability of good  first approximations of the posterior and likelihood, which themselves depend on the ability of GLLiM to find probabilistic mappings between  parameters and observations, in a regression-like manner using a Gaussian mixture structure. 
The construction of informative summary statistics is an important, but independent, research question that affects not only standard ABC methods, but also SBI methods based of neural networks, {\it e.g.} \cite{chen2021neural,radev2023bayesflow}. Finally, following the discussion in Section \ref{sec:results-summary}, we wish to emphasize that measuring the quality of posterior inference in SBI is certainly an important question deserving further research, and that multiple metrics should be considered when evaluating experiments. For a recent study on metrics to evaluate the quality of SBI, see \cite{bischoff2024practical} .

\subsubsection*{Acknowledgements}
 We thank Emeline Perthame (Institut Pasteur) for assistance with the xLLiM package. HH and UP acknowledge funding from the Swedish National Research Council (Vetenskapsrådet 2019-03924). UP acknowledges funding from the Chalmers AI Research Centre. 

\bibliography{biblio}
\bibliographystyle{tmlr}

\appendix

\section*{Appendix}

  \begin{enumerate}[label=\Alph*,leftmargin=*,labelsep=2ex,ref=\arabic*]
    \item SBIBM\dotfill \pageref{sec:sbibm}
    \item BIC computation for selecting $K$ \dotfill \pageref{sec:bic}
    \item SNPE-A \dotfill\pageref{sec:snpeA}
    \item SNL \dotfill\pageref{sec:snl}
    \item Two Moons model
      \begin{enumerate}[label*=.\arabic*,leftmargin=*,labelsep=2ex]
        \item Two Moons model definition \dotfill\pageref{sec:twomoons-model}
        \item Two moons: targeting the posterior directly vs targeting the likelihood\dotfill\pageref{sec:twomoons-post-vs-lik}
        \item Two moons: selection of $K$ via BIC\dotfill\pageref{sec:twomoons-bic}
        \item Two moons: Results with $R=4$\dotfill\pageref{sec:twomoons-postplots}
        \end{enumerate}
    \item Multiple hyperboloid model\dotfill\pageref{sec:hyperboloid}
    \begin{enumerate}[label*=.\arabic*,leftmargin=*,labelsep=2ex]
    \item Multiple hyperboloid model definition\dotfill\pageref{sec:hyperboloid-definition}
    \item Multiple hyperboloid model: selection of $K$ via BIC\dotfill\pageref{sec:hyperboloid-bic}
    \item Multiple hyperboloid model:  results with $R=4$\dotfill\pageref{sec:hyperboloid-posteriors}
    \end{enumerate}
    \item Bernoulli GLM model\dotfill\pageref{sec:bernoulli}
    \begin{enumerate}[label*=.\arabic*,leftmargin=*,labelsep=2ex]
    \item Bernoulli GLM model definition\dotfill\pageref{sec:bernoulli-model}
    \item Inference setup\dotfill\pageref{sec:bernoulli-setup}
    \item Results when using $R = 2$ for all algorithms\dotfill\pageref{sec:bernoulli-results_2rounds}
    \item Results when using $R=10$ for SNL/SNPE-C and $R=2$ for SeMPLE\dotfill\pageref{sec:bernoulli-results_10rounds}
    \end{enumerate}
    \item Ornstein-Uhlenbeck model\dotfill\pageref{sec:ou}
    \begin{enumerate}[label*=.\arabic*,leftmargin=*,labelsep=2ex]
    \item Inference setup\dotfill\pageref{sec:ou-inference}
    \item Results\dotfill\pageref{sec:ou-results}
    \begin{enumerate}[label*=.\arabic*,leftmargin=*,labelsep=2ex]
        \item Results when using $R=4$ for all algorithms\dotfill\pageref{sec:ou_R=4}
        \item Results when using $R=10$ for SNL/SNPE-C and $R = 4$ for SeMPLE\dotfill\pageref{sec:ou_R=4_and_R=10}
    \end{enumerate}
    \end{enumerate}
    \item Lotka-Volterra model (Markov jump process with measurement error) \dotfill\pageref{sec:lotka-volterra}
     \begin{enumerate}[label*=.\arabic*,leftmargin=*,labelsep=2ex]
         \item Model definition with three reactions \dotfill \pageref{sec:lotka-volterra-model}
         \item Inference setup\dotfill \pageref{sec:lotka-volterra-setup}
         \item Results \dotfill \pageref{sec:lotka-volterra-results}
     \end{enumerate}
    \item Twisted prior model with 20 parameters\dotfill\pageref{sec:twisted}
     \begin{enumerate}[label*=.\arabic*,leftmargin=*,labelsep=2ex]
     \item Model definition\dotfill\pageref{sec:twisted-model}
     \item Results\dotfill\pageref{sec:twisted-results}
     \end{enumerate}
     \item Biological model of the translation kinetics after mRNA transfection \dotfill\pageref{sec:mRNA}
     \begin{enumerate}[label*=.\arabic*,leftmargin=*,labelsep=2ex]
         \item Model definition\dotfill \pageref{sec:mrna-model}
         \item Results \dotfill \pageref{sec:mrna-results}
     \end{enumerate}
\end{enumerate}

\normalsize

\section{SBIBM}\label{sec:sbibm}

Our runs of SNL \citep{papamakarios2019} and SNPE-C \citep{greenberg2019} used the implementation given in the Python package \texttt{SBIBM} \citep{sbibm}, which we ran with default parameters. However, we forked the \texttt{SBIBM} GitHub repository and conveniently modified this to output  posterior samples at each round of the inference, instead of only from the last round as given in the default implementation. This made it easier to display the performance of SNL and SNPE-C at intermediate rounds. Additionally, we added time measurements to the SNL and SNPE-C algorithms in \texttt{SBIBM} for comparison of runtimes. Note that the multiple hyperboloid model and the Ornstein-Uhlenbeck process had to be implemented as new tasks in \texttt{SBIBM} to run these models with SNL and SNPE-C. The implementation of these tasks can also be found in the \texttt{SBIBM} fork mentioned above.

The default setting for SNPE-C in \texttt{SBIBM} is to utilize Neural Spline Flows (NSF) as density estimator with 50 hidden features and 10 atoms. Correspondingly, SNL by default uses Masked Autoregressive Flow for density estimation with 50 hidden features. The MCMC step of SNL utilizes vectorized slice sampling with 100 Markov chains and a thinning interval of 10 by default.

\section{BIC computation for selecting $K$}\label{sec:bic}

In GLLiM, the number $K$ of components in the Gaussian mixture model  has to be specified prior to performing the EM procedure. A too large value of $K$ may result in overfitting and unnecessary computational effort, while a too small value of $K$ may limit the ability to represent the relationship between $\btheta$ and $\by$. The Bayesian Information Criterion (BIC), used in \cite{deleforge} to guide the selection of  $K$, is given in equation \eqref{eq:bic}  
\begin{equation}
    \label{eq:bic}
    BIC = -2 \mathcal{L}(\hat{\bphi}) + D(\Tilde{\bphi}) \log N
\end{equation}
where $\Tilde{\bphi}$ is the GLLiM model parameter,  $\mathcal{L}(\hat\bphi)$ is the maximised value of the  GLLiM log-likelihood function at the MLE $\hat\bphi$, $D(\Tilde{\bphi})$ is the total number of parameters in the model and $N$ is the number of observations in the training dataset. We wish to select a $K$ returning a small BIC when GLLiM is fitted with $K$ components to a training dataset 
$\{\btheta_n, \by_n\}_{n=1}^{N}$ obtained by sampling parameters $\btheta_n \sim p(\btheta)$ from the prior and simulating the corresponding $\by_n \sim p(\by|\btheta_n)$ from the generative model.
The GLLiM log-likelihood is specified as (see eq. (4) in the main body),
\begin{equation}
    \mathcal{L(\Tilde{\bphi})} = \sum_{n=1}^{N} \log q_{\tilde{\bphi}}(\by_n, \btheta_n),
\end{equation}
with
\begin{equation}
    q_{\tilde{\bphi}}(\by_n, \btheta_n) = \sum_{k=1}^{K} 
    \mathcal{N}(\by_n; \tilde{\boldsymbol A}_k \btheta + \tilde{\boldsymbol b}_k, \tilde{\boldsymbol \Sigma}_k)\;  \mathcal{N}(\btheta_n; \tilde{\boldsymbol c}_k, \tilde{\boldsymbol \Gamma}_k)  \;
    \pi_k \; .
\end{equation}
The number of parameters that GLLiM needs to estimate is
\begin{equation}
    D(\boldsymbol \phi) = (K-1) + K(DL + D + L + \operatorname{nbpar}_\Sigma + \operatorname{nbpar}_\Gamma),
\end{equation}
where $\operatorname{nbpar}_\Sigma$ and $\operatorname{nbpar}_\Gamma$ are the number of parameters in the covariance matrices $\tilde{\Sigma}_k$ and $\tilde{\Gamma}_k$, respectively. The covariance structure of the matrices $\tilde{\Sigma}_k$ and $\tilde{\Gamma}_k$ can be constrained to reduce the number of parameters that GLLiM needs to estimate. One possible constraint is to set the covariance matrices to be isotropic, which means that they are set to be proportional to the identity matrix, which is the default in the \texttt{xLLiM} package \citep{xllim}. 

The GLLiM implementation in the \texttt{xLLiM} package removes a mixture component $k$ if the mixture probability 
becomes zero. Empirically it was found that when mixture probabilities becomes very small, GLLiM could occasionally crash with errors about failed internal matrix inversions. To prevent this, a threshold was set within SeMPLE to remove mixture components with probabilities $\tilde{\eta}_k(\btheta)$ below this threshold, which we set to be 0.005 for the multiple hyperboloid, Ornstein-Uhlenbeck, Lotka-Volterra and the biological SDE model, zero for Two Moons (hence no component was deleted) or 0.03 for Bernoulli GLM and SLCP models. This check was repeated in every \textcolor{black}{round} of SeMPLE. This shows the need for setting  reasonable initial number of mixture components $K$, as setting a too large value inflates the problem of small mixture component probabilities. However, in several cases our inference tasks worked out-of-the-box by setting the initial value to $K=20$ or 30.

\section{SNPE-A}\label{sec:snpeA}

In \cite{papamakarios2016} a parametric approximation to the exact posterior is learned via sequential updates, and the method is denoted SNPE-A (that is, Sequential Neural Posterior Approximation A\footnote{This is not the name given in \cite{papamakarios2016} but follows the taxonomy established in \cite{durkan2020contrastive}.}). This approximation is acquired by training a mixture density network (MDN, \citealp{bishop}), which is a Gaussian mixture where the means and covariances are parameterised via neural networks. However, no method of estimating the number of mixture components is provided in SNPE-A.
SNPE-A is given in Algorithm \ref{alg:snpe-a_train_prop_prior} and \ref{alg:snpe-a_train_posterior}. The MDN with parameters $\bphi$ is denoted by $q_{\bphi}(\btheta\g\by)$ and the proposal distribution for parameters $\btheta$ is denoted $\tilde{p}(\btheta)$ and called ``proposal prior''. Therefore, parameters $\btheta_n\sim \tilde{p}(\btheta)$ are drawn and corresponding data $\by_n$ are simulated from the generative model $p(\by\g\btheta_n)$, and the MDN $q_{\bphi}(\btheta\g\by)$ is trained on the data set $\{\btheta_n,\by_n\}_{n=1}^{N}$ to obtain a posterior approximation.

\cite{papamakarios2016} show that using preliminary posterior fits to guide future simulations drastically reduces the number of simulations required to learn an accurate posterior approximation. In terms of notation, this means that parameters are proposed from $\tilde{p}(\btheta)$, which can be different from the prior $p(\btheta)$. However, not sampling from the prior distribution requires a correction factor to target the true posterior distribution. The motivation behind this is given in Proposition 1 in \cite{papamakarios2016}, and ultimately they suggest to estimate the posterior by
\begin{equation*}
    \hat{p}(\btheta\g\by_o) \propto \frac{p(\btheta)}{\tilde{p}(\btheta)} q_{\bphi}(\btheta\g \by_o).
\end{equation*}
As mentioned in \cite{papamakarios2016}, in Algorithm \ref{alg:snpe-a_train_prop_prior} as long as $q_\phi (\btheta|\by)$ has only one Gaussian component ($K = 1$) then $\tilde{p}(\btheta)$
remains a single Gaussian throughout. This Gaussian approximation can be used as a rough but cheap approximation to the true posterior,
or it can serve as a good proposal prior in Algorithm \ref{alg:snpe-a_train_posterior} for fine-tuning a non-Gaussian
multi-component posterior. When the second strategy is used, {\it i.e.} Algorithm \ref{alg:snpe-a_train_posterior} is employed, they reuse the single-component
neural density estimator learnt in Algorithm \ref{alg:snpe-a_train_prop_prior} to initialize $q_\phi$ in Algorithm \ref{alg:snpe-a_train_posterior}, and in this case the weights in the final
layer of the MDN are replicated $K$ times, with small random perturbations to break symmetry.
Notice,
in SNPE-A the proposal prior $\tilde{p}(\btheta)$ trained in Algorithm \ref{alg:snpe-a_train_prop_prior} is restricted to be a Gaussian distribution, and the prior $p(\btheta)$ is assumed to be a uniform or Gaussian distribution to allow analytical calculation of $\tilde{p}(\btheta\g\by = \by_o)$. 

In SeMPLE we do not have such restrictions, meaning that  we can consider any prior $p(\btheta)$, however this comes at the price of introducing MCMC sampling (as in SNL). However, unlike for SNPE-A, in SeMPLE we are not constrained to reuse the estimated one-component Gaussian and replicate this $K$ times to obtain a $K$-components mixture, instead the GLLiM algorithm automatically determines the means and covariance matrices for each of the $K$ components, which as such can be different between components, thus adding flexibility to the shapes of the posterior distributions that can be targeted.

\begin{figure}[htbp]
\begin{tabular}{cc}
\begin{minipage}{0.5\textwidth}
\begin{algorithm}[H]
    \begin{algorithmic}[]
    \State Initialize $q_{\bm{\phi}}\br{\bm{\theta}\g\by}$ with one component ($K=1$).
    \State $\tilde{p}\br{\bm{\theta}} \leftarrow \prob{\bm{\theta}}$.
    \Repeat
    \For{$n=1:N$}
    \State sample $\bm{\theta}_n \sim \tilde{p}\br{\bm{\theta}}$
    \State sample $\by_n \sim \prob{\by\g \bm{\theta}_n}$
    \EndFor
    \State retrain $q_{\bm{\phi}}\br{\bm{\theta}\g\by}$ on $\set{\bm{\theta}_n, \by_n}_{n=1}^{N}$
    \State $\tilde{p}\br{\bm{\theta}} \leftarrow \frac{\prob{\bm{\theta}}}{\tilde{p}\br{\bm{\theta}}}\,q_{\bm{\phi}}\br{\bm{\theta}\g\by_o}$
    \Until{$\tilde{p}\br{\bm{\theta}}$ has converged}
    \caption{SNPE-A: Training of proposal prior}
    \label{alg:snpe-a_train_prop_prior}
    \end{algorithmic}
\end{algorithm}
\end{minipage}
&
\begin{minipage}{0.5\textwidth}
\begin{algorithm}[H]
\begin{algorithmic}
    \State Initialize $q_{\bm{\phi}}\br{\bm{\theta}\g\vect{x}}$ with $K$ components. \\
    \{if $q_{\bm{\phi}}$ available by Algorithm~\ref*{alg:snpe-a_train_prop_prior}\\
    initialize by replicating its\\
    one component $K$ times\}
    \For{$n=1:N$}
    \State sample $\bm{\theta}_n \sim \tilde{p}\br{\bm{\theta}}$
    \State sample $\by_n \sim \prob{\by\g \bm{\theta}_n}$
    \EndFor
    \State train $q_{\bm{\phi}}\br{\bm{\theta}\g\by}$ on $\set{\bm{\theta}_n, \by_n}_{n=1}^{N}$
    \State $\hatprob{\bm{\theta}\g \by=\by_o} \leftarrow \frac{\prob{\bm{\theta}}}{\tilde{p}\br{\bm{\theta}}}\,q_{\bm{\phi}}\br{\bm{\theta}\g\by_o}$
    \caption{SNPE-A: Training of posterior}
    \label{alg:snpe-a_train_posterior}
\end{algorithmic}
\end{algorithm}
\end{minipage}
\end{tabular}
\vspace*{-0.2cm}
\end{figure}

\section{SNL}\label{sec:snl}

Sequential Neural Likelihood (SNL) \citep{papamakarios2019}  sequentially trains a conditional neural density estimator for the likelihood function, instead of training an estimator for the posterior as in SNPE-A.  This eliminates the need for corrections stemming from the proposal distribution not being the prior, but introduces an additional MCMC step (in the form of slice sampling) to sample from the posterior distribution, by employing the estimation  $q_{\bphi}(\by_o \g\btheta)$ of the likelihood. SNL is detailed in Algorithm \ref{alg:snl}, which is the same as Algorithm 1 in \cite{papamakarios2019}. The main difference with SeMPLE is that, while SNL trains $q_\phi(\by|\btheta)$ and then plugs the latter into MCMC for posterior sampling, SeMPLE trains $q_{\tilde{\phi}}(\by|\btheta)$ and also obtain (as an \textit{immediate} algebraic by-product of the  $q_{\tilde{\phi}}(\by|\btheta)$ training) a proposal distribution $q_\phi(\btheta|\by_o)$ that is used in the MCMC step.

\begin{algorithm}[H]
    \caption{Sequential Neural Likelihood (SNL)}
    \label{alg:snl}
\begin{algorithmic}
    \State \textbf{Input:} observed data $\by_o$, estimator $q_{\phi}(\by\g \btheta)$, number of rounds $R$, simulations per round $N$.
    \State \textbf{Output:} approximate posterior $\hat{p}(\btheta\g \by_0)$.
    \\
    \State Set $\hat{p}_0(\btheta\g\by_o) = p(\boldsymbol\theta)$ and $\mathcal{D} = \{\}$ 
    \For{$r=1:R$}
    \For{$n=1:N$}
    \State sample $\boldsymbol\theta_n \sim \hat{p}_{r-1}(\boldsymbol\theta\g\by_o)$ with MCMC
    \State simulate $\by_n \sim p(\by \g \boldsymbol\theta_n)$
    \State add $(\boldsymbol\theta_n,\by_n)$ into $\mathcal{D}$ 
    \EndFor
    \State {(re-)train} $q_{\bphi}(\by \g\btheta)$ on $\mathcal{D}$ and set $\hat{p}_r(\btheta \g\by_o) \propto q_{\bphi}(\by_o \g\btheta) \, p(\boldsymbol\theta)$
    \EndFor
    \State \Return $\hat{p}_R(\btheta \g \by_o)$
\end{algorithmic}
\end{algorithm}

\clearpage

\section{Two moons model}\label{sec:twomoons}

\subsection{Two moons model definition}\label{sec:twomoons-model}

The Two Moons model, described in \cite{greenberg2019}, generates $\by \in \mathbb{R}^2$ for a given parameter $\boldsymbol \theta \in \mathbb{R}^2$ according to
\begin{align}
a &\sim \mathcal{U}(-\frac{\pi}{2}, \frac{\pi}{2}) \\
r &\sim \mathcal{N}(0.1, 0.01^2) \\
\boldsymbol{p}&= (r \cos(a) + 0.25, \ r \sin(a)) \\
 \by^\top &= \boldsymbol{p} + \left(-\frac{|\theta_1+\theta_2|}{\sqrt{2}}, \ \frac{-\theta_1+\theta_2}{\sqrt{2}} \right).
\end{align}
Same as in \cite{greenberg2019}, we set uniform priors $\theta_1\sim\mathcal{U}(-1, 1)$, $\theta_2\sim\mathcal{U}(-1, 1)$ . By construction, this toy example has a posterior distribution with two crescent shapes for a given observation.

\subsection{Two moons: targeting the posterior directly vs targeting the likelihood}\label{sec:twomoons-post-vs-lik}

\begin{figure}[h] 
    \centering
    \includegraphics[width=0.4\textwidth]{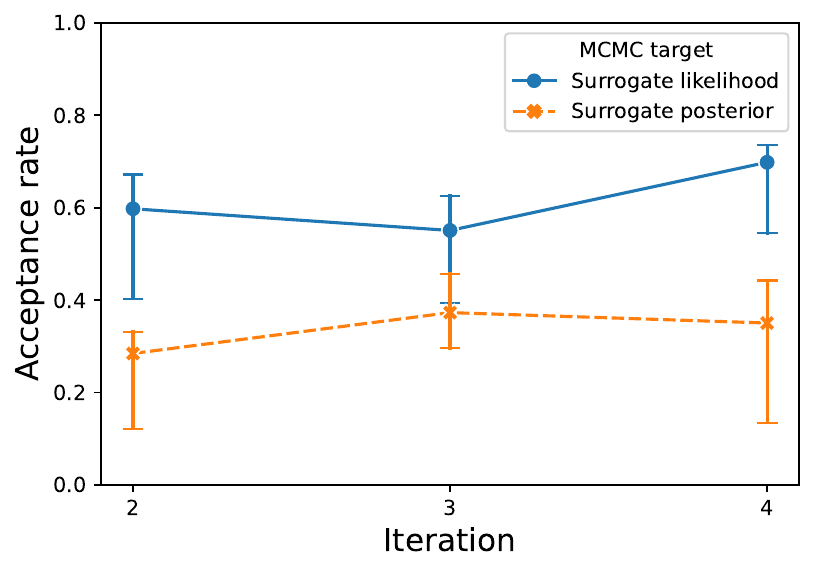}
    \caption{Two Moons: Median acceptance rate of the Metropolis-Hastings step of 10 SeMPLE runs with different data sets. Error bars show min/max values.}
    \label{fig:two_moons_accrate}
\end{figure}

\begin{figure}[h]
    \captionsetup[subfigure]{justification=centering}
     \centering
     \begin{subfigure}[b]{0.35\textwidth}
         \centering
         \includegraphics[width=\textwidth]{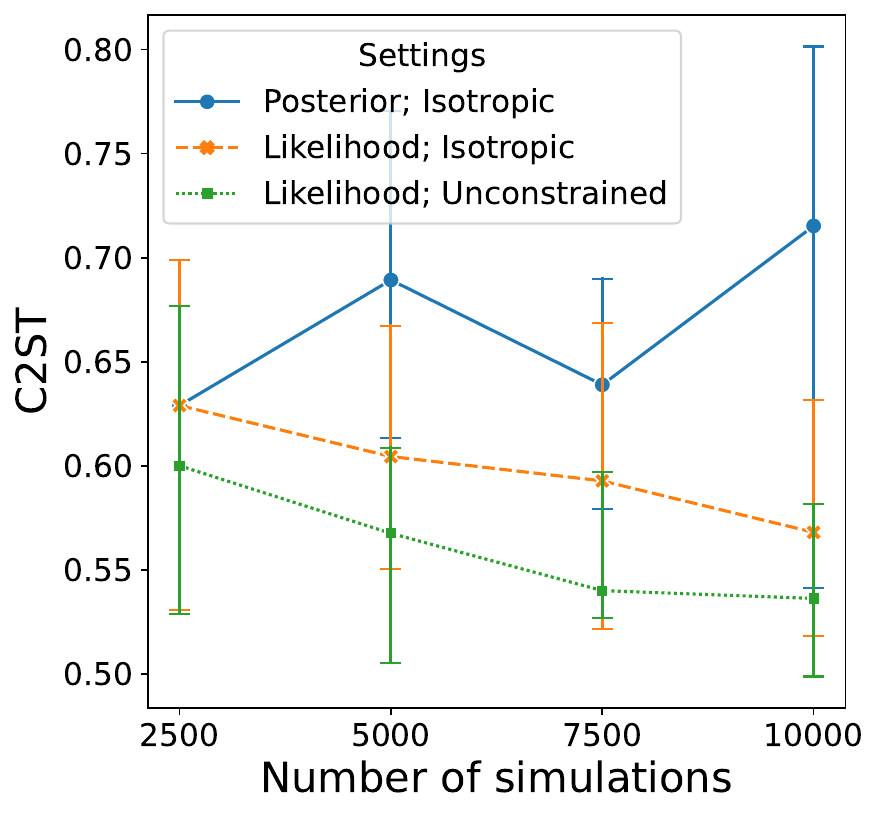}
         \caption{}
         \label{fig:TM_settings_c2st_vs_sims}
     \end{subfigure}
     \begin{subfigure}[b]{0.35\textwidth}
         \centering
         \includegraphics[width=\textwidth]{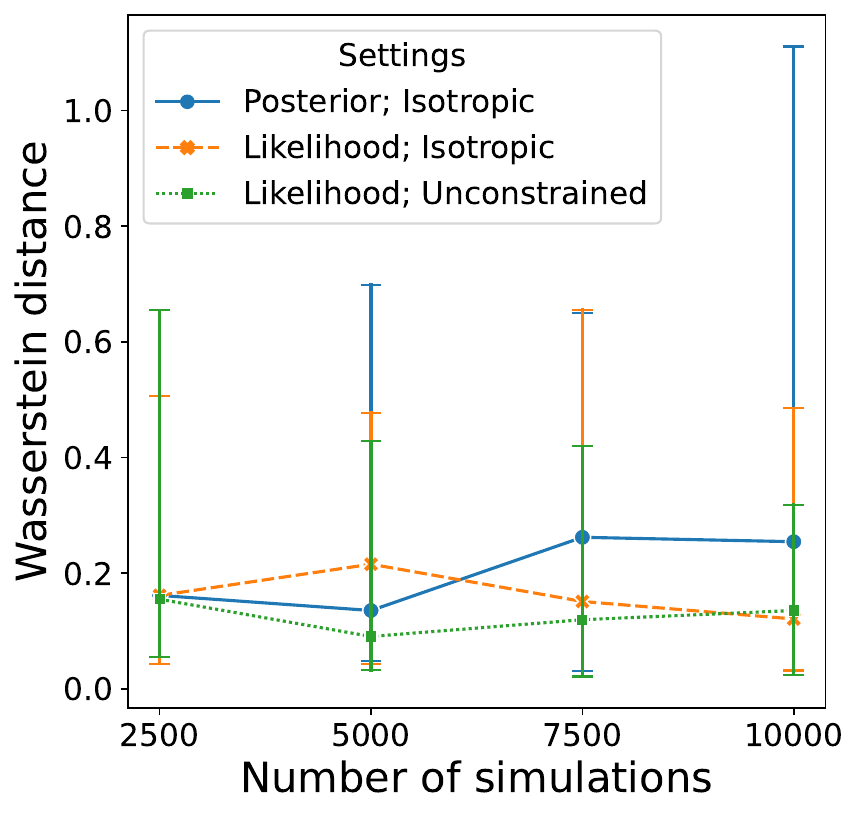}
         \caption{}
         \label{fig:TM_settings_emdp2_vs_sims}
     \end{subfigure}
     \caption{Two Moons: Median C2ST and Wasserstein distance on 10 SeMPLE runs with different data sets vs number of model simulations. The three different SeMPLE settings refers to whether the surrogate likelihood (reported as ``likelihood; isotropic'' or ``likelihood; unconstrained'') or the ``corrected'' surrogate posterior (reported as ``Posterior; isotropic'') is used in the MCMC target distribution, and constraints  used for the GLLiM covariance matrices (``isotropic'' or ``unconstrained''). Error bars show min/max values.}
     \label{fig_supp:TM_settings_metric_vs_sims}
\end{figure}

In the main paper we discussed the two possibilities of targeting the posterior $\tilde{p}_{r+1}(\btheta) \propto q_{\tilde{\bphi}_r}(\by_o \g \btheta) p(\btheta)$ by first obtaining an estimate $q_{\tilde{\bphi}_r}(\by_o \g \btheta)$ of the likelihood  and then sampling via MCMC from $\tilde{p}_{r+1}(\btheta)$, as opposed to the option of using the GLLiM surrogate posterior $q_{\bphi_r}(\btheta\g \by_o)$ to update the posterior $\tilde{p}_{r+1}(\btheta) = \frac{p(\btheta)}{\tilde{p}_{r}(\boldsymbol\theta)} q_{\bphi_r}(\btheta\g \by_o)$, and then sampling from the latter via MCMC. We hinted at the fact that it is preferable to follow the first route, which we used to obtain all our results, and here we provide results in support of our recommendation. 
Figure \ref{fig_supp:TM_settings_metric_vs_sims} shows the C2ST metric and the Wasserstein distance as a function of the number of model simulations for different SeMPLE settings, including different choices for the covariance matrices  $\tilde{\bm{\Sigma}}_k$ (``isotropic'' or ``full unconstrained'') of the mixture models. We first considered the ``isotropic'' covariances case (specified in \texttt{xLLiM} via the option \verb|cstr$Sigma="i"|), which means that all the $K$ covariance matrices are proportional to the diagonal matrix, however these matrices are all different between the $K$ components. With this option, we ran SeMPLE by targeting $\tilde{p}_{r+1}(\btheta) \propto q_{\tilde{\bphi}_r}(\by_o \g \btheta) p(\btheta)$ (``likelihood; isotropic'' in the legend of Figure \ref{fig_supp:TM_settings_metric_vs_sims}) and also ran SeMPLE by targeting $\tilde{p}_{r+1}(\btheta) = \frac{p(\btheta)}{\tilde{p}_{r}(\boldsymbol\theta)} q_{\bphi_r}(\btheta\g \by_o)$ (``posterior; isotropic'' in the legend of Figure \ref{fig_supp:TM_settings_metric_vs_sims}). Targeting $\tilde{p}_{r+1}(\btheta) \propto q_{\tilde{\bphi}_r}(\by_o \g \btheta) p(\btheta)$ is clearly the preferred route according to C2ST. This is less clear by looking at the Wasserstein distance, however the computation of this particular metric may not be accurate with multimodal targets (while we recommend to use Wasserstein distances for unimodal targets), and we refer the reader to section \ref{sec:results-summary} in the main paper for a discussion about metrics to evaluate posterior inference.
Additionally, we explored the results of assuming full matrices $\tilde{\bm{\Sigma}}_k$ without constraints (specified in \texttt{xLLiM} via the option \verb|cstr$Sigma=" "|). This setting is run together with our (recommended) method of sampling from $\tilde{p}_{r+1}(\btheta) \propto q_{\tilde{\bphi}_r}(\by_o \g \btheta) p(\btheta)$ (``likelihood; unconstrained'' in the legend of Figure \ref{fig_supp:TM_settings_metric_vs_sims}). This SeMPLE configuration produces the best results. This may not be surprising, as for two-moons $\by_0$ has dimension 2, therefore the covariances $\tilde{\bm{\Sigma}}_k$ have dimensions $2\times 2$, thus not inducing a large number of parameters $\tilde{\bphi}$ to estimate via EM. Notice that the C2ST value, when using our recommended method but with isotropic covariance matrices, is still lower than C2ST as obtained with SNPE-C (see the main paper). Regarding the performance of the MCMC algorithm, Figure \ref{fig:two_moons_accrate} shows that the acceptance rate is noticeably higher with the ``likelihood'' option rather than the ``posterior'' one, which strengthens our choice. While a higher acceptance rate does not denote, per-se, any special ability for an algorithm to efficiently explore the posterior, however this fact coupled with the already displayed high quality inference provided by SeMPLE for this example, means that Figure \ref{fig:two_moons_accrate} further shows that we can obtain a larger number of representative posterior samples when targeting $\tilde{p}_{r+1}(\btheta) \propto q_{\tilde{\bphi}_r}(\by_o \g \btheta) p(\btheta)$.

Based on these results, it was concluded that using the surrogate likelihood to target $q_{\tilde{\phi}_r}(\by_o \g \btheta) p(\btheta)$ in the MCMC step, with the surrogate posterior $q_{\phi_r}(\btheta\g \by_o)$ as proposal distribution in the Metropolis-Hastings algorithm, is a superior method overall, which we used in all the other simulation studies.

\subsection{Two moons: selection of $K$ via BIC}\label{sec:twomoons-bic}

\begin{figure}[h]
    \centering
    \includegraphics[width=0.5\textwidth]{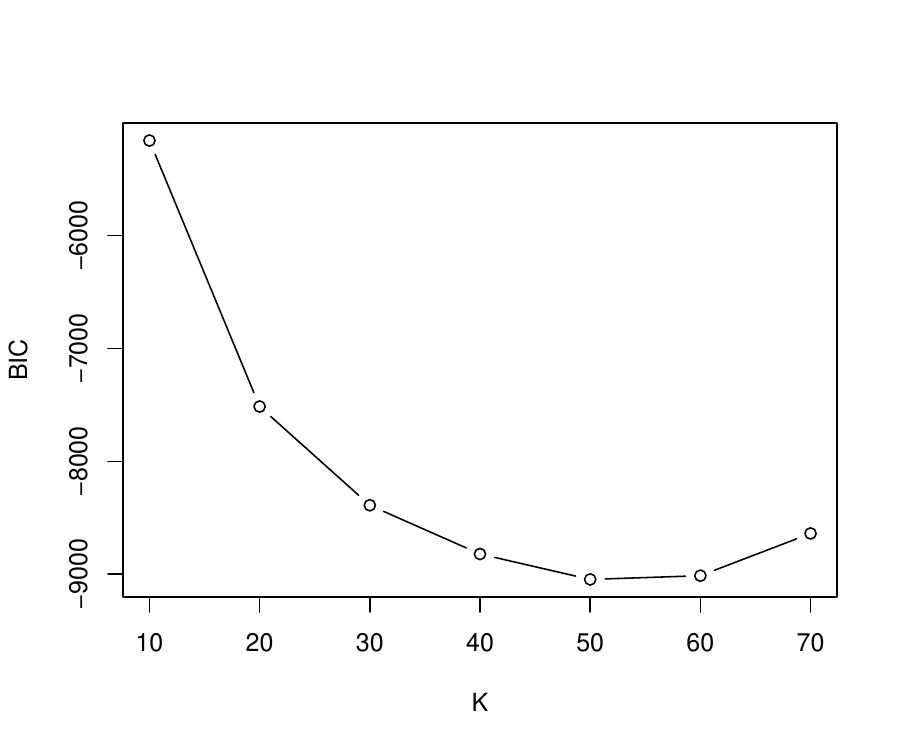}
    \caption{Two Moons. BIC with respect to the number of mixture components $K$.}
    \label{fig:bic_two_moons}
\end{figure}

We illustrate how to select a starting value for $K$ using BIC. However, we recall that it is not strictly necessary to choose $K$ in a very precise way, as its value is let decrease during the SeMPLE run. All BIC values are computed on the same prior-predictive data set $\{\boldsymbol\theta_n, \by_n\}_{n=1}^{N}$ with $N=2\,500$, obtained prior to running SeMPLE. Figure \ref{fig:bic_two_moons} shows BIC as a function of $K$. The minimum BIC of the evaluated values is obtained at $K=50$ but it decreases sharply from $K=10$ to $K=30$. By the Occam's razor principle, $K=30$ was used as input to SeMPLE. Empirical experiments showed that a larger value of $K$ did not improve results. The runtime to compute all BIC values in Figure \ref{fig:bic_two_moons} was 93 seconds (on a desktop computer with a 6-core (12 threads) AMD
Ryzen 5 2600 CPU). Note that the BIC computations are ``amortised'', that is are independent of the observed data set $\by_o$, and only have to be performed once before running SeMPLE, and can thus be recycled whenever a different observed data set is considered.

\subsection{Two moons. Results with $R=4$}\label{sec:twomoons-postplots}

In the main paper, we have considered the inference results based on a total of $10^4$ model simulations, uniformly distributed across $R=10$ rounds for SNPE-C and SNL (as per \texttt{SBIBM} defaults), and $R=4$ rounds for SeMPLE. Results showed that SeMPLE performed better and in particular that SNL (which is also an MCMC-based procedure, similarly to SeMPLE) was returning the worst performance. Here we show the performance of both SNPE-C and SNL with $R=4$, again using a total of $10^4$ model simulations for each method. That is, for each method we show the inference obtained after 2500, 5000, 7500 and 10000 model simulations.
Figures \ref{fig:two_moons_density_scatter_obs1}-\ref{fig:two_moons_density_scatter_obs10} shows scatter plots (colored by density value) of the posterior samples from the last \textcolor{black}{round} (r=4) of each algorithm and for each of the 10 observed data sets in SBIBM. SeMPLE consistently matches the reference posterior, unlike SNL and SNPE-C. 
\foreach \x in {1,2,3,4,5,6,7,8,9,10}
{
\begin{figure}[tp]
    \centering
    \includegraphics[width=\textwidth]{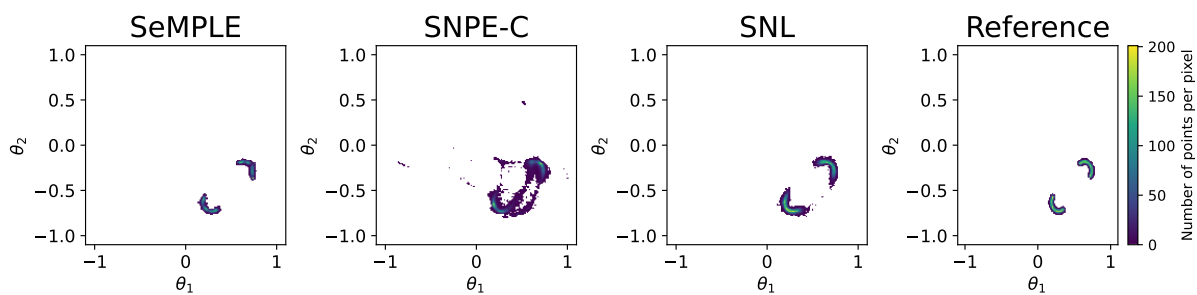}
    \caption{Two Moons using $R=4$ rounds for every method. Density scatter plot of posterior samples from the last \textcolor{black}{round} (r=4) of each algorithm. Observed data set number \x \, in SBIBM.}
    \label{fig:two_moons_density_scatter_obs\x}
\end{figure}
}

Moreover, it is even more interesting to note that SeMPLE performs already well after a single \textcolor{black}{round}, see
Figures \ref{fig:two_moons_density_scatter_all_obs1}-\ref{fig:two_moons_density_scatter_all_obs10} showing posterior samples from every \textcolor{black}{round} of each algorithm and for each of the 10 observed data sets. While SNL improves across \textcolor{black}{rounds}, this is not the case for SNPE-C.
\foreach \x in {1,2,3,4,5,6,7,8,9,10}
{
\begin{figure}[h]
    \centering
    \includegraphics[width=\textwidth]{TwoMoons/TM_density_sims_vs_alg/observation\x.pdf}
    \caption{Two Moons using $R=4$ rounds for every method: inference using observed dataset number \x \, from SBIBM. Posterior samples from each algorithm round, starting from the first round (upper row) to the fourth (bottom row).}
    \label{fig:two_moons_density_scatter_all_obs\x}
\end{figure}
}

\begin{figure}[h]
    \centering
    \includegraphics[scale=0.4]{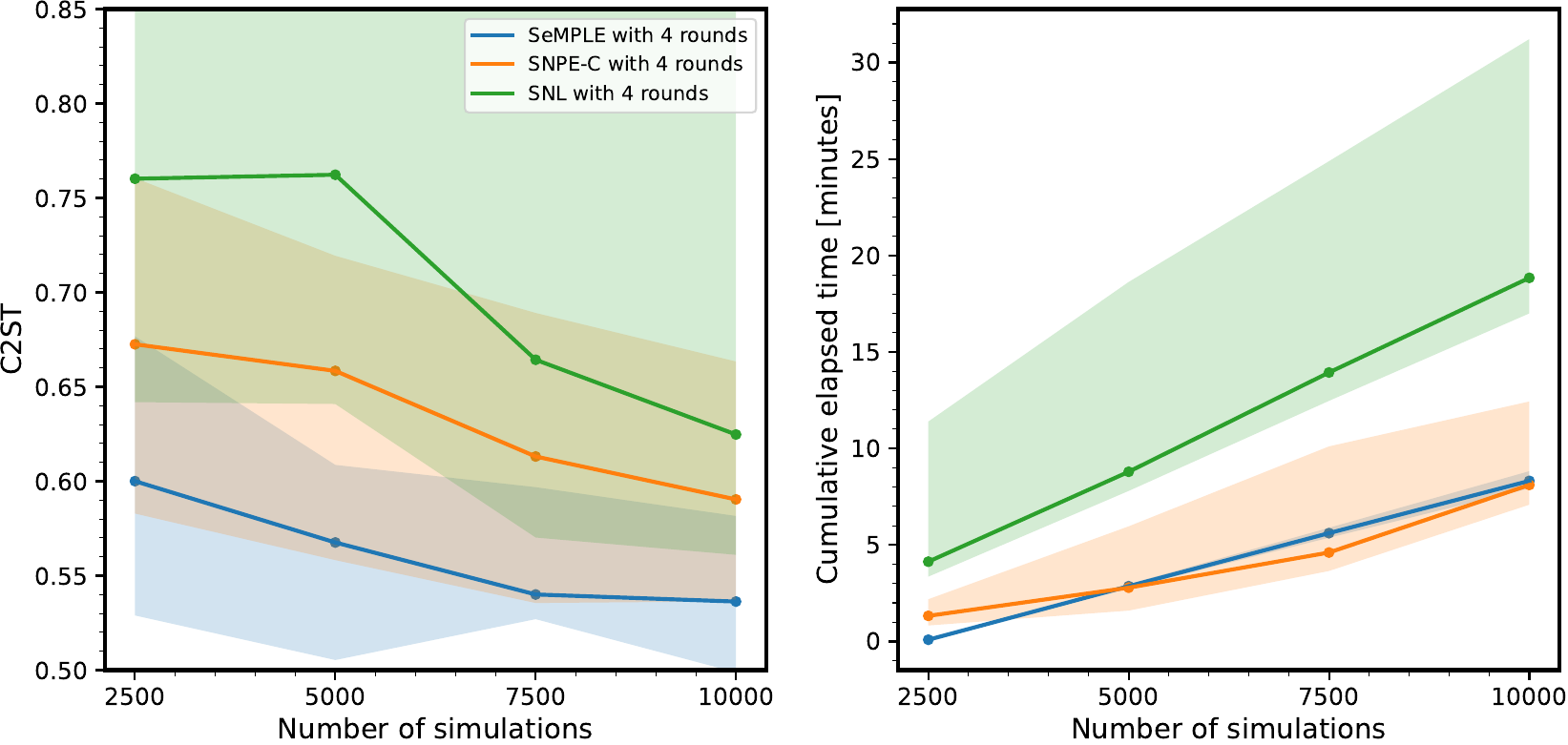}
    \caption{Two Moons using $R=4$ rounds for every method. Median C2ST (left) and median cumulative runtime in minutes (right) for 10 runs with different data sets vs the number of model simulations. Shaded bands enclose the min and max values.}
    \label{fig:TM_metrics_4rounds}
\end{figure}

Figure \ref{fig:TM_metrics_4rounds} shows C2ST and runtime values in the case when $R=4$ is used for all algorithms. Compared to the results in the main paper with $R=10$ rounds, SNPE-C and SNL perform slightly worse with $R=4$ rounds in terms of the C2ST metric. The SNPE-C and SNL runtimes are however reduced, with SNPE-C having essentially identical runtime as SeMPLE, although with significantly higher variance. SNL is approximately twice as slow as the other algorithms.

\clearpage

\section{Multiple hyperboloid model}\label{sec:hyperboloid}

\subsection{Multiple hyperboloid model definition}\label{sec:hyperboloid-definition}

This model was introduced by \cite{forbes}.
With reference to applications in audio processing, the unknown parameter $\boldsymbol \theta \in \mathbb{R}^2$ can be interpreted as the coordinates of a sound source location on the plane, while the two pairs of 2-dimensional parameters $\boldsymbol m^1 = (\boldsymbol m_1^1, \boldsymbol m_2^1)$ and $\boldsymbol m^2 = (\boldsymbol m_1^2, \boldsymbol m_2^2)$ can be interpreted as the location of two pairs of microphones aimed at detecting the sound source. An observation $\boldsymbol y \in \mathbb{R}^d$ has the likelihood function
\begin{equation}
\label{eq:hyperboloid_likelihood}
    p(\by\g\btheta) = \frac{1}{2} \mathcal{S}(\boldsymbol y; F_{\boldsymbol m^1}(\boldsymbol\theta) \mathbbm{1}_d, \sigma^2 I_d, \nu) + \frac{1}{2} \mathcal{S}(\boldsymbol y; F_{\boldsymbol m^2}(\boldsymbol\theta) \mathbbm{1}_d, \sigma^2 I_d, \nu),
\end{equation}
where
\begin{equation}
\label{eq:hyperboloid_likelihood_supp}
    F_m(\boldsymbol \theta) = (||\boldsymbol \theta - \boldsymbol m_1 ||_2 - ||\boldsymbol \theta - \boldsymbol m_2 ||_2), \, \text{with} \, \, \boldsymbol m=(\boldsymbol m_1, \boldsymbol m_2).
\end{equation}
The likelihood in equation \eqref{eq:hyperboloid_likelihood} is a mixture of two Student's t-distributions with a $d$-dimensional location parameter having all dimensions equal to $F_{\boldsymbol m^1}(\btheta)$, $F_{\boldsymbol m^2}(\btheta)$ respectively, a diagonal scale matrix $\sigma I_d$ and $\nu$ degrees of freedom. In our example, the observed data vector $\by_o$ has length $d=10$, and $\nu=3$, $\sigma=0.01$, $\boldsymbol m_1^1 = (-0.5,0)^T$, $\boldsymbol m_2^1 = (0.5,0)^T$, $\boldsymbol m_1^2 =(0,-0.5)^T$ and $\boldsymbol m_2^2=(0,0.5)^T$. The prior is set to be a uniform $\mathcal{U}(-2, 2$) for both components of $\btheta$, and $\btheta$ is the only parameter to infer. Data generating parameter is $\btheta^* = (1.5,1)$ as in \cite{forbes}, which represents the coordinates of the location of a sound source. 

Since the multiple hyperboloid model is not implemented in \texttt{SBIBM}, there exists no observed data sets or reference posterior samples in \texttt{SBIBM}. To produce an observed data set $\by_0$, the model simulator $p(\by\g\btheta^*)$ was run with the true parameter value $\btheta^*= (1.5,1)$. Since the posterior shape varies widely depending on the true parameter value, and to keep the connection to the experiment in \cite{forbes}, only a single  observed data set was produced. Instead, the metric results were averaged over 10 independent runs with this same observed data set. To produce a reference posterior sample corresponding to the produced observed data set, the Metropolis-Hastings algorithm was used. The sample size of the reference posterior sample was set to $10^4$.
 
\subsection{Multiple hyperboloid model: selection of $K$ via BIC}\label{sec:hyperboloid-bic}

\begin{figure}[ht]
    \centering
    \includegraphics[width=0.45\textwidth]{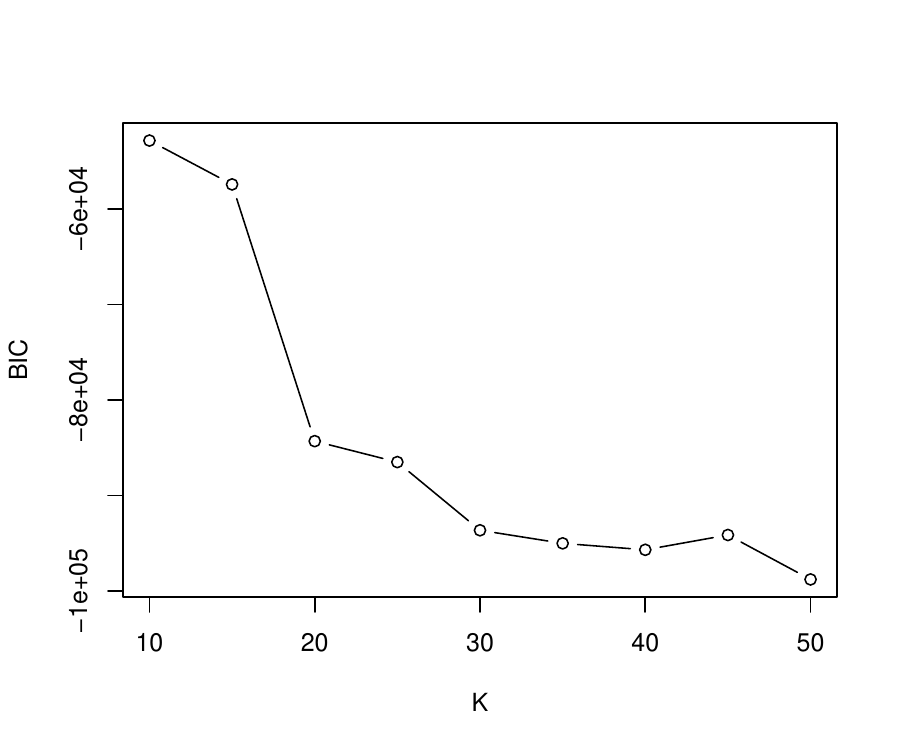}
    \caption{Multiple hyperboloid. BIC with respect to the number of Gaussian mixture components K.}
    \label{fig:bic_hyperboloid}
\end{figure}

Figure \ref{fig:bic_hyperboloid} shows the BIC for different values of K computed by fitting GLLiM, prior to running SeMPLE, using the same prior-predictive data set $\{\boldsymbol\theta_n, \by_n\}_{n=1}^{N}$ with $N=10\,000$. Isotropic covariance matrices were used in GLLiM as it was found that using fully-specified covariance matrices did not improve the results. The decrease in the BIC slows down significantly at K=30, which motivates a starting value of K=40 in SeMPLE. This is close to the value of K=38 used in \cite{forbes} (which was also selected for the same multiple hyperboloid model via BIC). The runtime to compute all the BIC values in Figure \ref{fig:bic_hyperboloid} was 270 seconds = 4.5 minutes. Note once again that the BIC computation is independent of the observed data set and has to be performed only once before running SeMPLE.

\subsection{Multiple hyperboloid model: results with $R=4$}\label{sec:hyperboloid-posteriors}

\begin{figure}[h]
    \centering
    
    \begin{subfigure}[b]{0.48\linewidth}
        \includegraphics[scale=0.45]{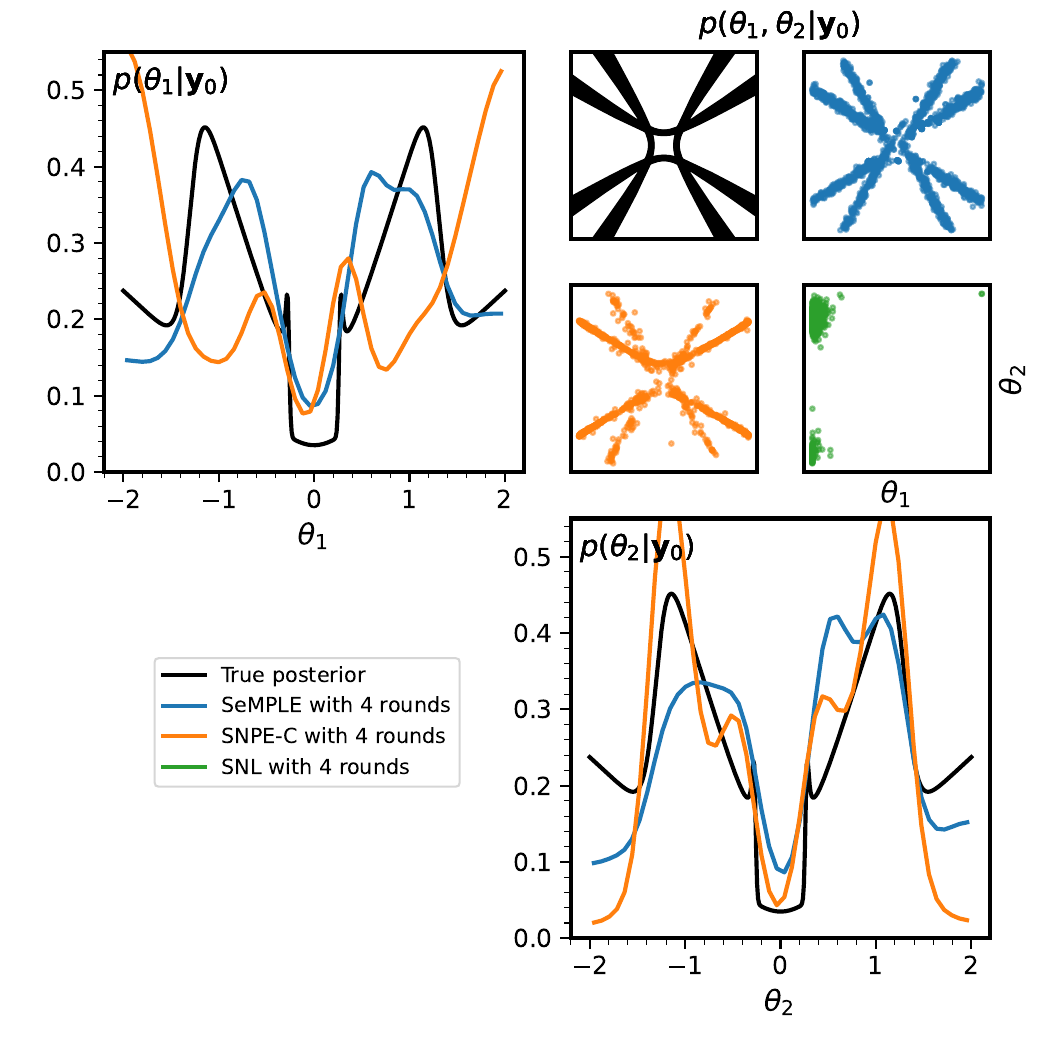}
    \caption{Repetition 1}
    \end{subfigure}
    \hfill
    \begin{subfigure}[b]{0.48\linewidth}
        \includegraphics[scale=0.45]{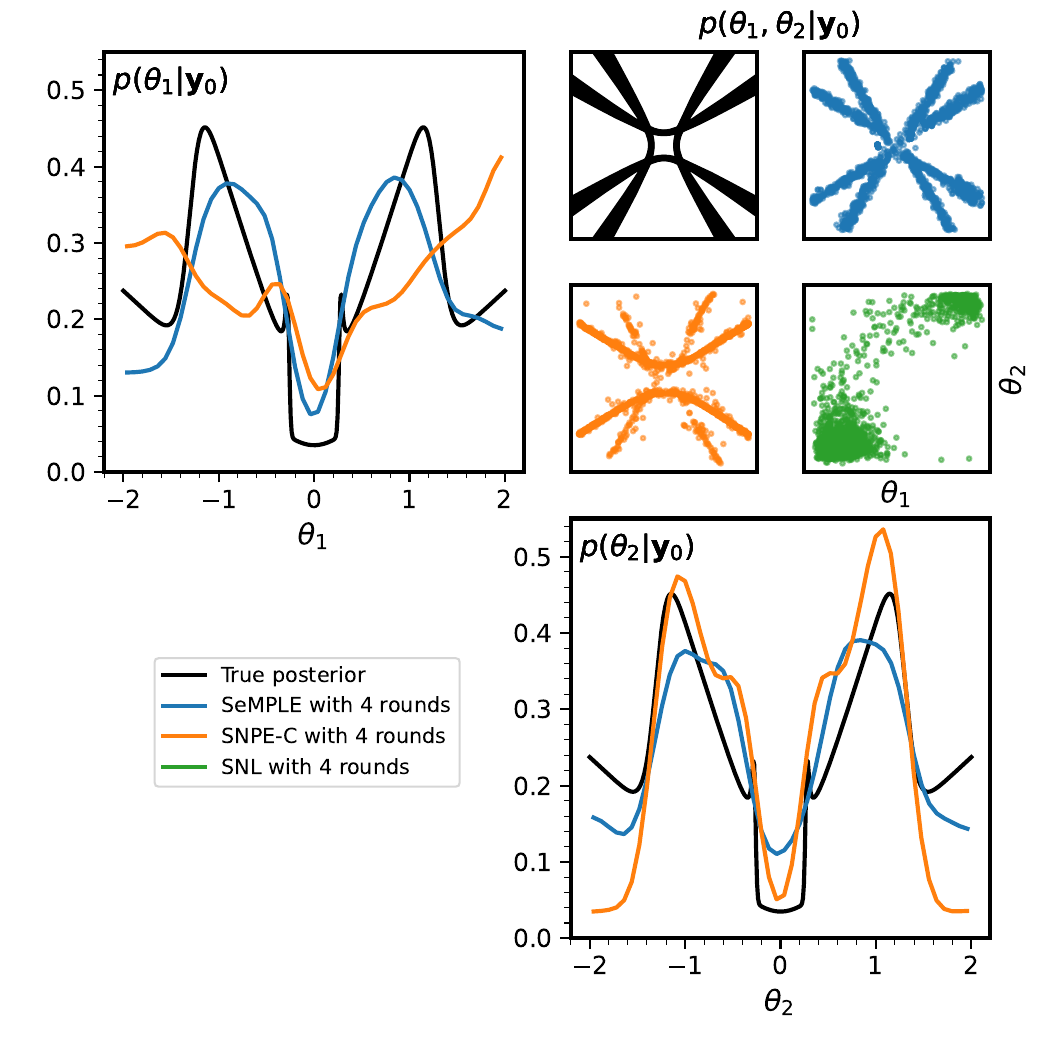}
    \caption{Repetition 2}
    \end{subfigure}
    \hfill
    \begin{subfigure}[b]{0.48\linewidth}
        \includegraphics[scale=0.45]{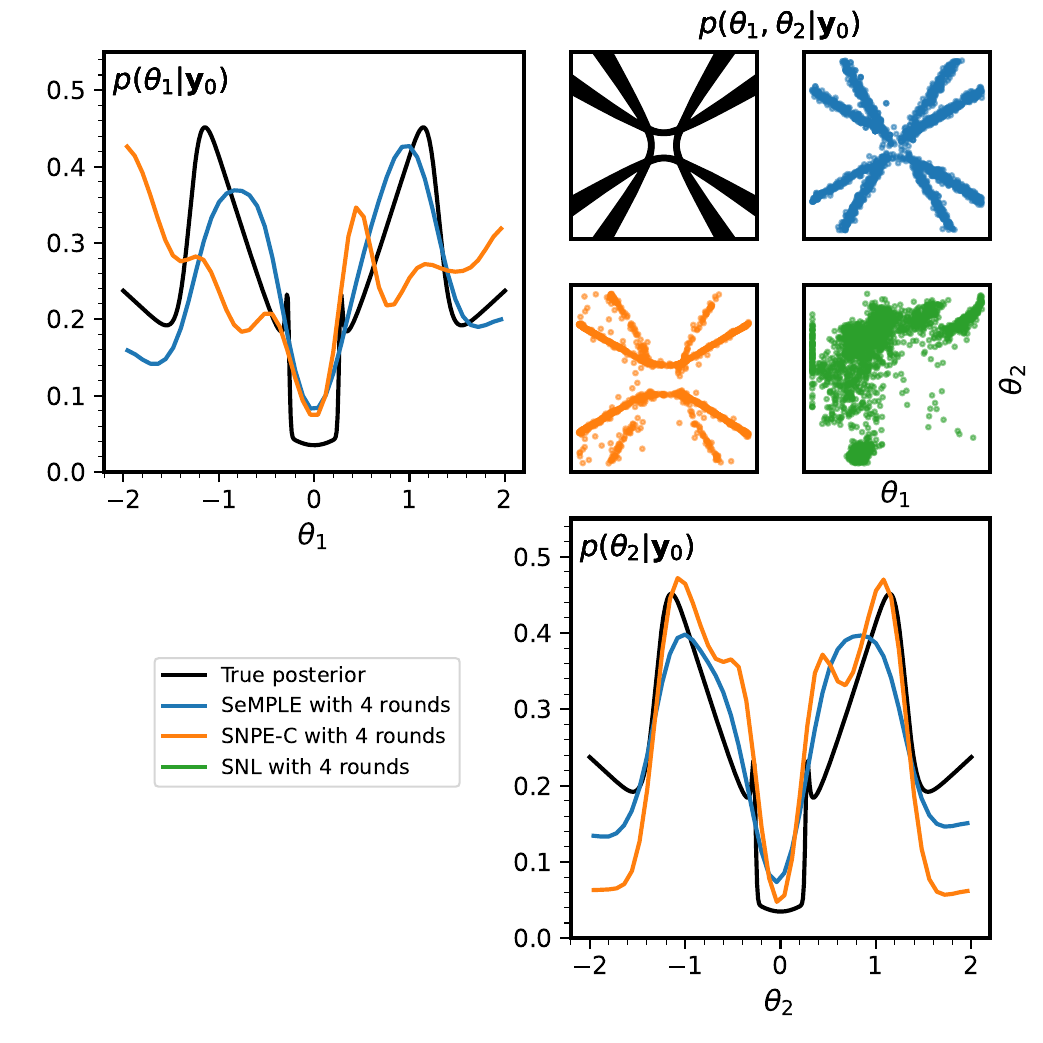}
        \caption{Repetition 3}
    \end{subfigure}
    \hfill
    \begin{subfigure}[b]{0.48\linewidth}
        \includegraphics[scale=0.45]{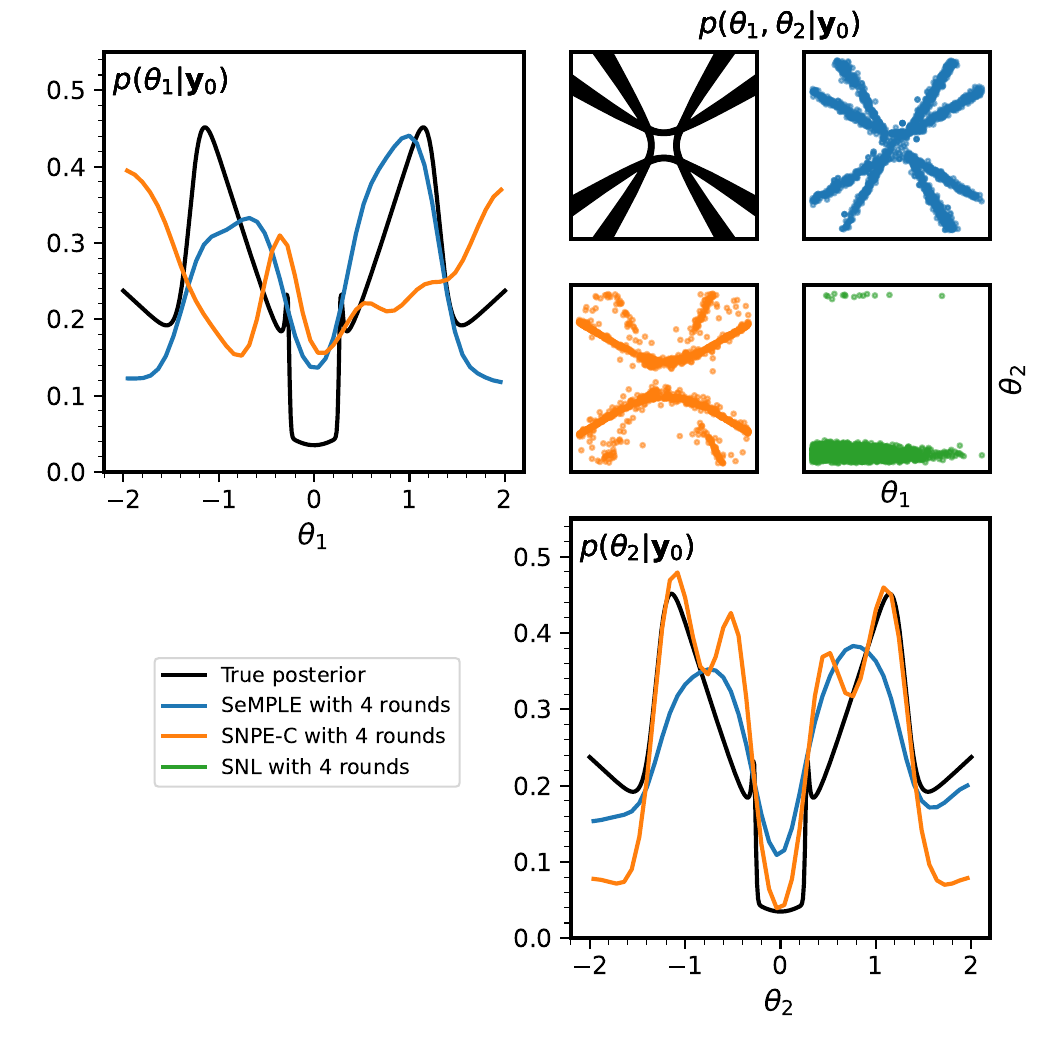}
    \caption{Repetition 4}
    \end{subfigure}
    
    \caption{Multiple hyperboloid model. Pair plots of posterior samples from the last \textcolor{black}{round} ($r=4$) of each algorithm. The SNL marginal posteriors are not reported for ease of reading, since the SNL inference fails.}

    \label{fig:hyperboloid_pairplot_run1-4}

\end{figure}

\begin{figure}[h]
    \centering
    
    \begin{subfigure}[b]{0.48\linewidth}
        \includegraphics[scale=0.45]{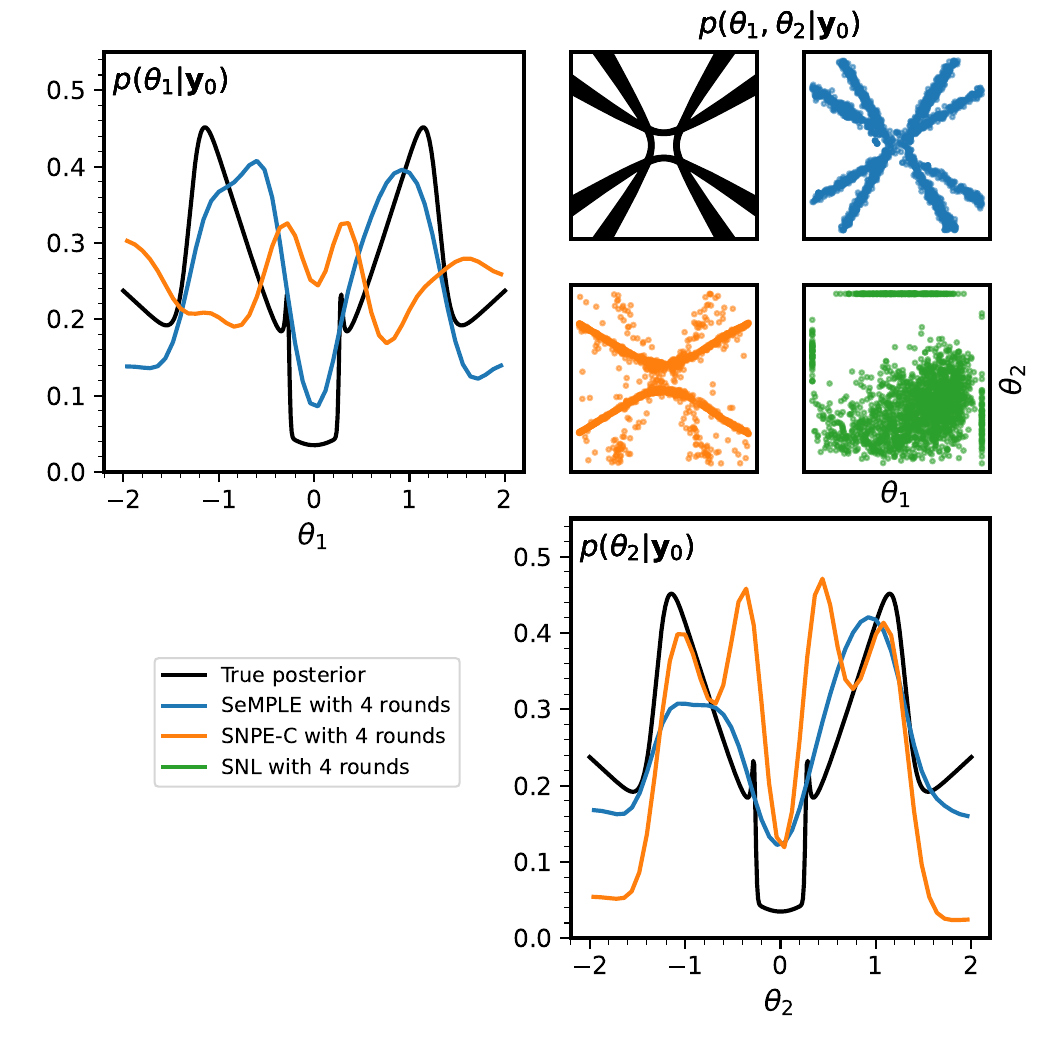}
    \caption{Repetition 5}
    \end{subfigure}
    \hfill
    \begin{subfigure}[b]{0.48\linewidth}
        \includegraphics[scale=0.45]{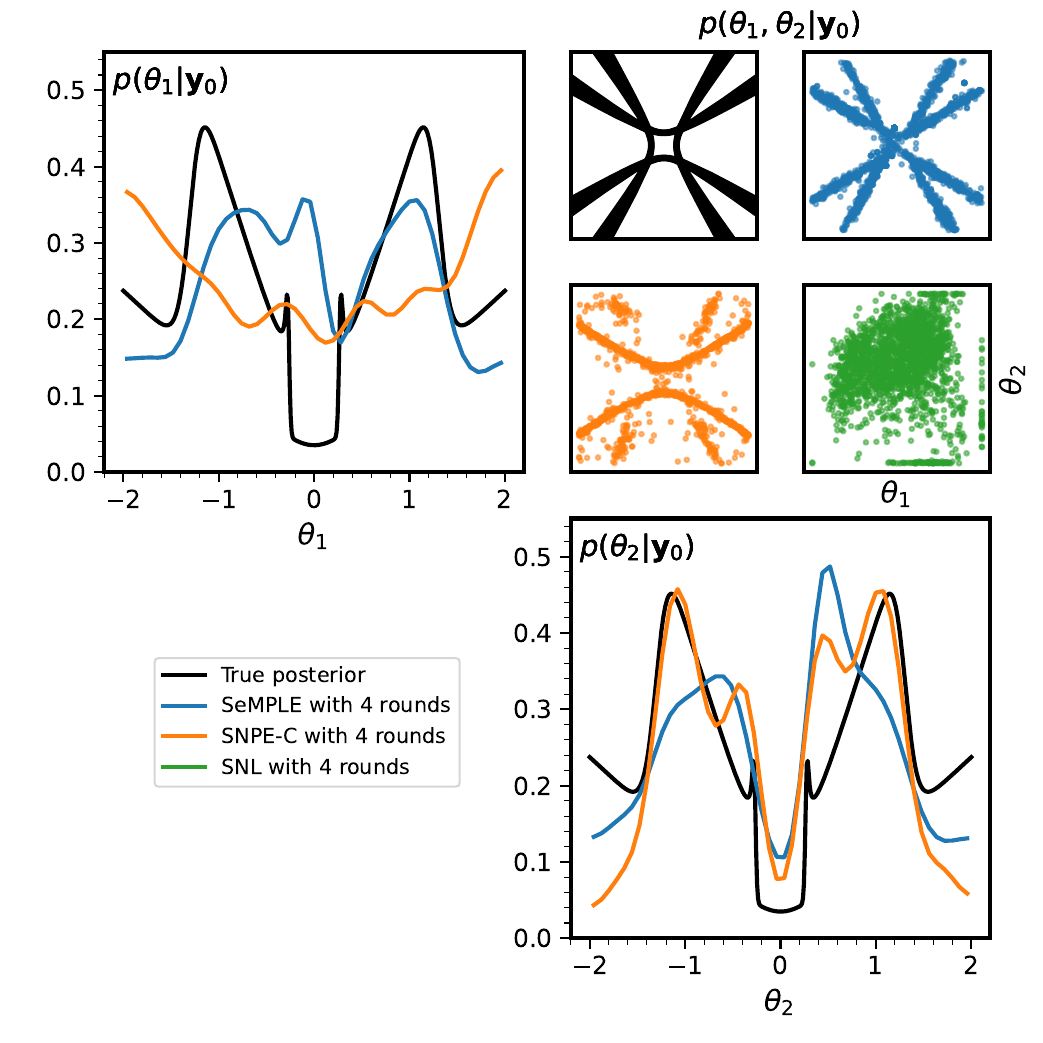}
    \caption{Repetition 6}
    \end{subfigure}
    \hfill
    \begin{subfigure}[b]{0.48\linewidth}
        \includegraphics[scale=0.45]{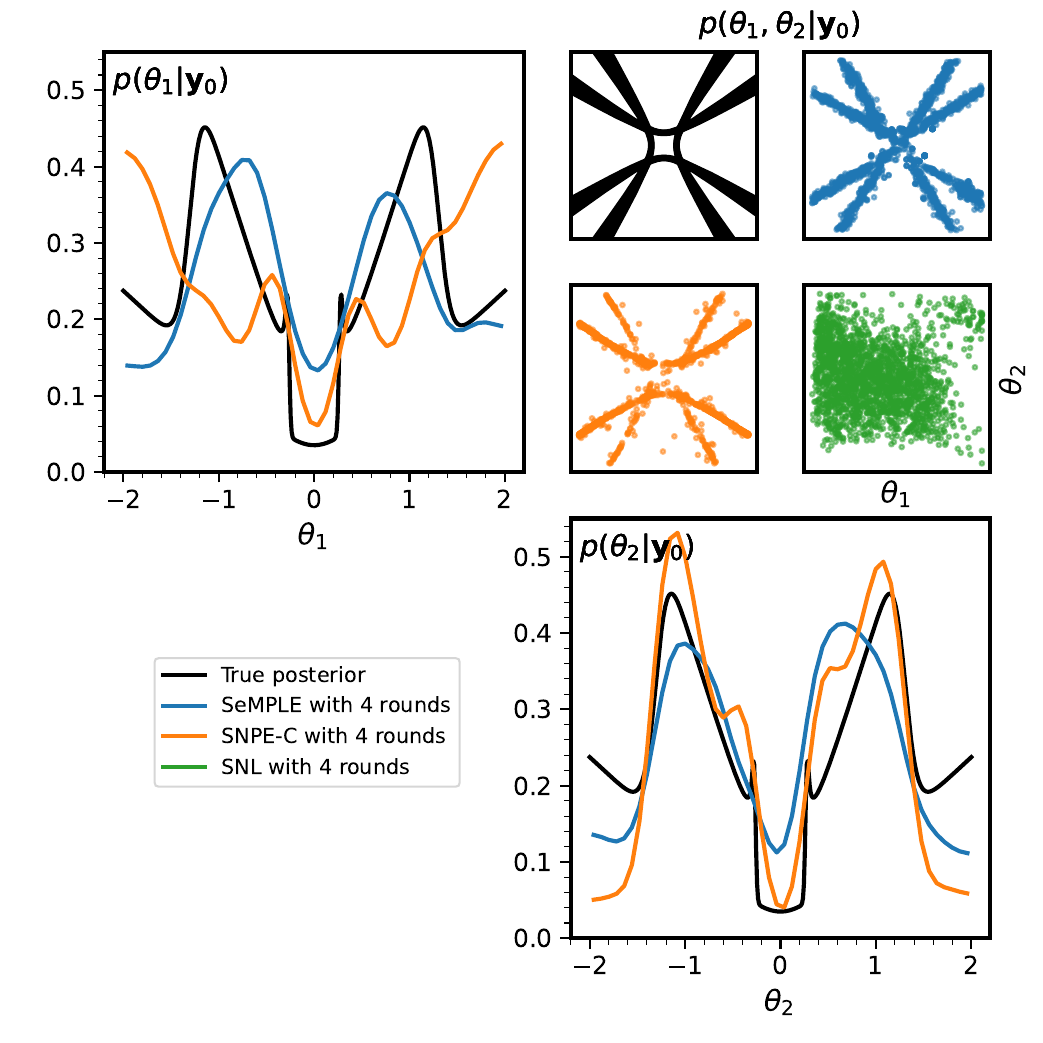}
        \caption{Repetition 7}
    \end{subfigure}
    \hfill
    \begin{subfigure}[b]{0.48\linewidth}
        \includegraphics[scale=0.45]{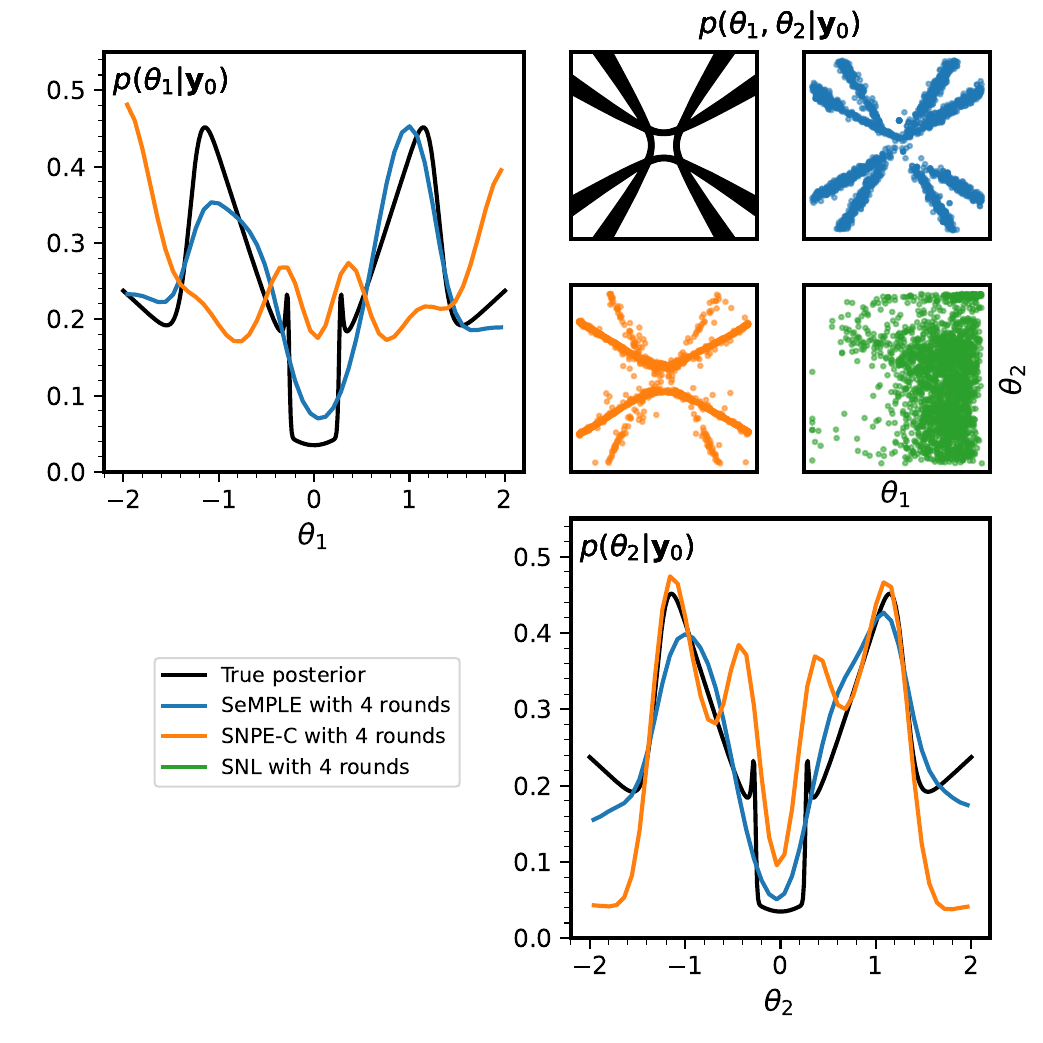}
    \caption{Repetition 8}
    \end{subfigure}
    
    \caption{Multiple hyperboloid. Pair plots of posterior samples from the last \textcolor{black}{round} ($r=4$) of each algorithm. The SNL marginal posteriors are not reported for ease of reading, since the SNL inference fails.}

    \label{fig:hyperboloid_pairplot_run5-8}

\end{figure}

\begin{figure}[h]
    \centering
    
    \begin{subfigure}[b]{0.48\linewidth}
        \includegraphics[scale=0.45]{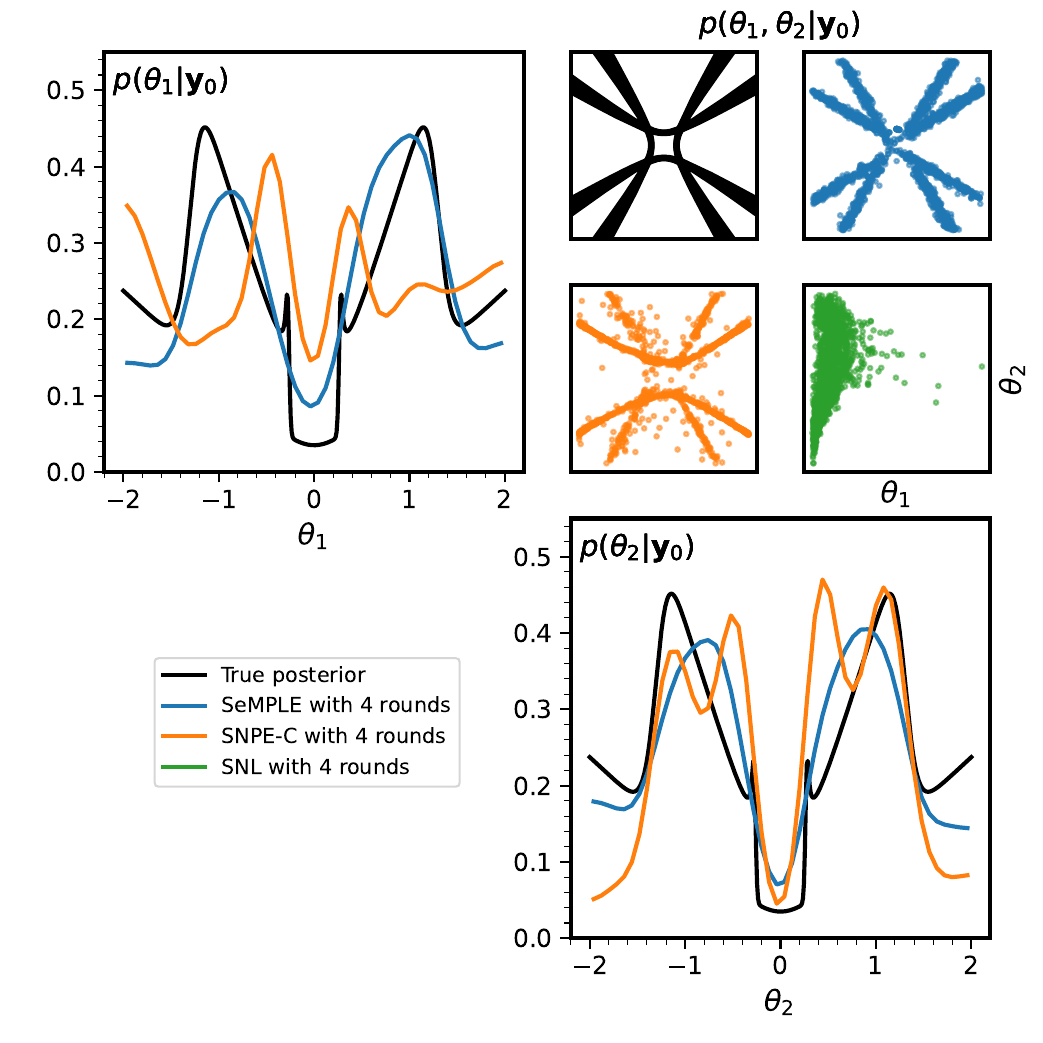}
    \caption{Repetition 9}
    \end{subfigure}
    \hfill
    \begin{subfigure}[b]{0.48\linewidth}
        \includegraphics[scale=0.45]{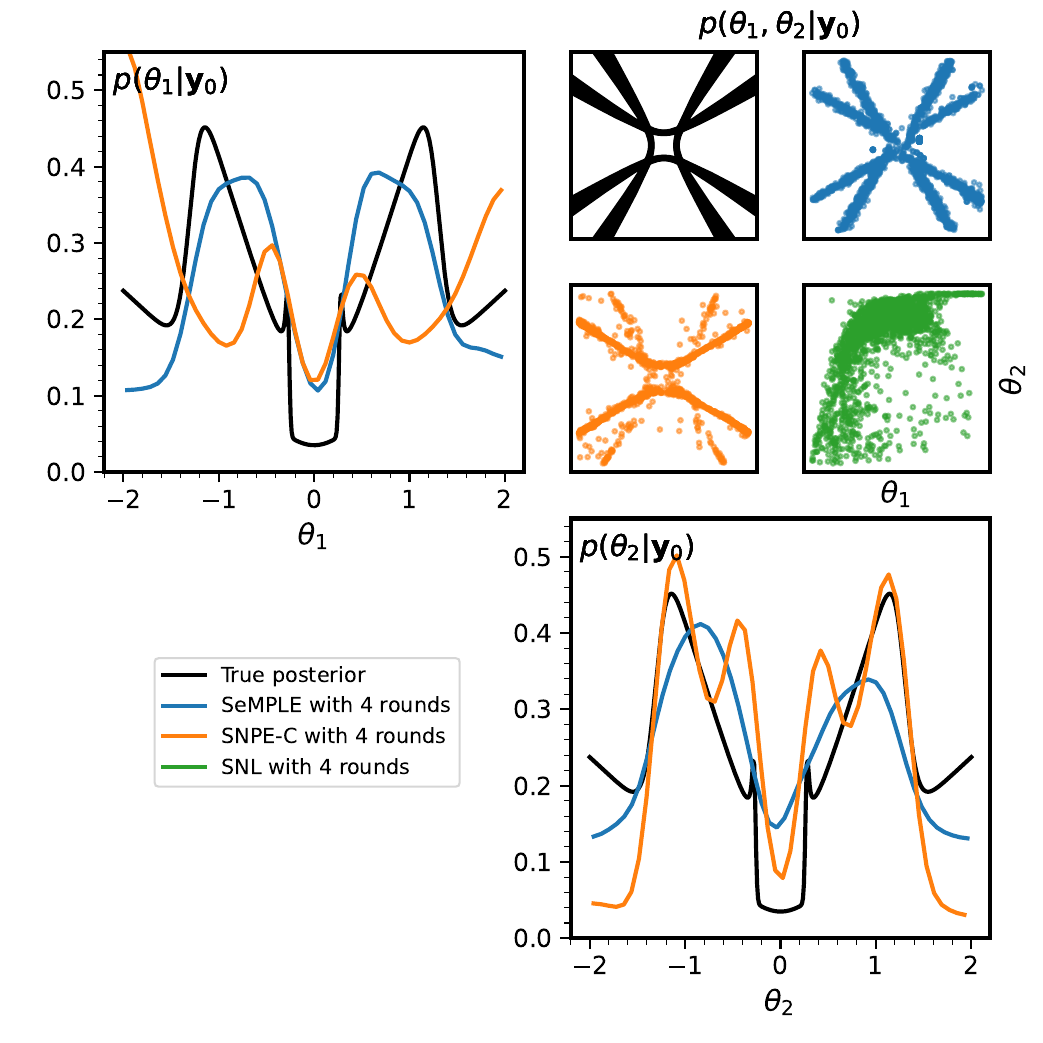}
    \caption{Repetition 10}
    \end{subfigure}
    
    \caption{Multiple hyperboloid. Pair plots of posterior samples from the last \textcolor{black}{round} ($r=4$) of each algorithm. The SNL marginal posteriors are not reported for ease of reading, since the SNL inference fails.}

    \label{fig:hyperboloid_pairplot_run9-10}

\end{figure}

In the main paper we have considered the inference results based on a total of $4\times 10^4$ model simulations, uniformly distributed across $R=10$ rounds for SNPE-C and SNL (as per \texttt{SBIBM} defaults), and $R=4$ rounds for SeMPLE. Results showed that SeMPLE performed better and in particular that SNL (which is also an MCMC-based procedure, similarly to SeMPLE) was returning the worst performance. Here we show the performance of both SNPE-C and SNL with $R=4$, again using a total of $4\times 10^4$ model simulations for each method. 
Figures \ref{fig:hyperboloid_pairplot_run1-4}-\ref{fig:hyperboloid_pairplot_run9-10} show posterior samples from the last \textcolor{black}{round} (r=4) of each algorithm for each of the 10 independent algorithm runs. We can clearly notice that SNL fails completely and therefore, for ease of reading, the SNL marginal posteriors are not reported. However, the differences between SNPE-C and SeMPLE are more subtle. It appears that occasionally SeMPLE samples from central areas, where there should not be any posterior mass, while SNPE-C seems to undersample some of the  {\it branches}. Therefore, we also plot the corresponding kernel-smoothed marginals in Figures \ref{fig:hyperboloid_pairplot_run1-4}-\ref{fig:hyperboloid_pairplot_run9-10}. There, it appears that the marginals from SeMPLE are overall closer to the reference marginals, which is in agreement with the better C2ST attained by SeMPLE (see the main text).

\begin{figure}[h]
    \centering
    \includegraphics[scale=0.4]{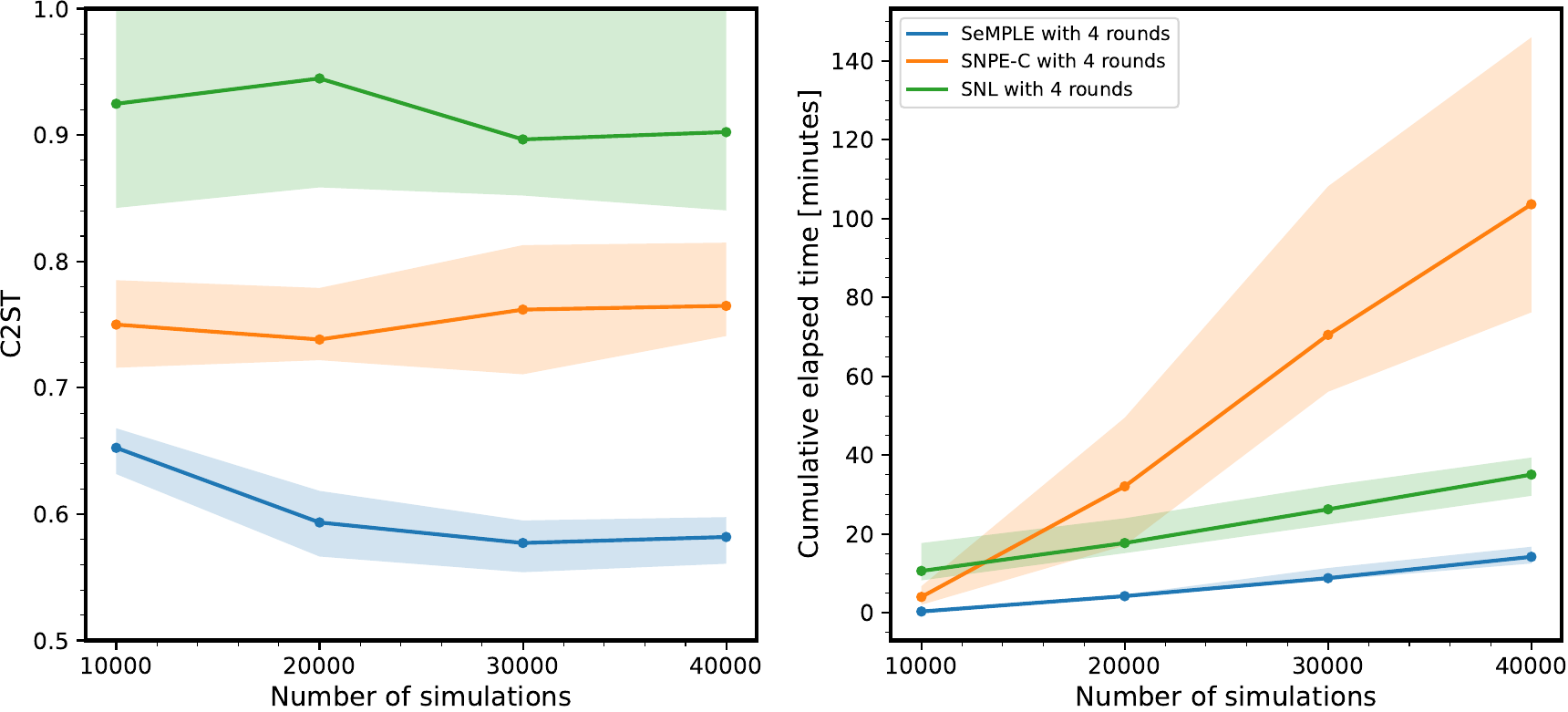}
    \caption{Hyperboloid. Median C2ST (left) and median cumulative runtime in minutes (right) for 10 runs with the same data vs the number of model simulations. Shaded bands enclose the min and max values.}
    \label{fig:hyperboloid_metrics_4rounds}
\end{figure}

Figure \ref{fig:hyperboloid_metrics_4rounds} shows C2ST and runtime values in the case when $R=4$ rounds for all algorithms. The C2ST results of SNPE-C and SNL are very similar with $R=4$, meaning that SNL fails completely and SeMPLE still performs much better than SNPE-C. In terms of runtime, SNPE-C is still about 7 times slower than SeMPLE, disregarding SNL as the inference from this method is not useful.

\clearpage

\section{Bernoulli GLM model}\label{sec:bernoulli}

\subsection{Bernoulli GLM model definition}\label{sec:bernoulli-model}

For the model definition, please see section T.5 in the appendix of \cite{sbibm}. The prior distribution and model parameters can also be found there.

\subsection{Inference setup}\label{sec:bernoulli-setup}
\begin{figure}[h]
    \centering
    \includegraphics[width=0.35\textwidth]{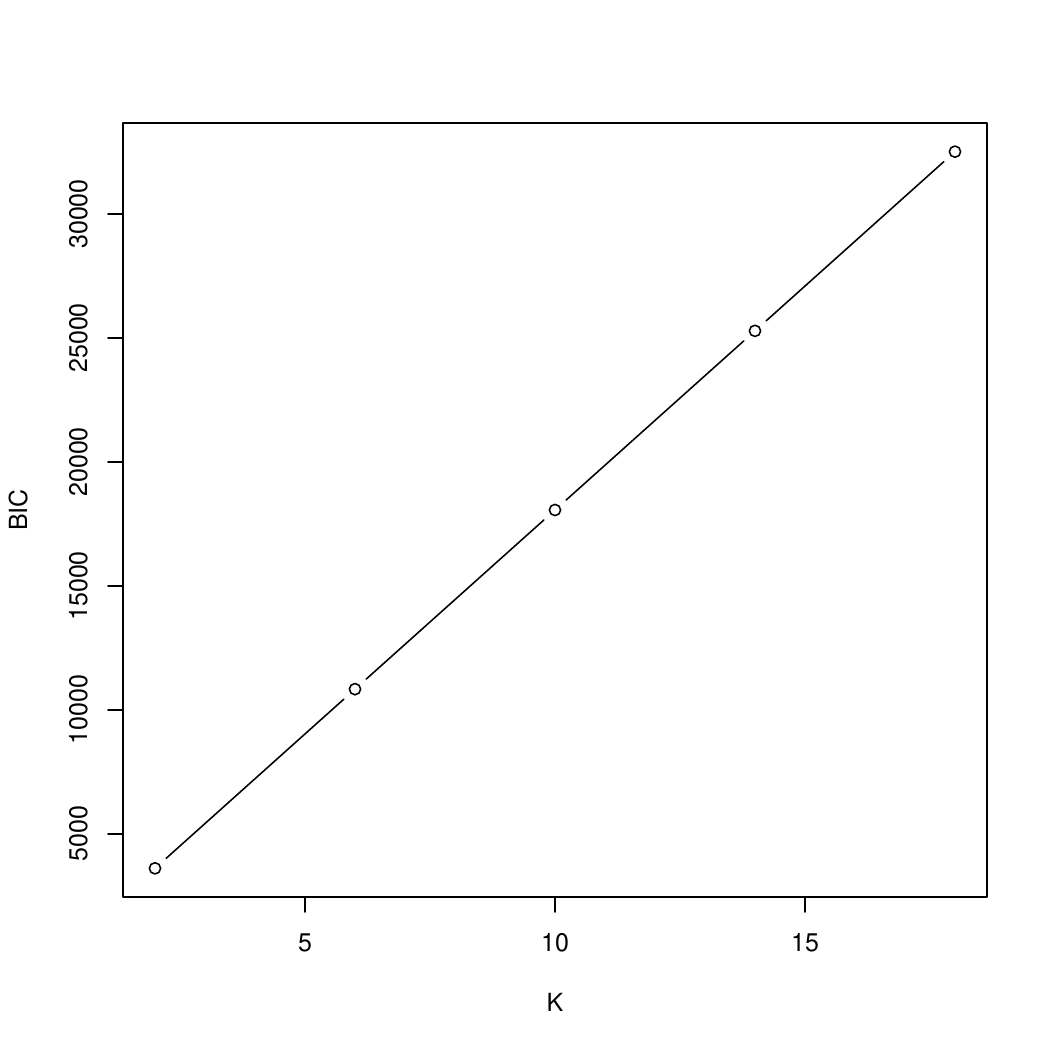}
    \caption{Bernoulli GLM model. BIC with respect to the number of Gaussian mixture components $K$.}
    \label{fig:bic_bernoulli}
\end{figure}

For all inference algorithms, we executed runs with the 10 different observed data sets provided in \texttt{SBIBM}. The total simulation budget was always set to $10,000$ model simulations, but two different setups were used for the number of algorithm rounds. In the first setup all algorithms used $R=2$ algorithm rounds and in the second setup SNL and SNPE-C used $R=10$ rounds while SeMPLE still used $R=2$ rounds. The reason for running $R=2$ rounds with SeMPLE was that for this model the output from $r=2$ produced accurate inference but had some stickiness in the Markov chain that caused problems in the GLLiM learning in round $r=3$. Hence, the best inference results with SeMPLE were obtained with $R=2$.
The GLLiM covariance structure was set to be a full unconstrained matrix as our experiments showed that this improved the inference results compared to a constrained covariance matrix structure.
Figure \ref{fig:bic_bernoulli} shows  BIC for several values of $K$, computed by fitting GLLiM using the same prior-predictive data set $\{\boldsymbol\theta_n, \by_n\}_{n=1}^{N}$ with $N=2\,500$ prior to running SeMPLE. The minimum value is obtained at $K=2$ but because of this small value, $K$ was instead set to 10 as this gives more flexibility to the model without resulting in significantly longer runtimes. The runtime to compute all BIC values in Figure \ref{fig:bic_bernoulli} was 16 seconds. Note that the BIC computation is independent of the observed data set and only has to be performed once before running SeMPLE.

\subsection{Results when using $R=2$ for all algorithms}\label{sec:bernoulli-results_2rounds}

\begin{figure}[tp]
     \centering
     \begin{subfigure}[b]{0.65\textwidth}
         \centering
         \includegraphics[width=\textwidth]{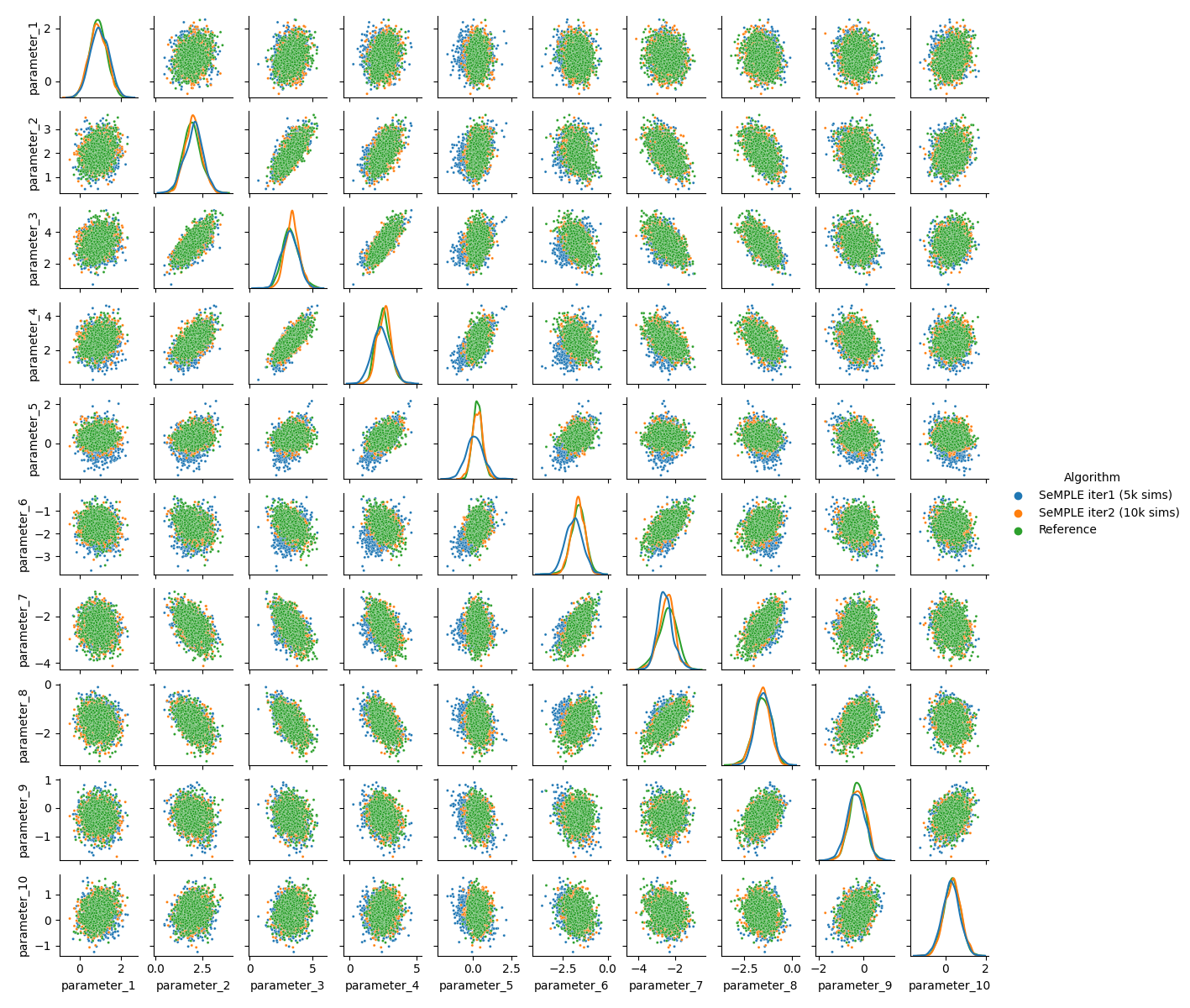}
         \caption{}
     \end{subfigure}
     \caption{Bernoulli GLM model. Posterior plots obtained with SeMPLE and obtained at round $r=1$ after 5,000 model simulations, and at $r=2$ after 10,000 model simulations.}
     \label{fig:GLM_posteriors}
\end{figure}

\begin{figure}[h]
     \centering
     \begin{subfigure}[b]{0.4\textwidth}
         \centering
         \includegraphics[width=\textwidth]{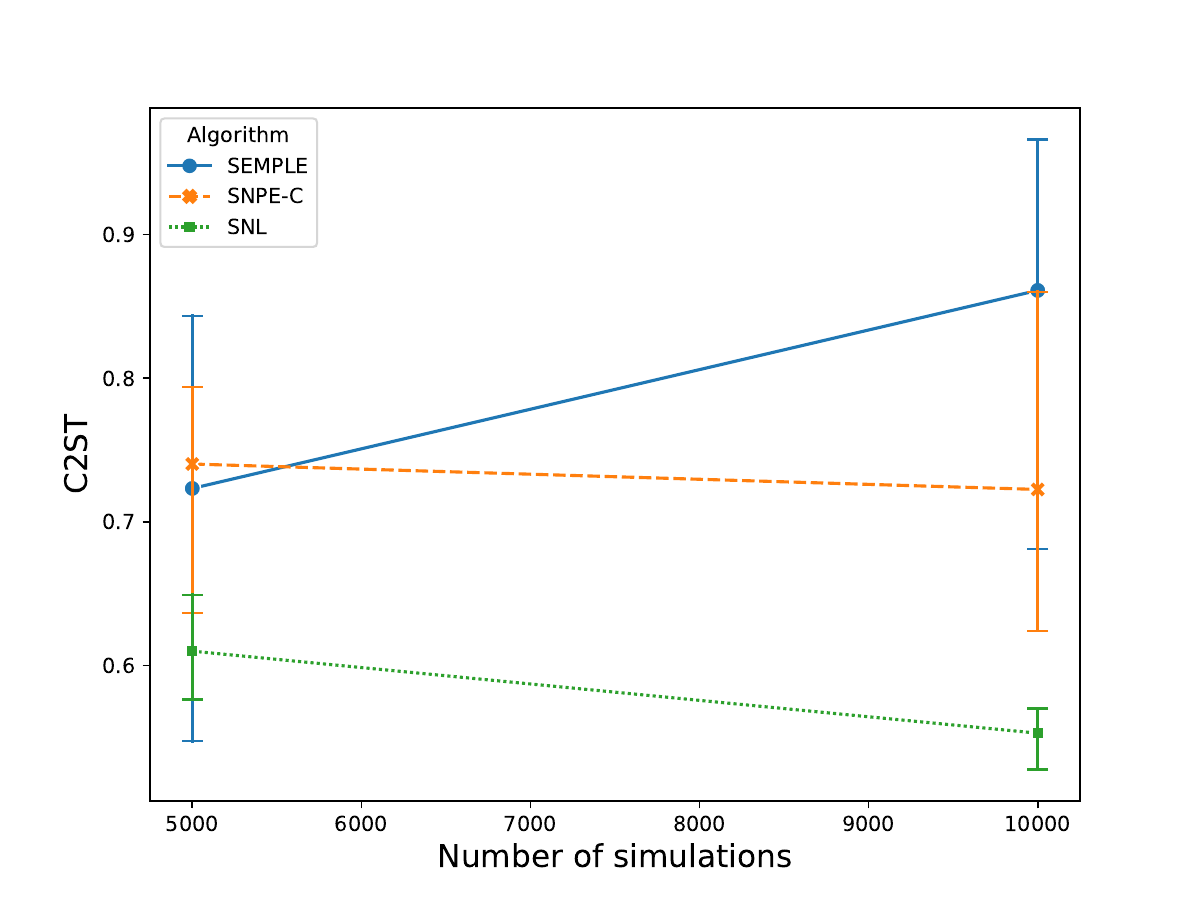}
         \caption{}
     \end{subfigure}
          \begin{subfigure}[b]{0.4\textwidth}
         \centering
         \includegraphics[width=\textwidth]{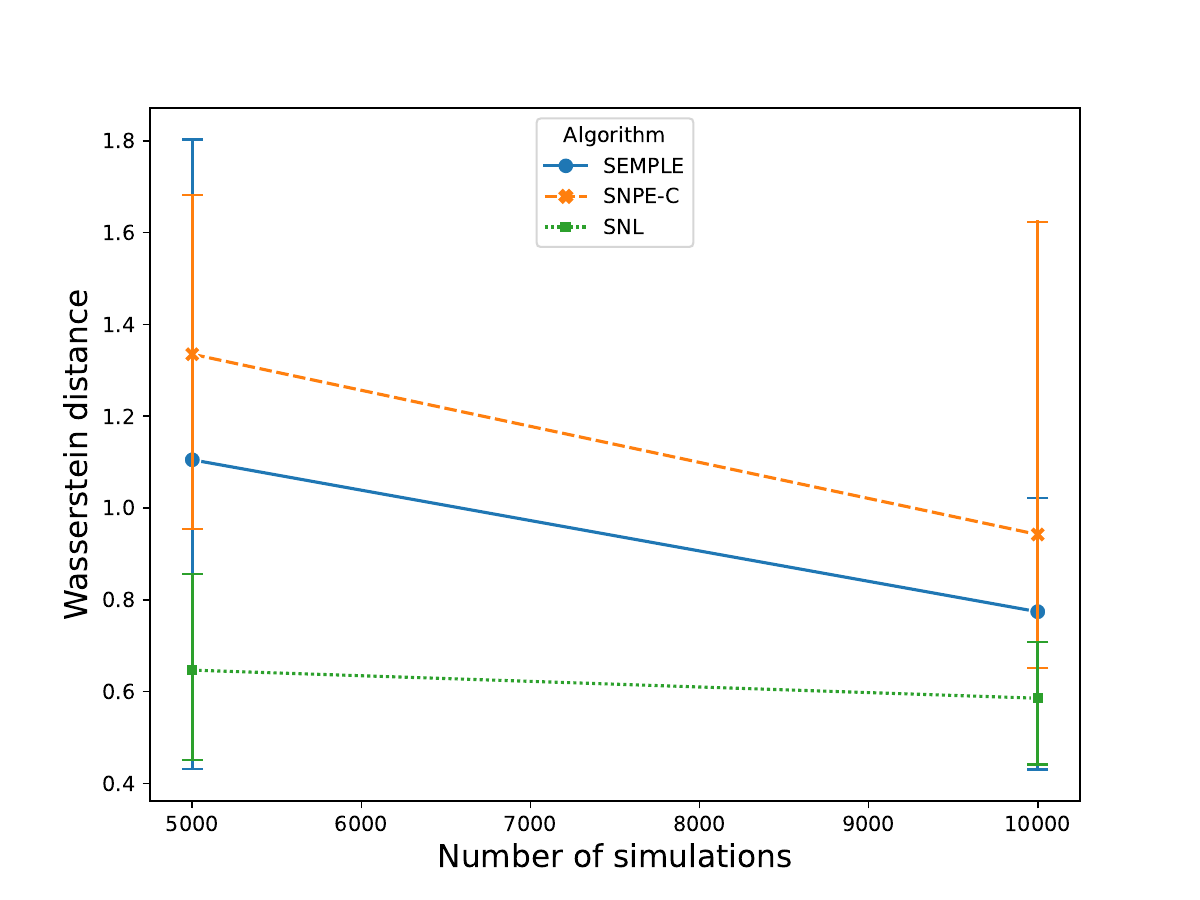}
         \caption{}
     \end{subfigure}
     \caption{Bernoulli GLM  model with $R=2$ for all algorithms. (a) Median C2ST of 10 runs with different data sets vs number of model simulations. (b) Median Wasserstein distance of 10 runs with different data sets vs number of model simulations. Error bars show min/max values. \textbf{For unimodal posteriors (as in this case) Wasserstein distances should be preferred to C2ST.}}\label{fig:GLM_settings_metric_vs_sims_2rounds}
\end{figure}

\begin{figure}[tp]
     \centering
     \begin{subfigure}[b]{0.4\textwidth}
         \centering
         \includegraphics[width=\textwidth]{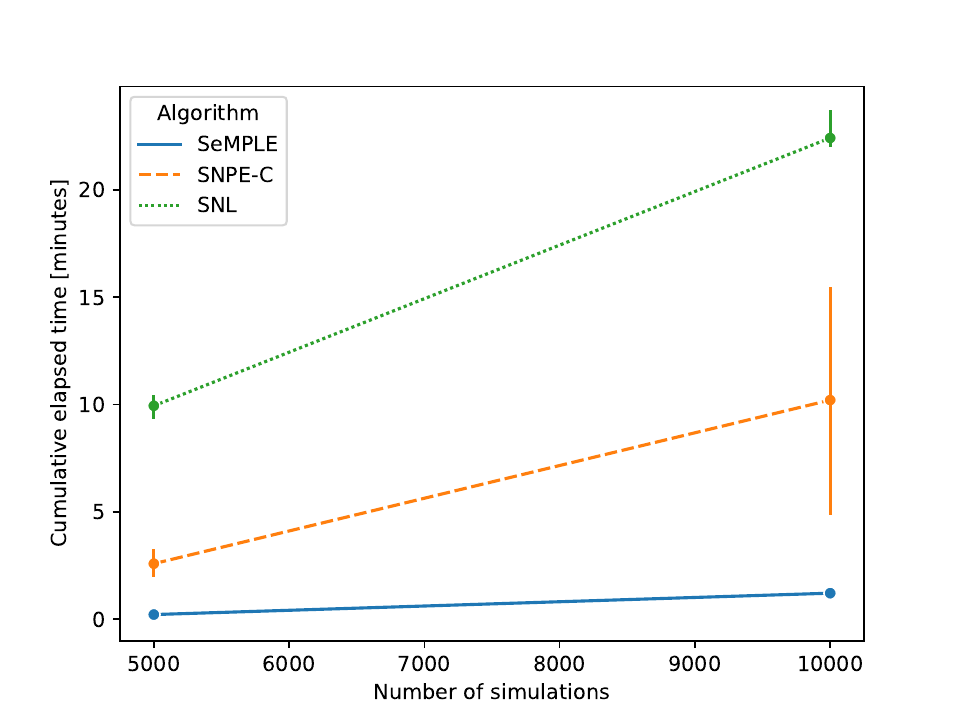}
     \end{subfigure}
     \caption{Bernoulli GLM model with $R=2$. Median running times of 10 runs using several methods. Error bars show min/max values.}
     \label{fig:GLM_runtimes}
\end{figure}

Here we compare results via SNL, SNPE-C and SeMPLE with the true posterior using $R=2$ rounds for all algorithms. Since we consider a total budget of 10,000 model simulations, Figure \ref{fig:GLM_settings_metric_vs_sims_2rounds} shows the C2ST and Wasserstein performance evaluated at 5,000 an 10,000 model simulations. We found that C2ST can be problematic with unimodal targets, and we see this also with the Ornstein-Uhlenbeck results (which also has a unimodal posterior), where the performance of the several algorithms seems to be better evaluated using Wasserstein distances. In fact, as from Figure \ref{fig:GLM_posteriors} it is clear that the results from SeMPLE at $r=2$ are noticeably improved compared to $r=1$, and this is captured by the Wasserstein distances in Figure  \ref{fig:GLM_settings_metric_vs_sims_2rounds}(b), whereas C2ST produces a contradicting result in Figure  \ref{fig:GLM_settings_metric_vs_sims_2rounds}(a). We therefore alert the reader that with unimodal targets C2ST is at times not always consistent with other measures, such as posterior plots (see also the Ornstein-Uhlenbeck section), while we found C2ST useful with multimodal targets. 
Moreover, the inference from SeMPLE is not only accurate, but the computational gain provided by SeMPLE is striking, as illustrated in Figure \ref{fig:GLM_runtimes}, where SeMPLE is 8 times faster than SNPE-C and 19 times faster than SNL. These figures increase dramatically for SNL and SNPE-C in next section, where these methods are run with the default \texttt{SBIBM} setup.

\subsection{Results when using $R=10$ for SNL/SNPE-C and $R=2$ for SeMPLE}\label{sec:bernoulli-results_10rounds}

\begin{figure}[tp]
     \centering
     \begin{subfigure}[b]{0.4\textwidth}
         \centering
         \includegraphics[width=\textwidth]{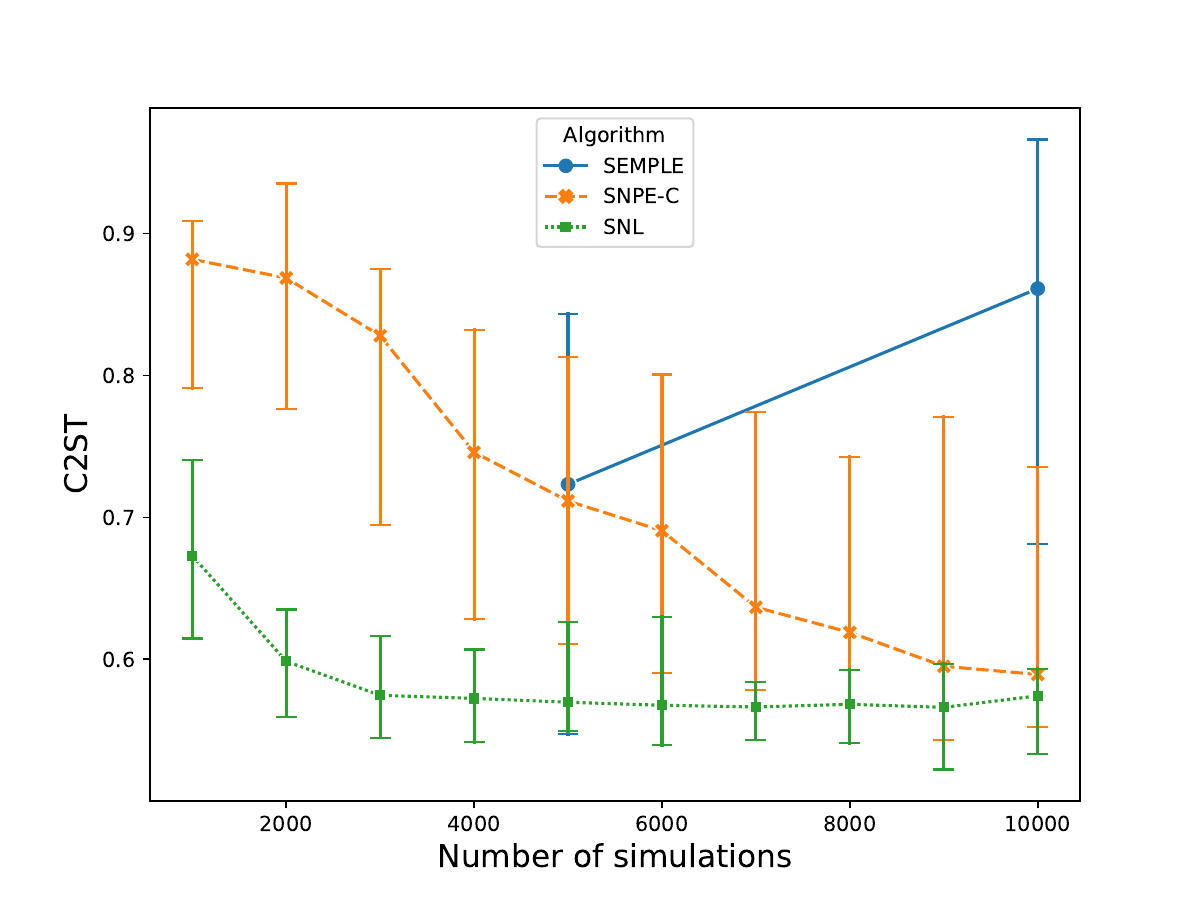}
         \caption{}
     \end{subfigure}
          \begin{subfigure}[b]{0.4\textwidth}
         \centering
         \includegraphics[width=\textwidth]{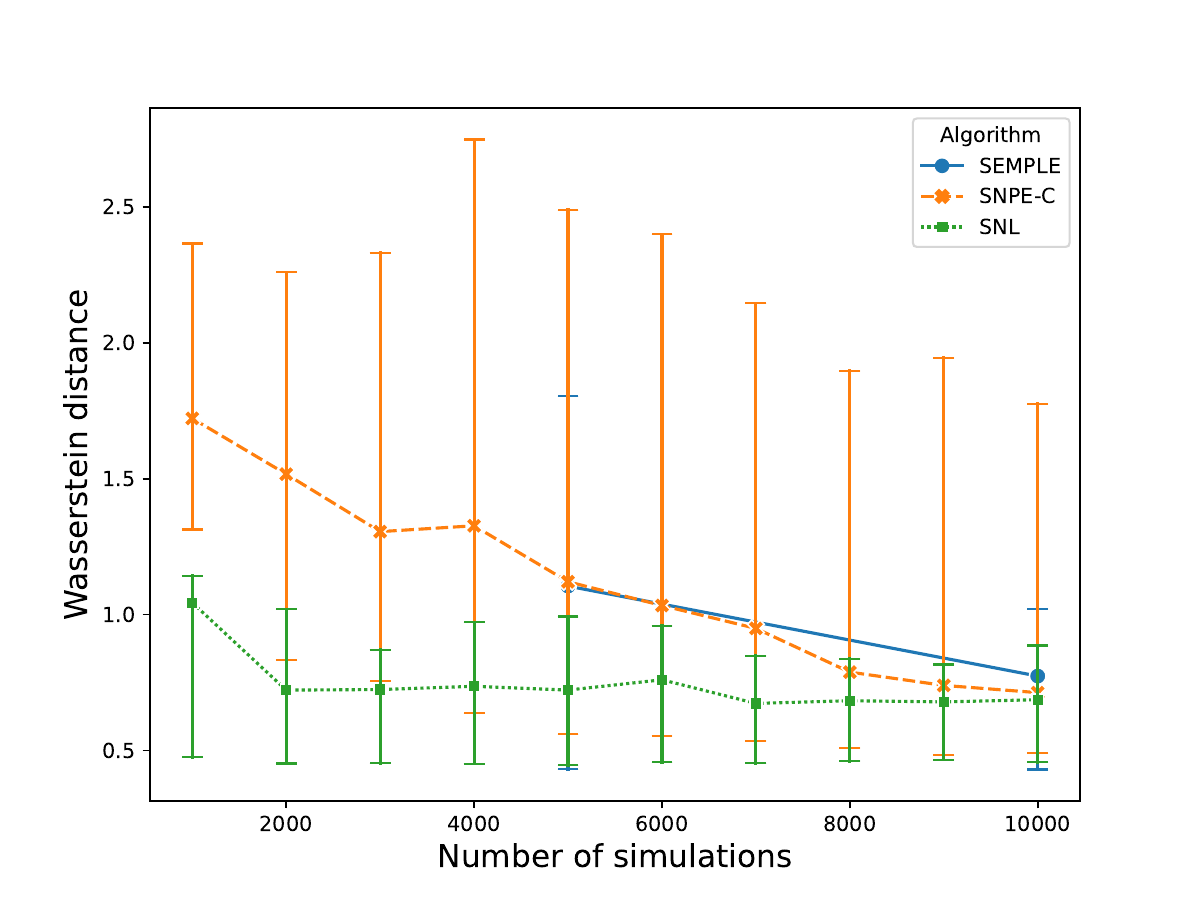}
         \caption{}
     \end{subfigure}
     \caption{Bernoulli GLM model with $R=10$ (SNL/SNPE-C) and $R=2$ (SeMPLE). (a) Median C2ST of 10 runs with different data sets vs number of model simulations. (b) Median Wasserstein distance of 10 runs with different data sets vs number of model simulations. Error bars show min/max values. \textbf{For unimodal posteriors (as in this case) Wasserstein distances should be preferred to C2ST.}}
     \label{fig:GLM_settings_metric_vs_sims_10rounds}
\end{figure}

\begin{figure}[tp]
     \centering
     \begin{subfigure}[b]{0.4\textwidth}
         \centering
         \includegraphics[width=\textwidth]{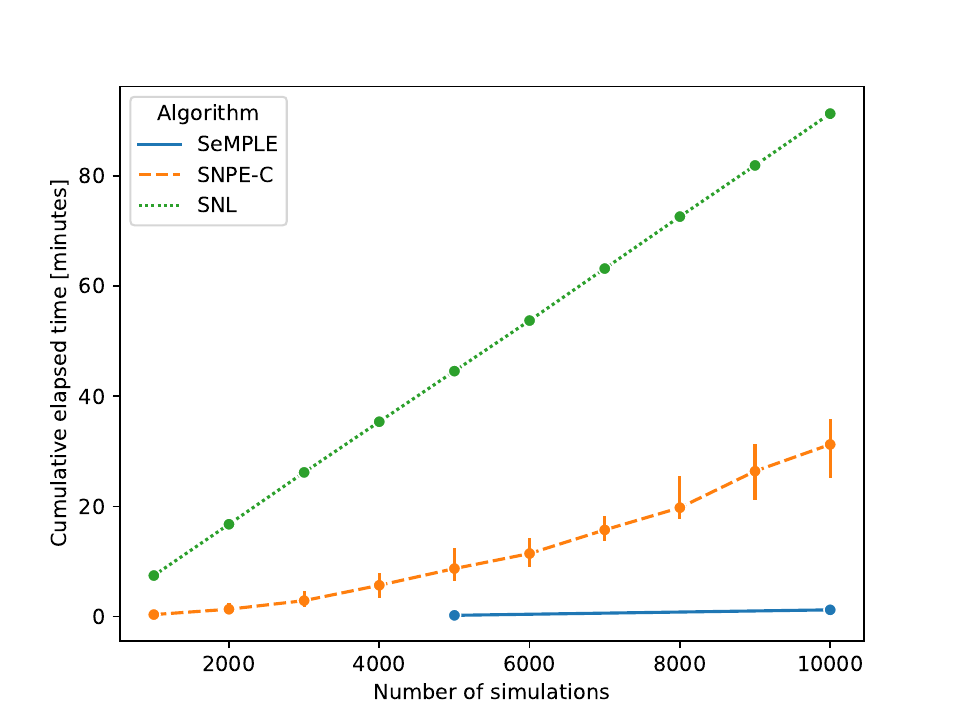}
     \end{subfigure}
     \caption{Bernoulli GLM model with $R=10$ (SNL/SNPE-C) and $R=2$ (SeMPLE). Median running times of 10 runs using several methods. Error bars show min/max values.}
     \label{fig:GLM_runtimes_10rounds}
\end{figure}

We executed the experiment anew, but this time by allocating the $10^4$ model simulations uniformly across 2 rounds for SeMPLE and 10 rounds for SNL and SNPE-C (as per \texttt{SBIBM} defaults). Results are in Figure \ref{fig:GLM_settings_metric_vs_sims_10rounds} and Figure \ref{fig:GLM_runtimes_10rounds}. Once more, we recommend the reader to consider the Wasserstein distances and not C2ST for this example, since that posterior is unimodal, as motivated in the previous section. The distances
show that on average SeMPLE is as accurate as SNL but not before $10^4$ model simulations are executed, while SNPE-C shows a large variability in the results. Once more, SeMPLE comes with the benefit of a much faster inference (Figure \ref{fig:GLM_runtimes_10rounds}): in this case SeMPLE is 26 times faster than SNPE-C and 74 times faster than SNL.

\clearpage

\section{Ornstein-Uhlenbeck}\label{sec:ou}

The Ornstein-Uhlenbeck process has original applications in physics to model the velocity of a Brownian particle \citep{ornstein}, and has found application in many contexts such as neuronal modelling and finance, to mention a few. In our work, it serves as an application where the data is a time series, while the likelihood function is tractable and therefore it is possible to obtain samples from the true posterior, to use as reference.
The Ornstein-Uhlenbeck process $\{X_t\}_{t\geq 0}$ is defined by the stochastic differential equation (SDE)
\begin{equation}
    \label{eq:OU_sde}
    dX_t = -\beta(X_t-\alpha)dt + \sigma dW_t,
\end{equation}
where $\beta>0$, $\alpha \in \mathbb{R}$, $\sigma>0$, $X_0=x_0$ and $W_t$ denotes the Wiener process (Brownian motion). The solution to equation \eqref{eq:OU_sde} is 
\begin{equation}
    \label{eq:OU_solution}
    X_t = \alpha + (x_0-\alpha)e^{-\beta t} + \sigma \int_{0}^{t} e^{-\beta(t-s)} dW_s.
\end{equation}
The process has known Gaussian transition densities: for a generic time $t$ and an arbitrary state $x_0$ at time $t_0=0$, the transition density is given as
\begin{equation}
    \label{eq:OU_transition}
    (X_t \g X_0=x_0) \sim \mathcal{N}\biggl(\alpha + (x_0-\alpha)e^{-\beta t}, \frac{\sigma^2}{2\beta}(1-e^{-2\beta t})\biggr).
\end{equation}
The transition density \eqref{eq:OU_transition} allows exact simulation of a solution path to the Ornstein-Uhlenbeck SDE. Starting at $t_0 = 0$ one can simulate $n-1$ further points of the solution at instants $t_1,\ldots, t_{n-1}$, letting $\Delta_i = t_i - t_{i-1}$, by using the transition recursively as following
\begin{equation}
    \label{eq:OU_transition_delta}
    (X_{t_i} \g X_{t_{i-1}} = x_{t_{i-1}}) \sim \mathcal{N}\biggl(\alpha + (x_0-\alpha)e^{-\beta \Delta_i}, \frac{\sigma^2}{2\beta}(1-e^{-2\beta \Delta_i})\biggr), \quad i=1,\ldots,n-1.
\end{equation}
This makes the Ornstein-Uhlenbeck model one of the very few examples of SDEs where it is trivial to evaluate the exact likelihood function.
The likelihood function for the parameters $\boldsymbol \theta = (\alpha,\beta,\sigma)$ is written as (due to the Markovianity of the solution of an SDE) 
\begin{equation}
    \label{eq:OU_likelihood}
    \mathcal{L}(x_0,\ldots,x_{n-1} \g \boldsymbol \theta) = \prod_{i=1}^{n-1} p(x_i \g x_{i-1};\boldsymbol \theta),
\end{equation}
where $p(x_i \g x_{i-1};\boldsymbol \theta)$ is the transition density in equation \eqref{eq:OU_transition_delta} (we assume $x_0$ a fixed constant throughout, otherwise a multiplicative term $p(x_0|\btheta)$ would have to appear in \eqref{eq:OU_likelihood}).
The interpretation of the parameter $\alpha$ is the asymptotic mean. This can be seen by letting $t \rightarrow \infty$ in equation \eqref{eq:OU_transition} which shows that the process' stationary distribution is $X_t \sim \mathcal{N}(\alpha, \frac{\sigma^2}{2\beta})$. The parameter $\sigma$ can be interpreted as the variation or the size of the noise, while $\beta$ can be interpreted as the growth-rate or how strongly the system reacts to perturbations.

\subsection{Inference setup}\label{sec:ou-inference}
\begin{figure}[h]
    \centering
    \includegraphics[width=0.4\textwidth]{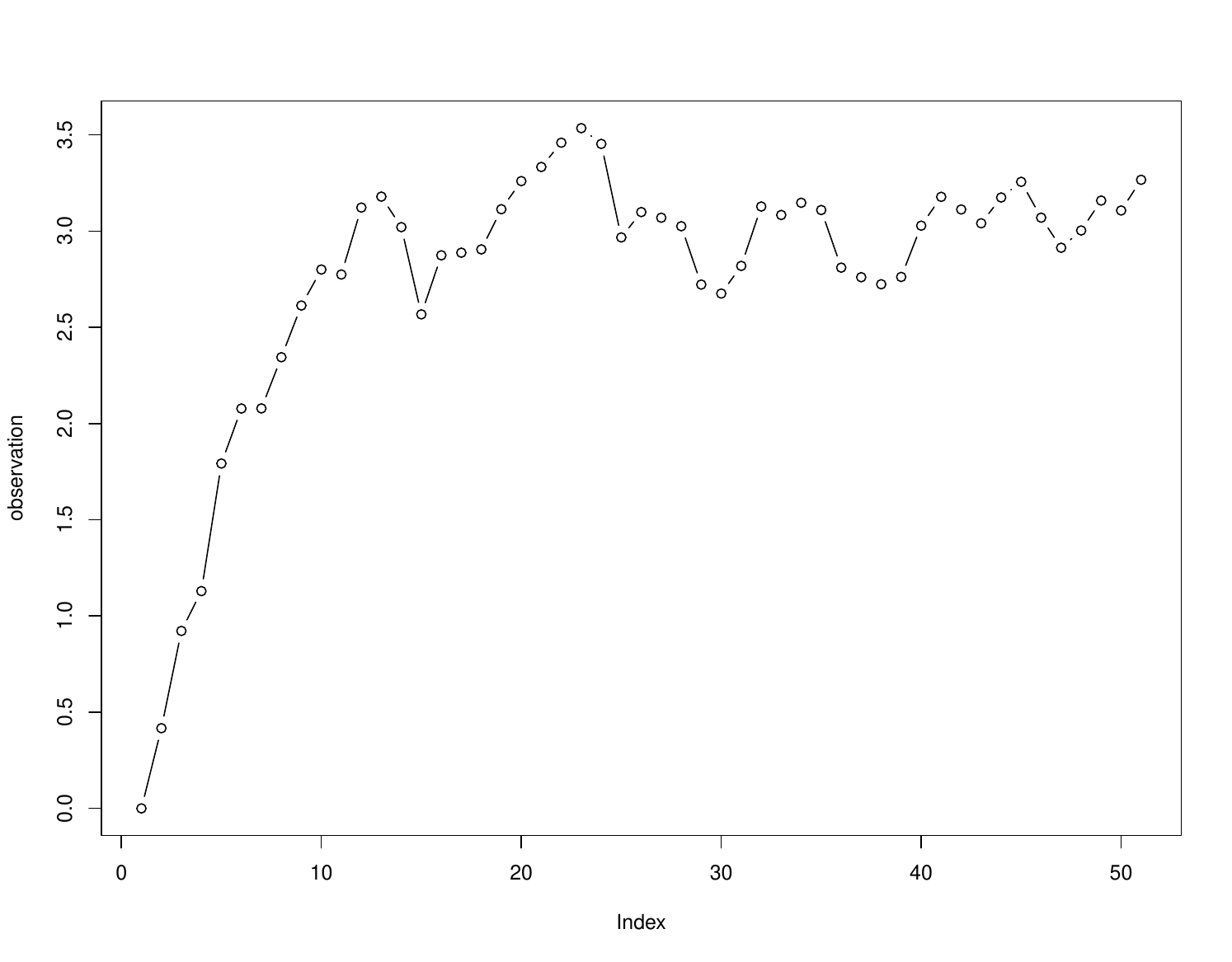}
    \caption{Ornstein-Uhlenbeck: Observed time series data of the process.}
    \label{fig:ornstein_uhlenbeck_observed_data}
\end{figure}

The fixed initial value was set to $x_0 = 0$. We set uniform priors $\alpha \sim \mathcal{U}(0,10)$, $\beta \sim \mathcal{U}(0,5)$ and $\sigma \sim \mathcal{U}(0,2)$. In this study, 50 points were simulated from the process to obtain a time series of length $n=51$ (including the fixed starting value $x_0$) using $\boldsymbol \theta = (\alpha, \beta, \sigma) = (3,1,0.5)$ as ground-truth.   The time frame of the process was set to $[t_0,T] = [0,10]$, and the discrete time points $t_1,\ldots, t_{n-1}$ set to be equally spaced in this interval. The resulting observed data is shown in Figure \ref{fig:ornstein_uhlenbeck_observed_data}.
Similarly to the multiple hyperboloid model, a single observed data set was produced by running the simulation model once with the true parameter values $\btheta = (3,1,0.5)$. 

\begin{figure}[h]
    \centering
    \includegraphics[width=0.5\textwidth]{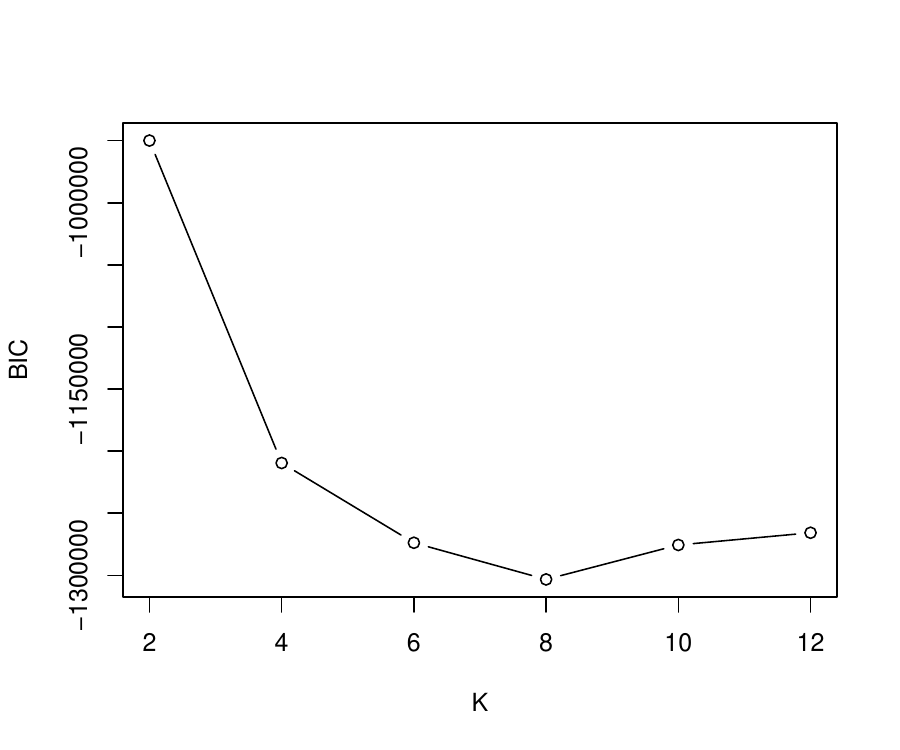}
    \caption{Ornstein-Uhlenbeck: BIC as a function of number of Gaussian mixture components $K$.}
    \label{fig:bic_OU}
\end{figure}

The covariance structure in GLLiM was set to be a full unconstrained matrix, which is reasonable given the time-dependent structure, and in your experiments it was found that this improved the inference results substantially.
Figure \ref{fig:bic_OU} shows the BIC for different values of $K$, computed by fitting surrogate likelihoods using the same prior-predictive data set $\{\boldsymbol\theta_n, \by_n\}_{n=1}^{N}$ with $N=10\,000$ prior to running SeMPLE. The minimum value is obtained at $K=8$ but because of this relatively small value, $K$ was instead set to 20 to have some buffer to remove unnecessary mixture components if needed. The runtime to compute all BIC values in Figure \ref{fig:bic_OU} was 973 seconds $\approx$ 16 minutes. Note that the BIC computation is independent of the observed data set and only has to be performed once before running SeMPLE.

\subsection{Results}\label{sec:ou-results}
The posterior is unimodal and therefore, following the results for the Bernoulli GLM example, we find it useful to report both the C2ST metric and Wasserstein distances. For all inference algorithms, we executed 10 independent runs using the same observed data set. We ran experiments using two configurations by always allocating the same total number of $4\times 10^4$ model simulations for all methods: the first  configuration uses $R=4$ rounds for all methods (section \ref{sec:ou_R=4}). The second configuration uses $R=4$ rounds for SeMPLE and $R=10$ for both SNL and SNPE-C (section \ref{sec:ou_R=4_and_R=10}), as $R=10$ is the default implementation in \texttt{SBIBM} for SNL and SNPE-C. It is instructive to look first at the case where $R=4$ is used for all algorithms, as there we can make useful remarks about the behaviour of C2ST and the Wasserstein distance.

\subsection{Results when using $R=4$ for all algorithms}\label{sec:ou_R=4}

\begin{figure}[h] 
    \captionsetup[subfigure]{justification=centering}
     \centering
     \begin{subfigure}[b]{0.39\textwidth}
         \centering
         \includegraphics[width=\textwidth]{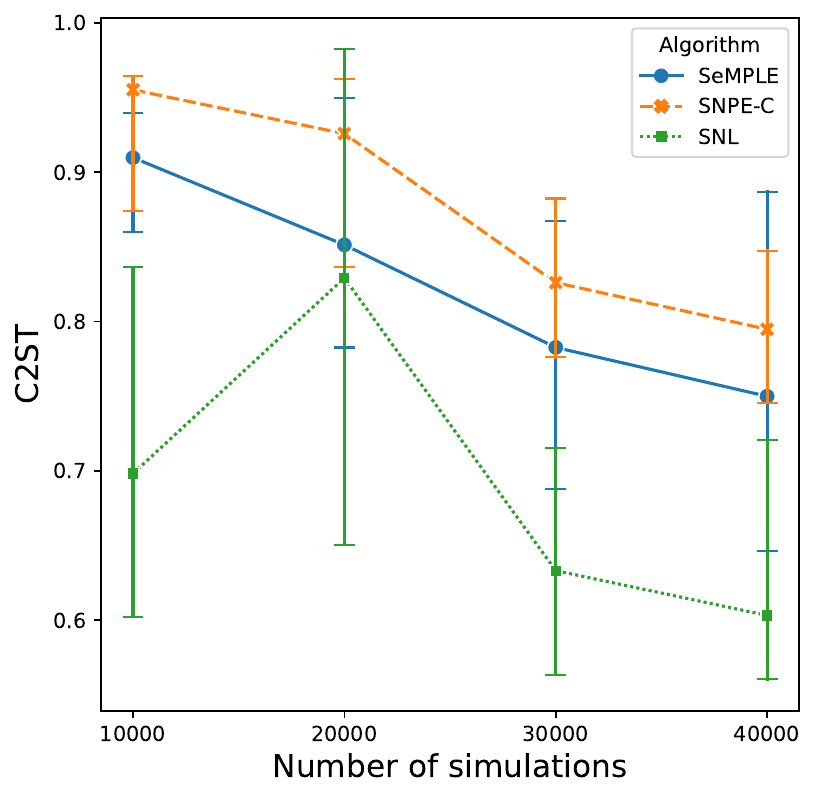}
         \caption{}
     \end{subfigure}
     \begin{subfigure}[b]{0.39\textwidth}
         \centering
         \includegraphics[width=\textwidth]{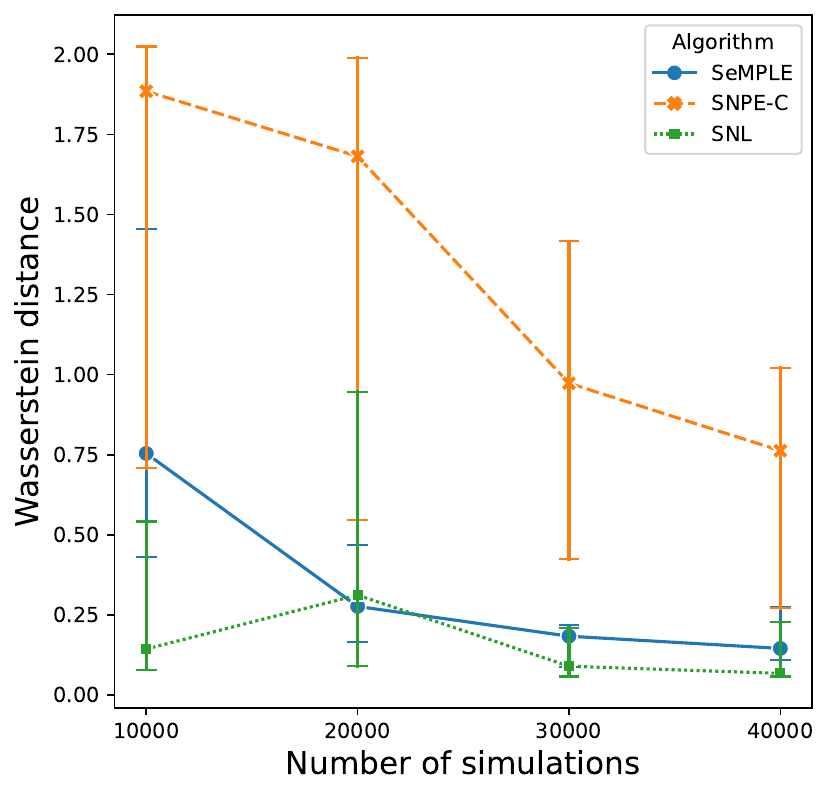}
         \caption{}
     \end{subfigure}
     \caption{Ornstein-Uhlenbeck, results when $R=4$ for all algorithms: median C2ST and Wasserstein distance across 10 independent runs (with the same observed data set) as a function of the number of model simulations. Error bars show min/max values. \textbf{For unimodal posteriors (as in this case) Wasserstein distances should be preferred to C2ST.}}
     \label{fig:OU_algorithm_metric_vs_sims}
\end{figure}

\begin{figure}[h]
    \centering
    \begin{subfigure}[b]{0.45\textwidth}
    \includegraphics[scale=0.27]{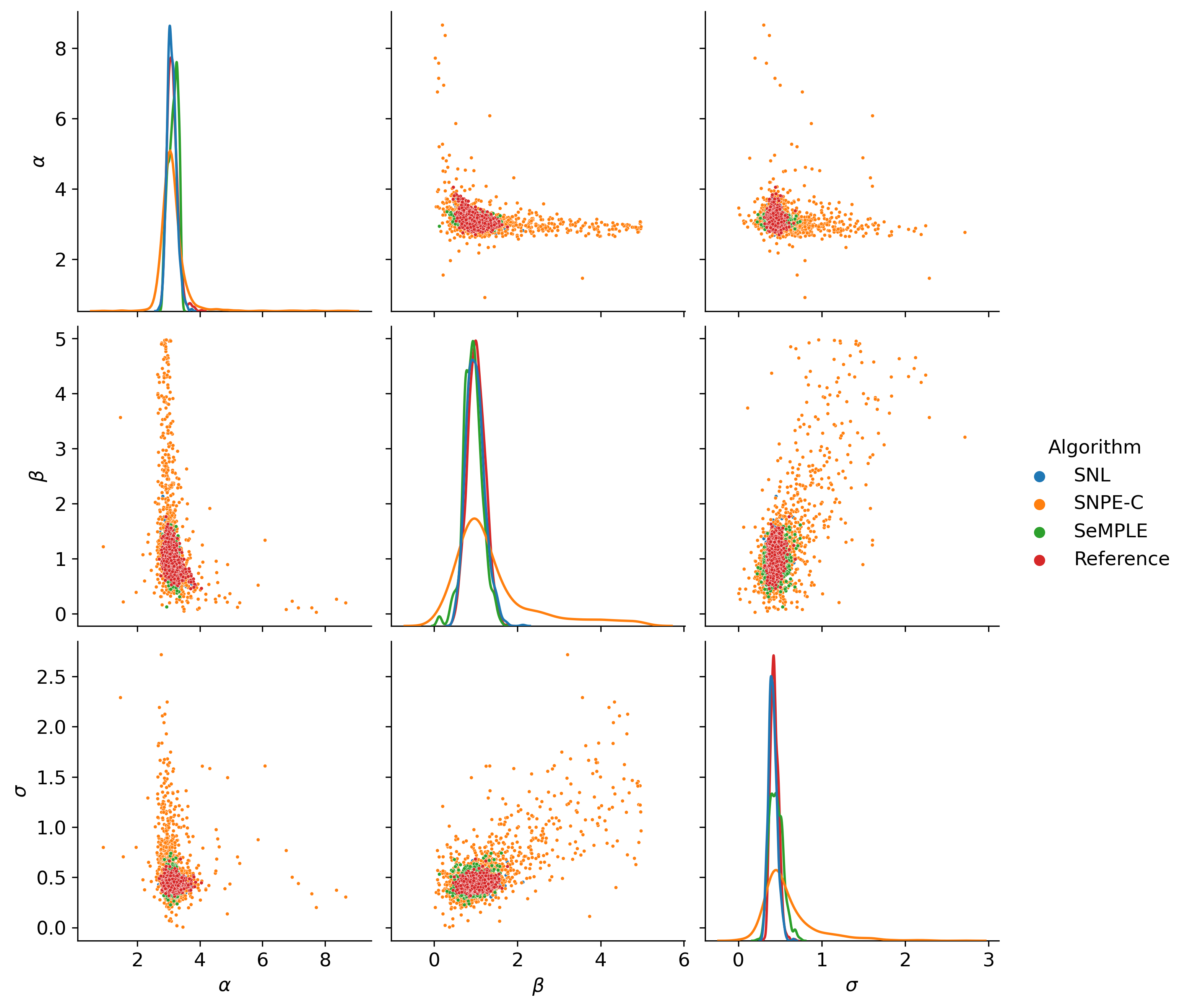}
\caption{Repetition 1.}
\end{subfigure}
\begin{subfigure}[b]{0.45\textwidth}
    \includegraphics[scale=0.27]{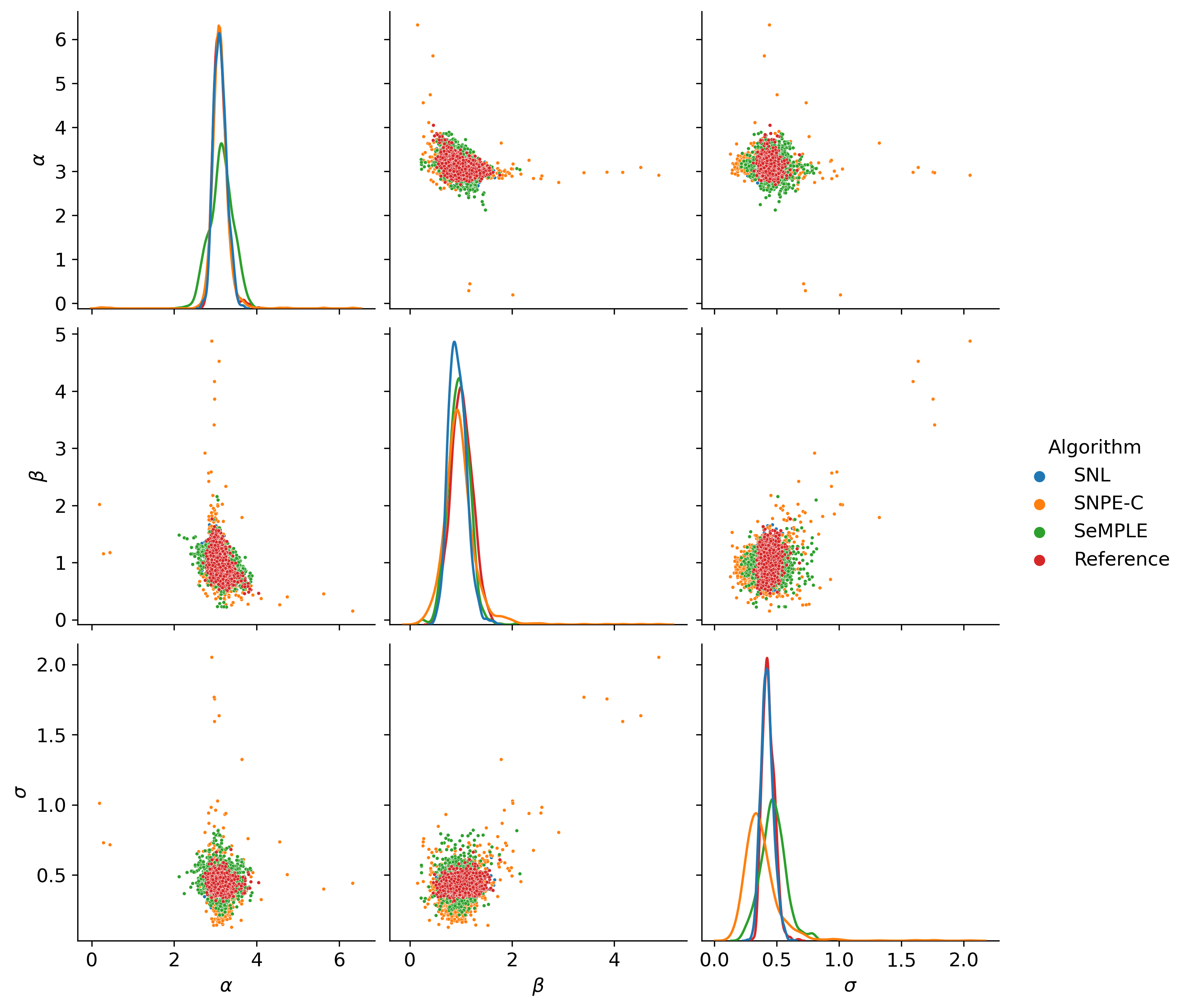}
\caption{Repetition 2.}
\end{subfigure}
\\
\begin{subfigure}[b]{0.45\textwidth}
    \includegraphics[scale=0.27]{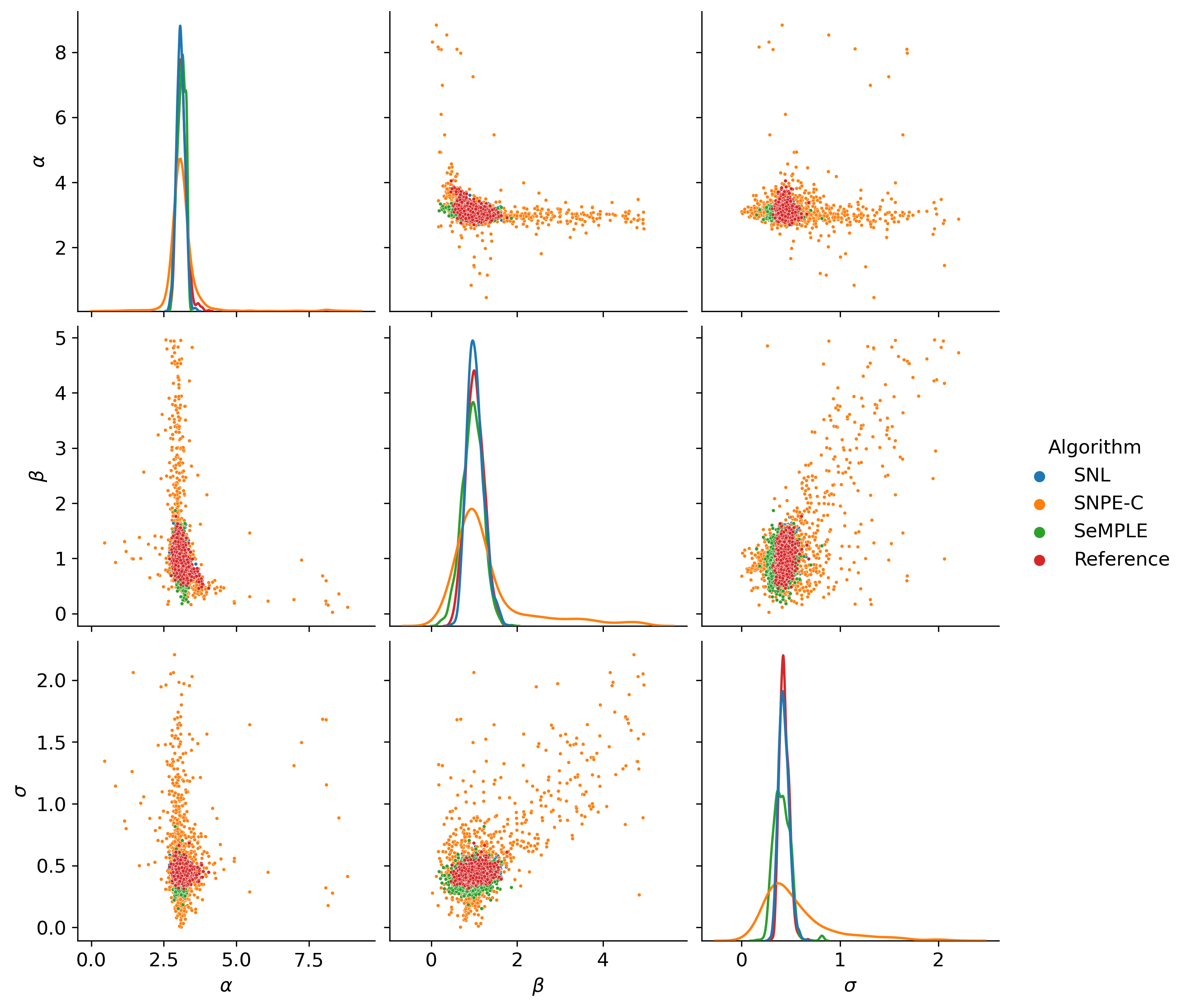}
\caption{Repetition 3.}
\end{subfigure}
\begin{subfigure}[b]{0.45\textwidth}
    \includegraphics[scale=0.27]{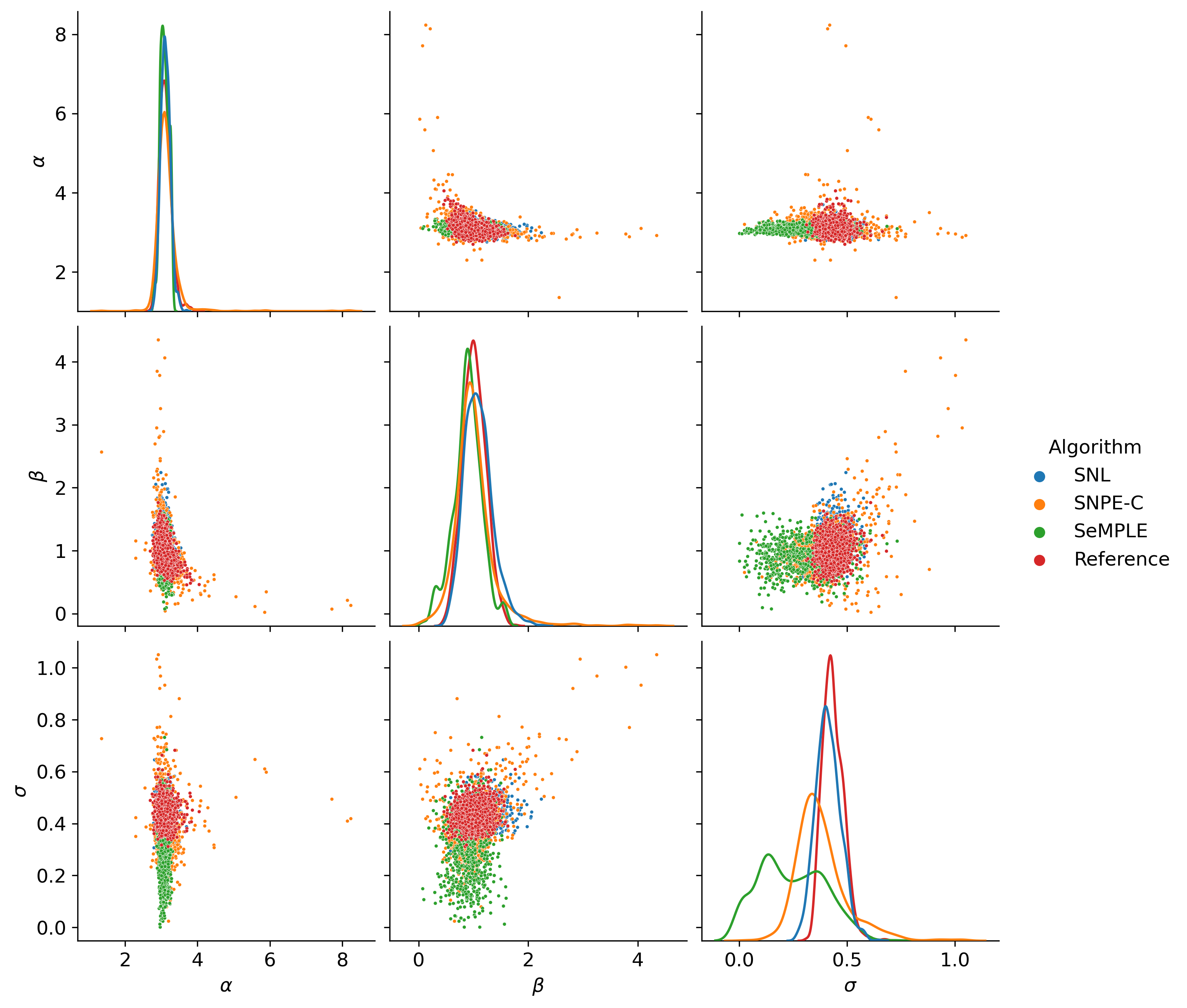}
\caption{Repetition 4.}
\end{subfigure}
\\
\begin{subfigure}[b]{0.45\textwidth}
    \includegraphics[scale=0.27]{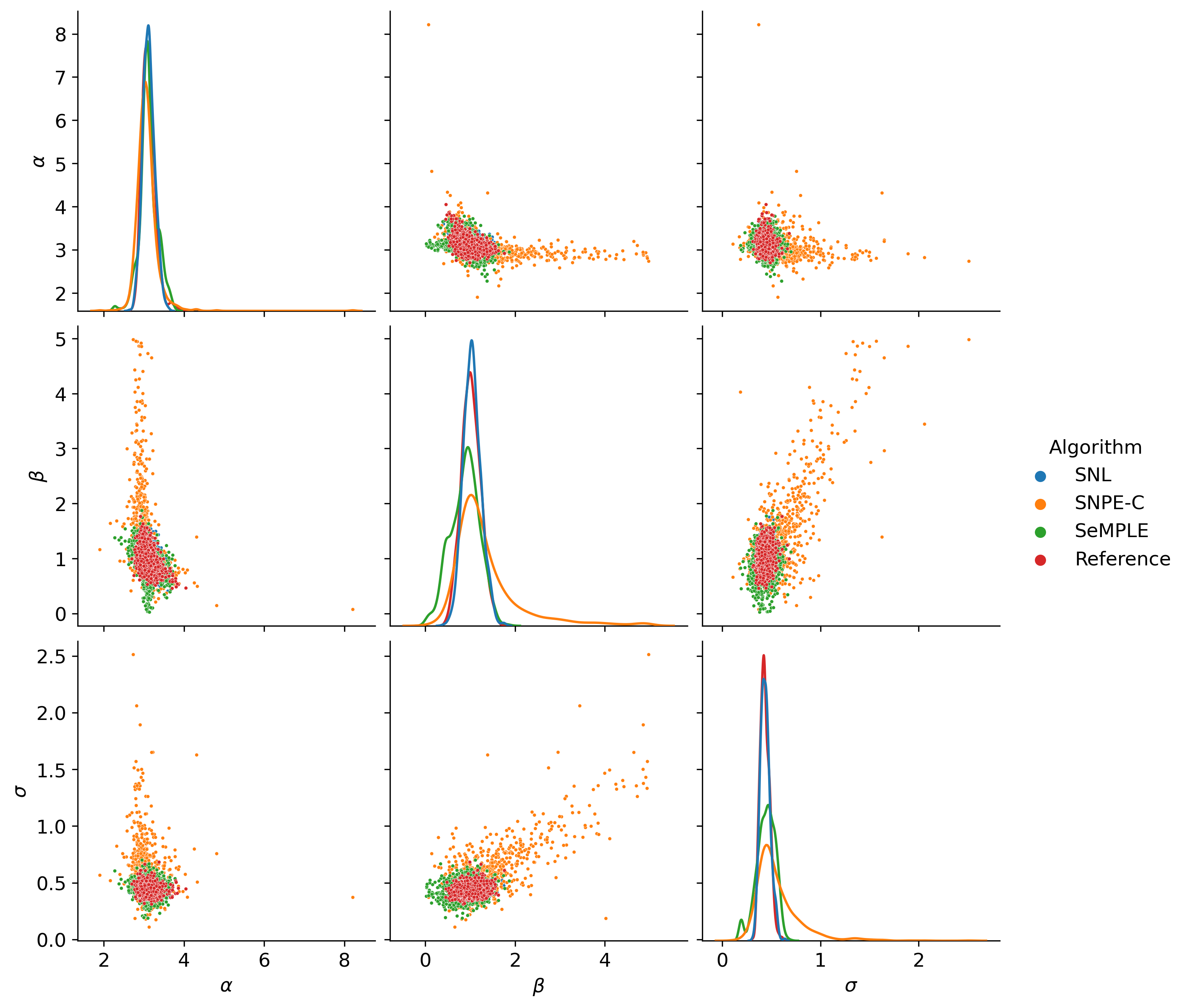}
\caption{Repetition 5.}
\end{subfigure}
\begin{subfigure}[b]{0.45\textwidth}
    \includegraphics[scale=0.27]{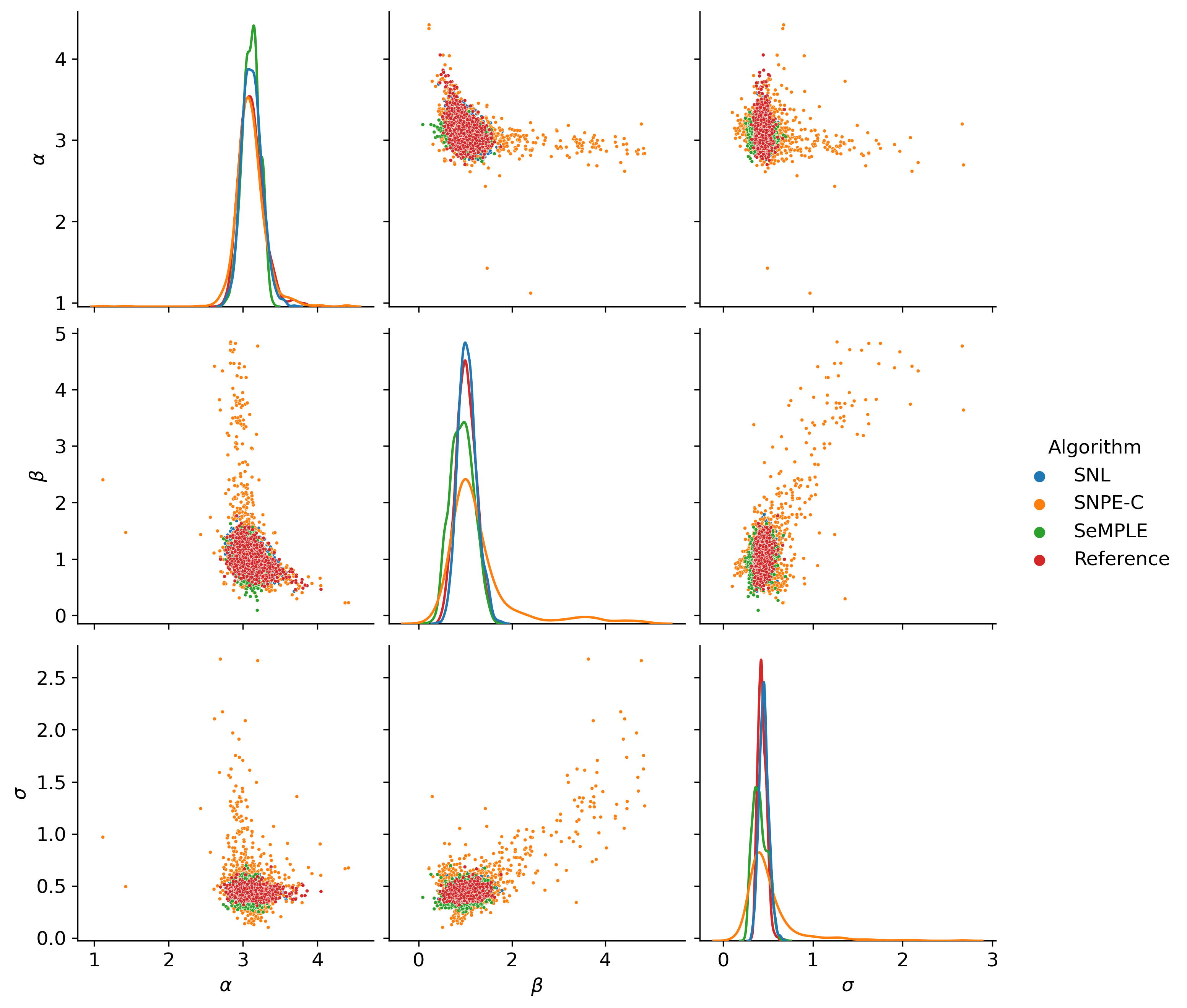}
\caption{Repetition 6.}
\end{subfigure}
    \caption{Ornstein-Uhlenbeck, results when $R=4$ for all algorithms, runs 1--6.: marginal posteriors and posterior draws from the last round (R=4) of each algorithm.}
\label{fig:ornstein_uhlenbeck_pairplot_run_1-6}
\end{figure}

\begin{figure}[h]
    \centering
\begin{subfigure}[b]{0.45\textwidth}
    \includegraphics[scale=0.27]{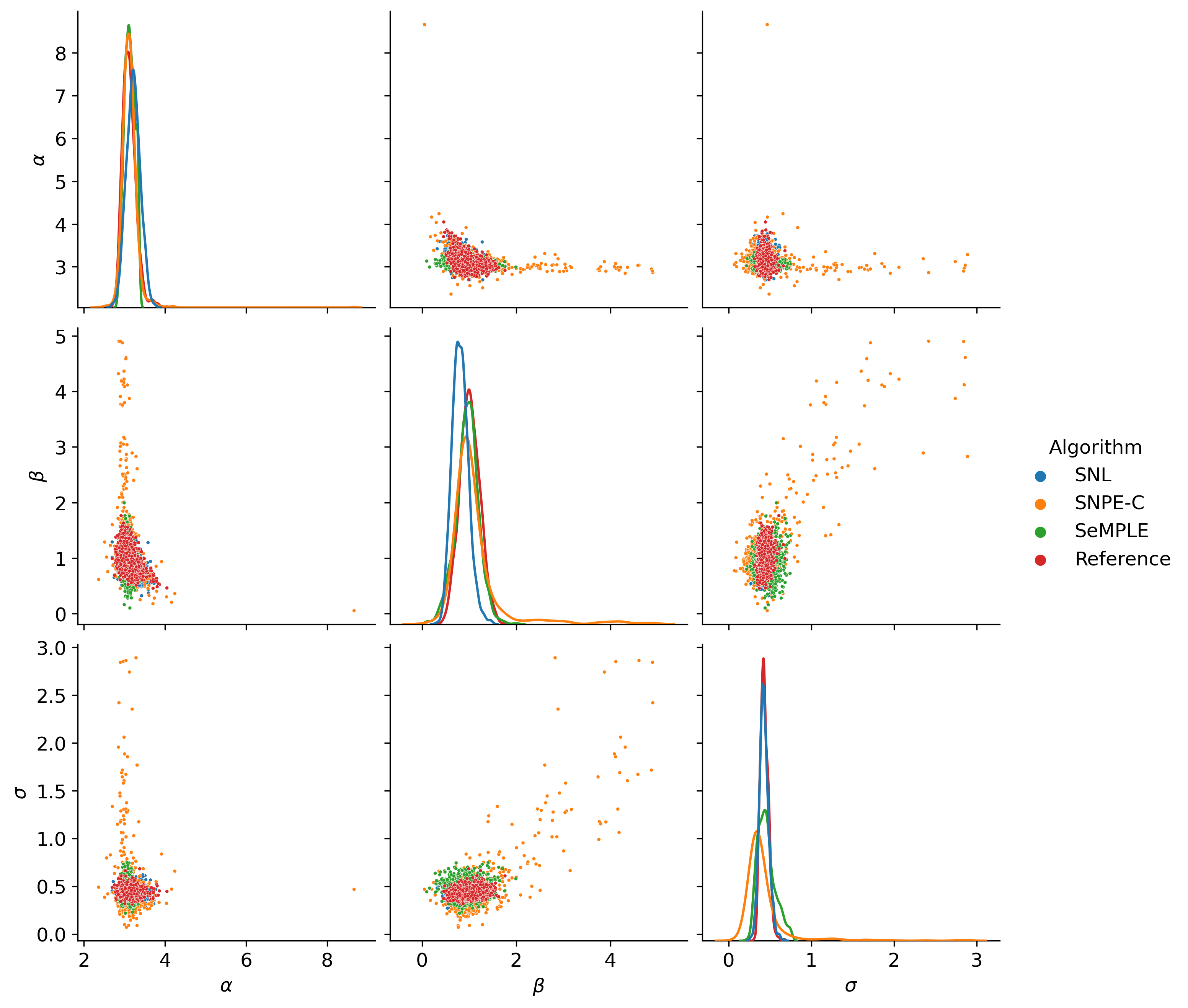}
\caption{Repetition 7.}
\end{subfigure}
\begin{subfigure}[b]{0.45\textwidth}
    \includegraphics[scale=0.27]{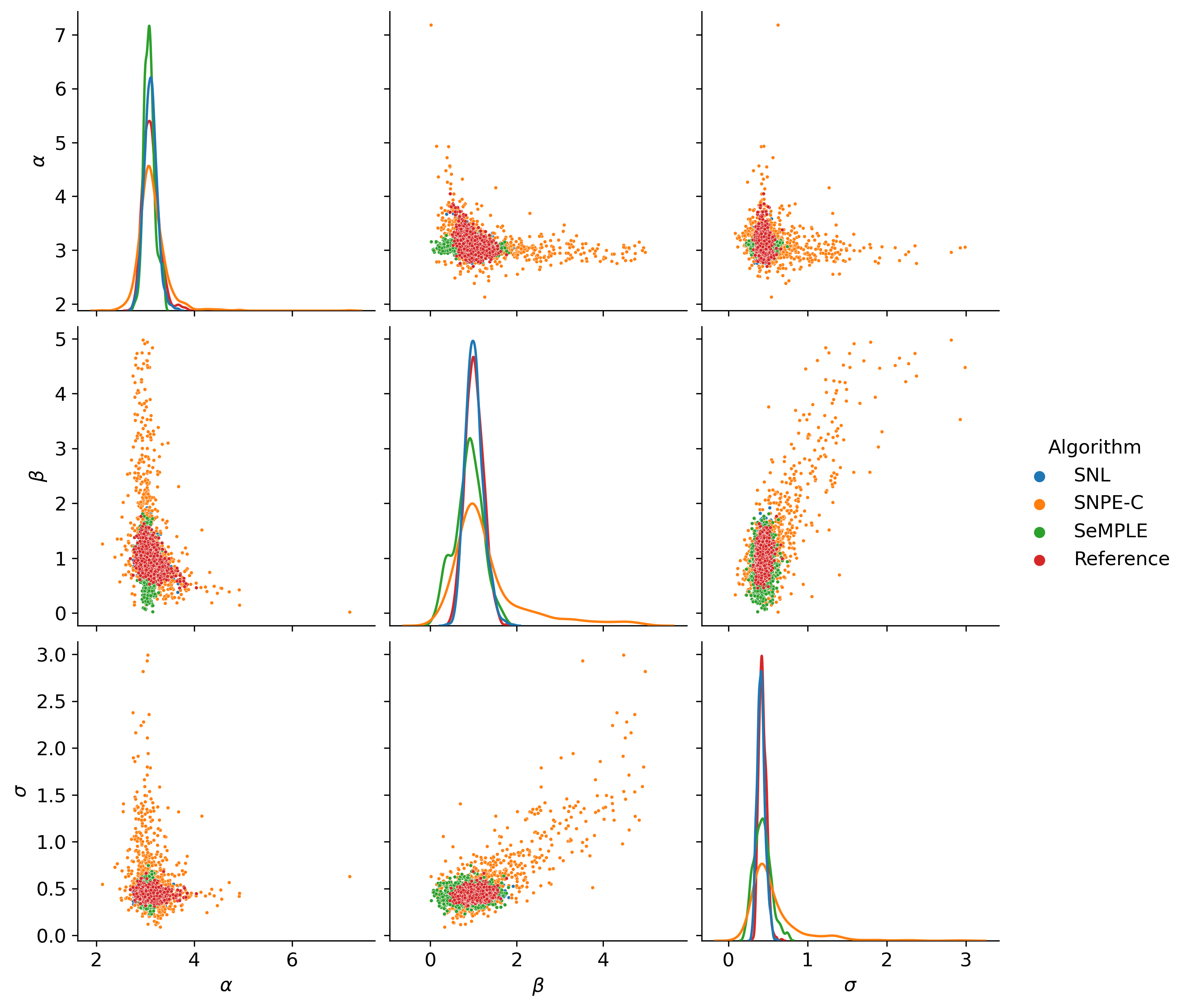}
\caption{Repetition 8.}
\end{subfigure}
\\
\begin{subfigure}[b]{0.45\textwidth}
    \includegraphics[scale=0.27]{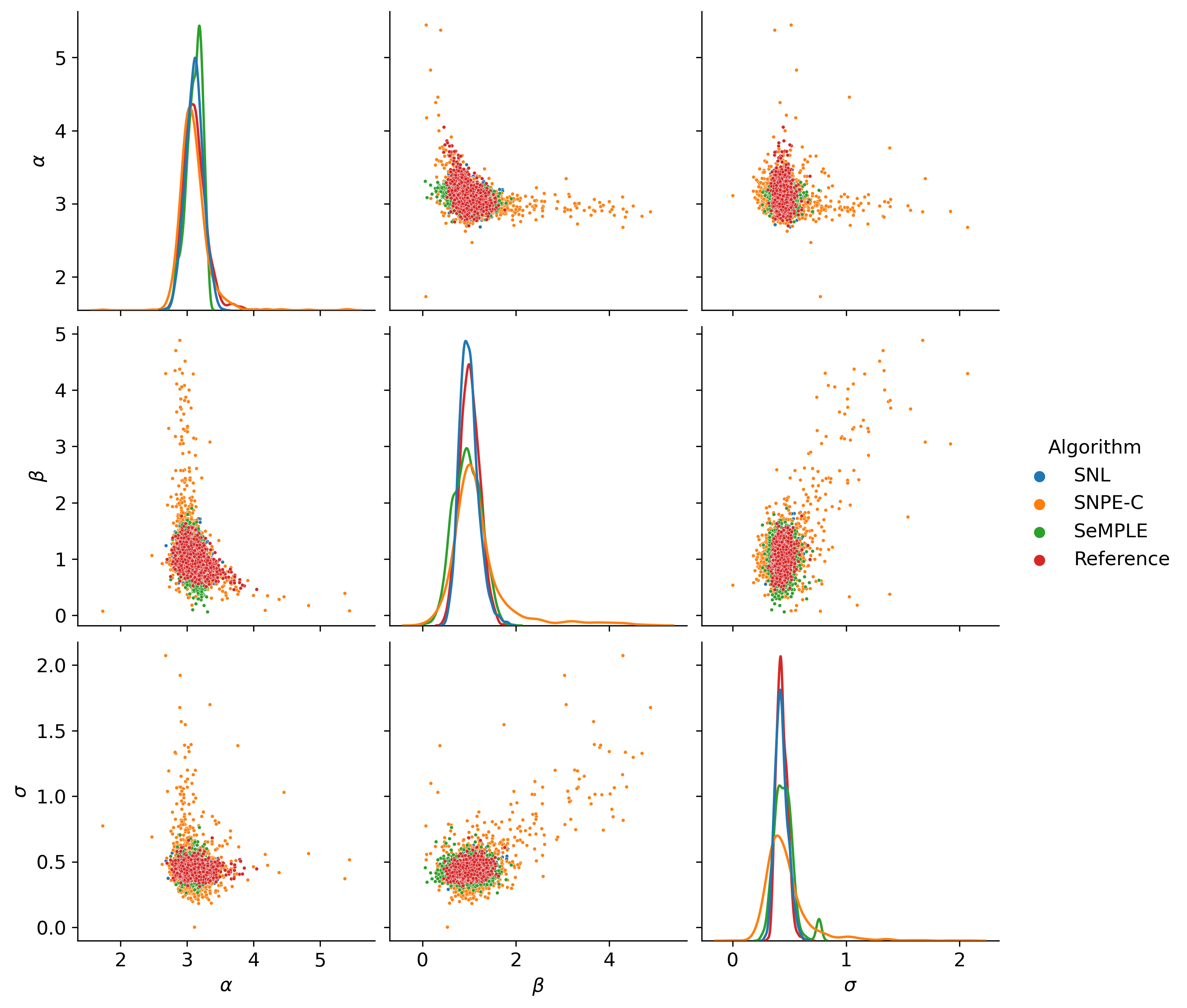}
\caption{Repetition 9.}
\end{subfigure}
\begin{subfigure}[b]{0.45\textwidth}
    \includegraphics[scale=0.27]{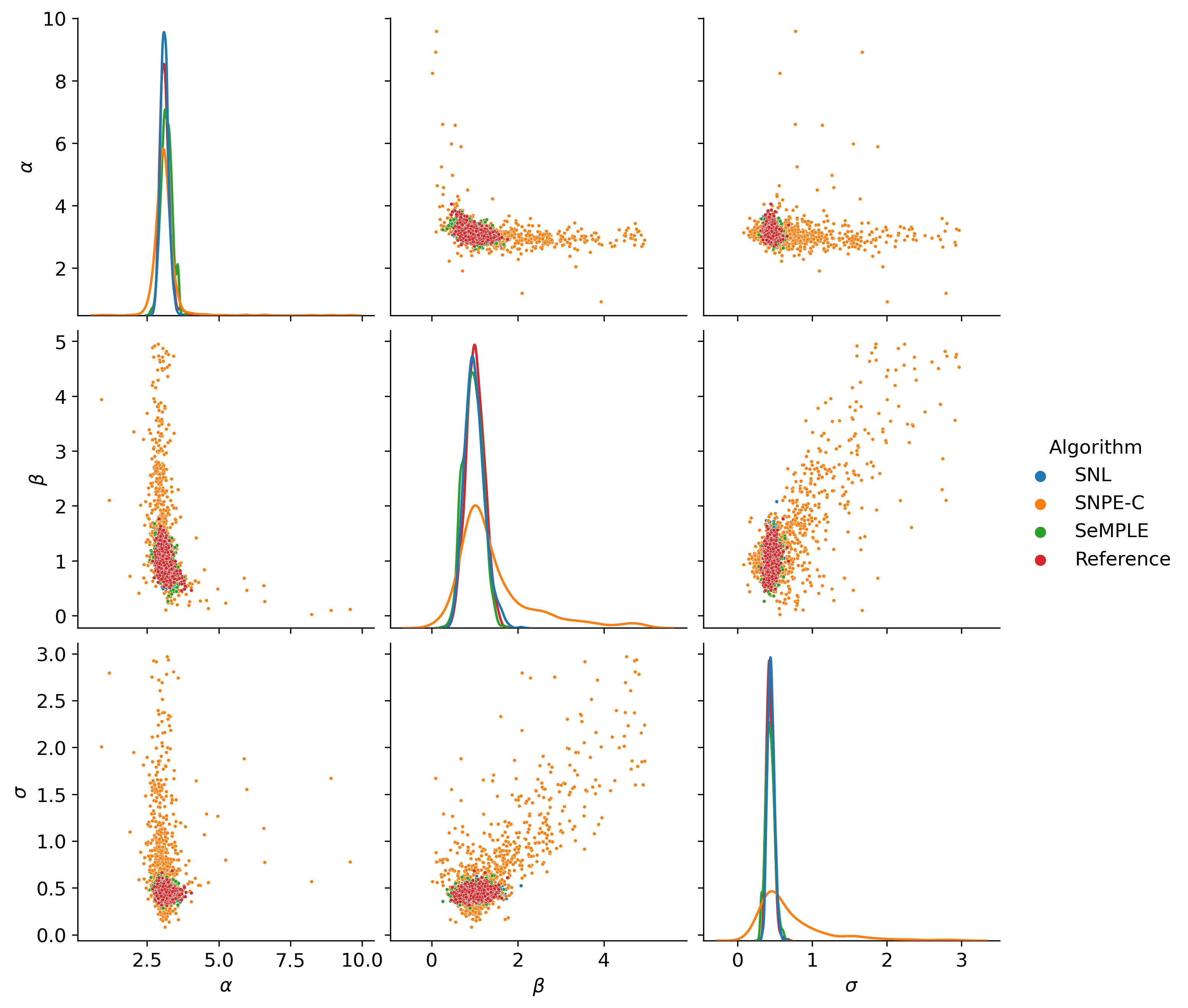}
\caption{Repetition 10.}
\end{subfigure}
    \caption{Ornstein-Uhlenbeck, results when $R=4$ for all algorithms, runs 7-10: marginal posteriors and posterior draws from the last round (R=4) of each algorithm.}
\label{fig:ornstein_uhlenbeck_pairplot_run_7-10}
\end{figure}

\begin{figure}[h]
    \centering
    \includegraphics[width=0.4\textwidth]{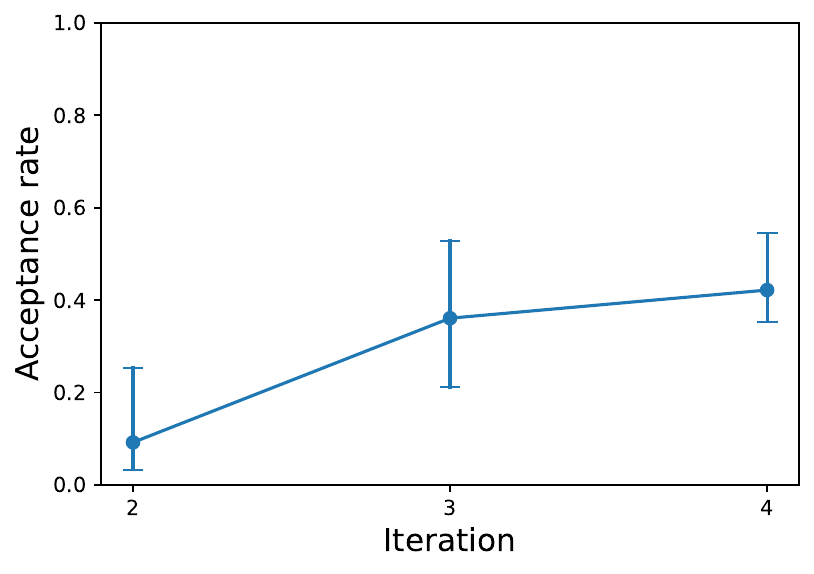}
    \caption{Ornstein-Uhlenbeck: Median acceptance rate of the Metropolis-Hastings step of 10 independent SeMPLE runs with the same data set. Error bars show min/max values.}
    \label{fig:OU_jass_accrate}
\end{figure}

Figure \ref{fig:OU_algorithm_metric_vs_sims} shows the C2ST metric and Wasserstein distance as a function of the number of model simulations for each algorithm. At round r=2 ($20\,000$ model simulations) SNL consistently shows an unexpected behaviour where the posterior sample is clearly worse than in the first round (r=1). This happened in some scenarios with SNL where the number of simulations in the first \textcolor{black}{round} is low enough. Apart from this, SNL scores best in both metrics after the full $40\,000$ model simulations (R=4 algorithm rounds). However, notice that the difference between SeMPLE and SNL is much smaller in terms of the Wasserstein distance. In fact, by looking at the marginal posteriors (Figures \ref{fig:ornstein_uhlenbeck_pairplot_run_1-6}--\ref{fig:OU_algorithm_metric_vs_sims_R=4_R=10}) it is clear that while SNL posteriors better approximate the true posterior, SeMPLE is a strong second best alternative, and provides clearly better posteriors compared to SNPE-C. However the latter result is not apparent when looking at the C2ST metric, where SeMPLE and SNPE-C do not seem strikingly different. Therefore, for this experiment with a unimodal target, similarly to the Bernoulli GLM example, we find that the Wasserstein distance provides a better summary of the quality of the posterior inference.
The Ornstein-Uhlenbeck process could be a simulation model where it is easier to estimate the likelihood compared to the posterior, which could explain the outcome of metric scores since both SNL and SeMPLE uses a surrogate likelihood to obtain posterior samples. 

\begin{figure}[h]
    \centering
    \includegraphics[width=0.4\textwidth]{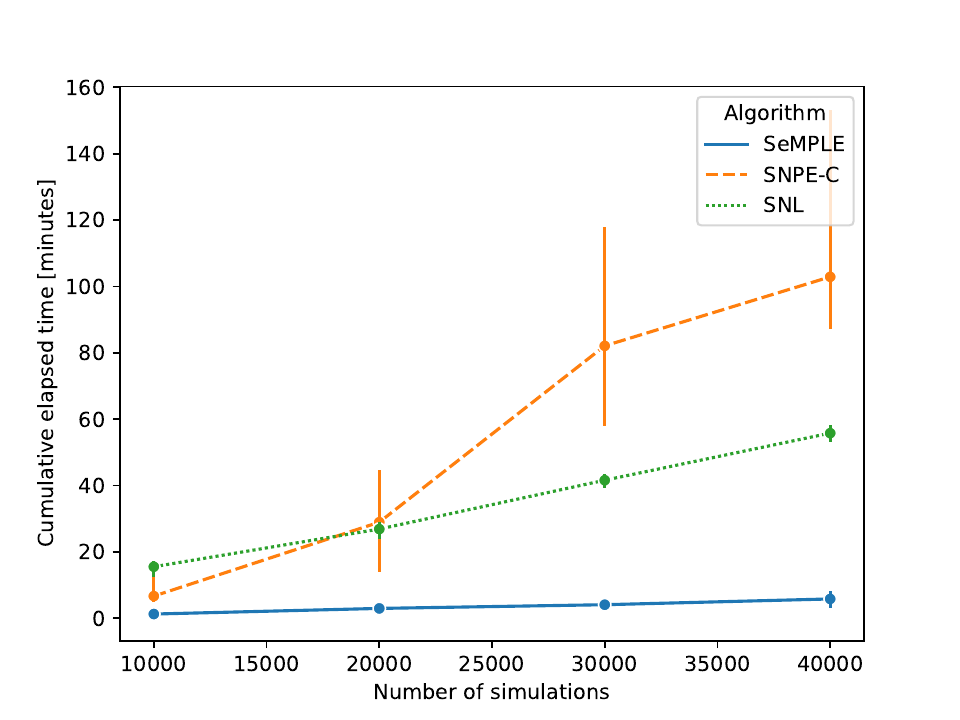}
    \caption{Ornstein-Uhlenbeck, results when $R=4$ for all algorithms: median cumulative runtime of 10 independent runs as a function of number of model simulations. Error bars show min/max values.}
    \label{fig:OU_time_vs_iter}
\end{figure}

In terms of runtime, Figure \ref{fig:OU_time_vs_iter} shows that SeMPLE has a much much smaller runtime compared to both SNL and SNPE-C, even if were to include the 16 minutes of BIC computations into account. Especially SNPE-C has much longer runtimes with this model, combined with the worst performance metric results out of all three algorithms. Plots of the posterior marginals and posterior draws from the last \textcolor{black}{round} (r=4) of each algorithm are in Figures \ref{fig:ornstein_uhlenbeck_pairplot_run_1-6}-\ref{fig:ornstein_uhlenbeck_pairplot_run_7-10}, for each of the 10 independent runs. Acceptance rates of the Metropolis-Hastings step of SeMPLE are in Figure \ref{fig:OU_jass_accrate}, showing that, particularly for the third and fourth round, the acceptance rate is fairly high (compatibly with being an MCMC algorithm).

\subsection{Results when using $R=10$ for SNL/SNPE-C and $R=4$ for SeMPLE}\label{sec:ou_R=4_and_R=10}

\begin{figure}[h] 
    \captionsetup[subfigure]{justification=centering}
     \centering
     \begin{subfigure}[b]{0.39\textwidth}
         \centering
         \includegraphics[width=\textwidth]{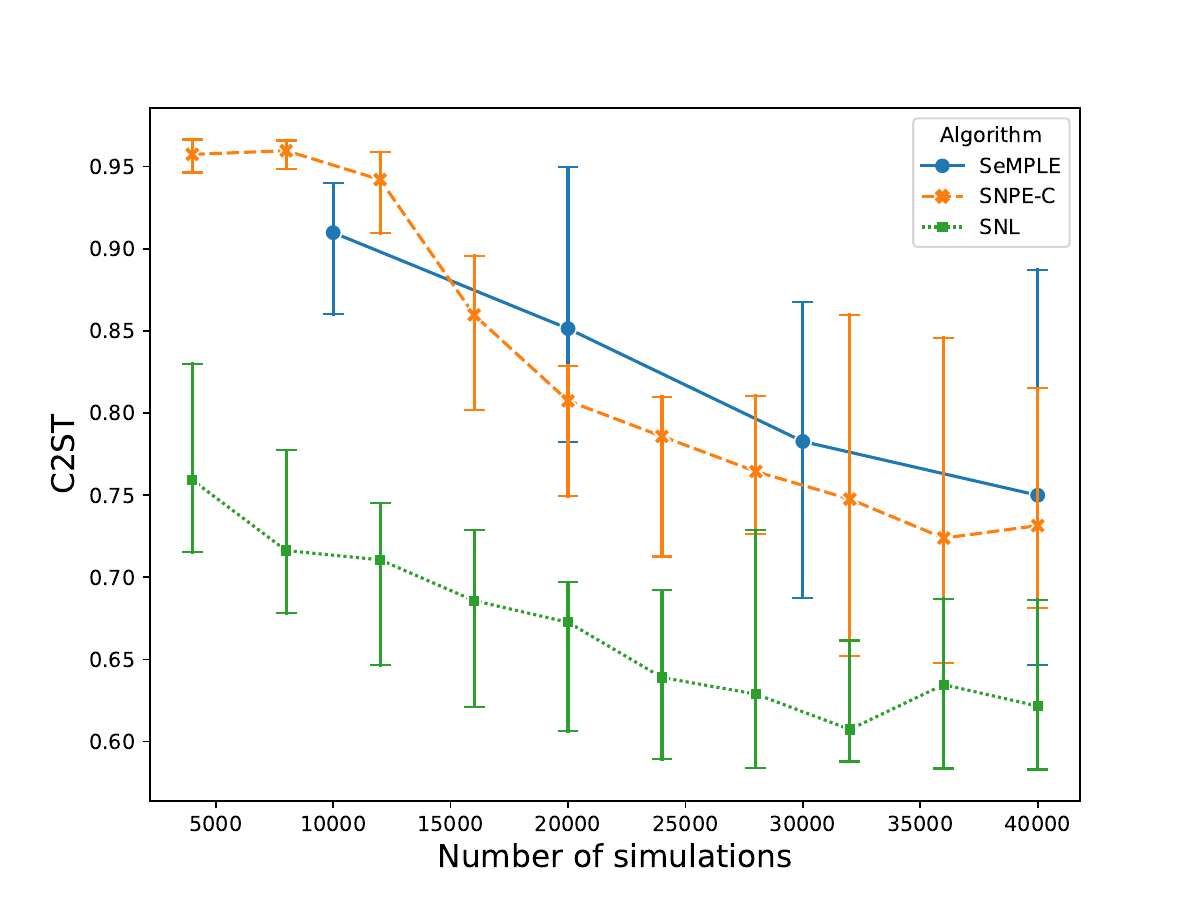}
         \caption{}
     \end{subfigure}
     \begin{subfigure}[b]{0.39\textwidth}
         \centering
         \includegraphics[width=\textwidth]{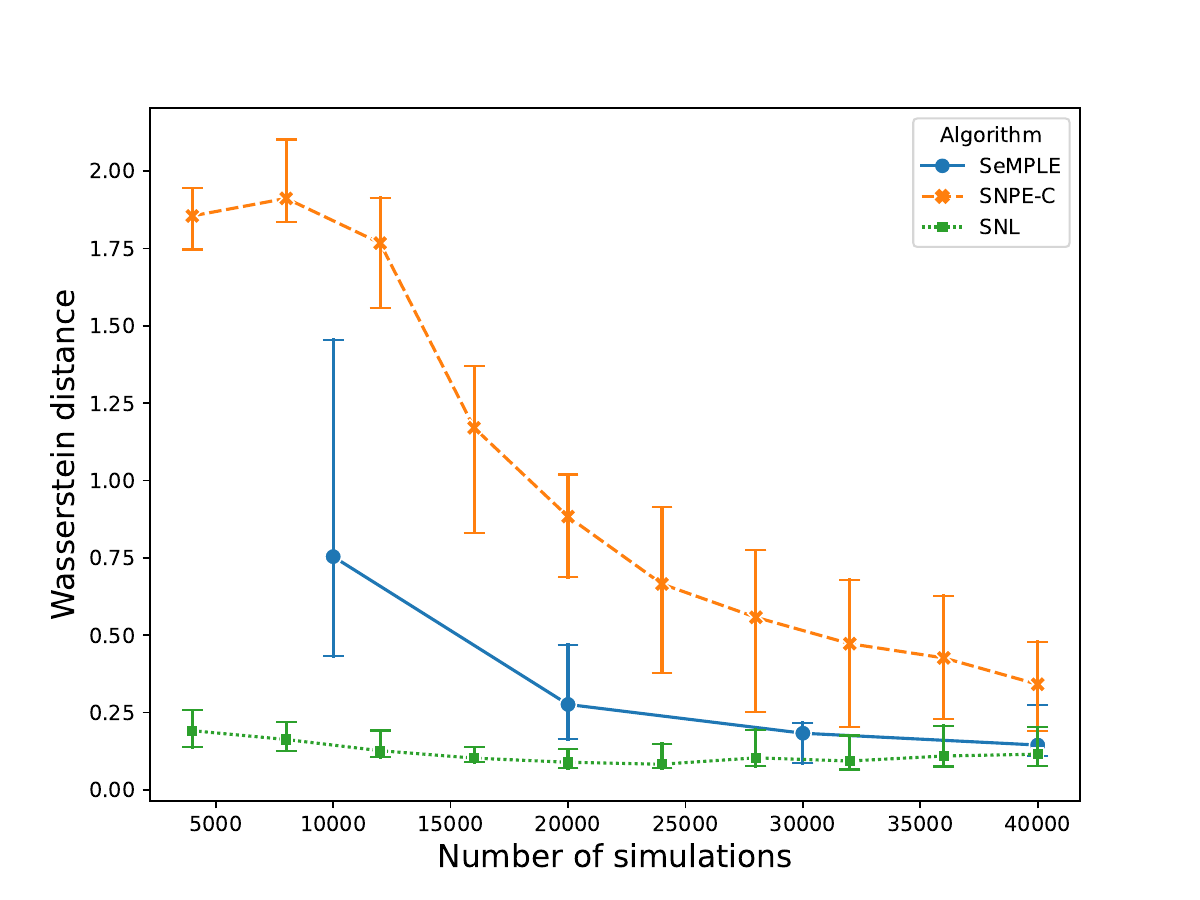}
         \caption{}
     \end{subfigure}
     \caption{Ornstein-Uhlenbeck, results with $R=4$ (SeMPLE) and $R=10$ (SNL/SNPE-C): median C2ST and Wasserstein distance across 10 independent runs (with the same observed data set) as a function of the number of model simulations. Error bars show min/max values. \textbf{For unimodal posteriors (as in this case) Wasserstein distances should be preferred to C2ST.}}
     \label{fig:OU_algorithm_metric_vs_sims_R=4_R=10}
\end{figure}

\begin{figure}[h]
    \centering
    \includegraphics[width=0.4\textwidth]{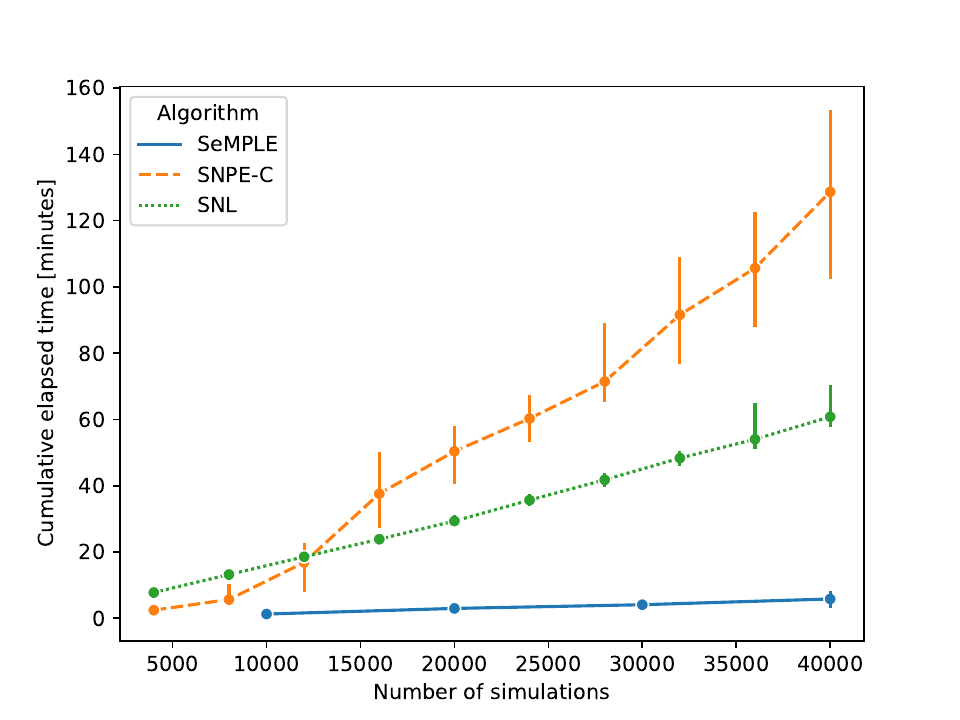}
    \caption{Ornstein-Uhlenbeck, results with $R=4$ (SeMPLE) and $R=10$ (SNL/SNPE-C): median cumulative runtime of 10 independent runs as a function of number of simulations. Error bars show min/max values.}
    \label{fig:OU_time_vs_iter_R=4_R=10}
\end{figure}

We executed the experiment anew, but this time by allocating the $4\times 10^4$ model simulations uniformly across 4 rounds for SeMPLE and 10 rounds for SNL and SNPE-C (as per \texttt{SBIBM} defaults). Results are in Figure \ref{fig:OU_algorithm_metric_vs_sims_R=4_R=10} and Figure \ref{fig:OU_time_vs_iter_R=4_R=10}. As explained in previous sections, for this experiment it is more trustworthy to consider the Wasserstein distances, hence Figure \ref{fig:OU_algorithm_metric_vs_sims_R=4_R=10}(b) in this case. Once again we show that, even by comparing with SNL and SNPE-C trained over many inference rounds (as per \texttt{SBIBM} defaults), SeMPLE is competitive. Moreover, in terms of runtime, if we include the 16 minutes required to run the BIC procedure to select $K$ (recall this is amortized so it can be reused with other observations), the median runtime for SeMPLE is around 21 minutes, while SNPE-C requires around 136 minutes to complete the 10 rounds on average. Hence SeMPLE implies a more than 6-fold acceleration, which becomes around 27-fold if we exclude the BIC procedure.

\clearpage

\section{Lotka-Volterra (Markov jump process with measurement error)}\label{sec:lotka-volterra}
The Lotka-Volterra predator prey model is a population dynamics model with two species. 
Its definition varies, depending on whether it is defined via
an ordinary differential equation (as in \citealp{sbibm}), a continuous time Markov Jump Process (MJP, \citealp{owen2015,owen2015scalable}) or a continuous time  approximation to the MJP given by a stochastic differential equation (as in \citealp{golightly2011bayesian}). Moreover, it could be seen as a model involving three (as in \citealp{owen2015}) or four (as in \citealp{papamakarios2016}) reactions.
We employ a stochastic MJP version of this model as described in \cite{owen2015}, characterized by a set of 3 reactions on the two species, with additional measurement error.

\subsection{Model definition with three reactions}\label{sec:lotka-volterra-model}
Denoting the two species, preys in number $X_{1}$ and predators in number $X_{2}$, evolution of the system is governed by the following three reactions:
\begin{align}
  \centering
  \begin{array}{rccc}
    \mathbf{R}_{1}: & {X}_{1} & \rightarrow & 2{ X}_{1} \\
    \mathbf{R}_{2}: & { X}_{1} + { X}_{2} & \rightarrow & 2{ X}_{2} \\
    \mathbf{R}_{3}: & { X}_{2} & \rightarrow & \emptyset.
  \end{array}
  \label{eq:lvmodel}
\end{align}
These reactions can be interpreted respectively as a prey birth, a predator prey interaction resulting in the death of a prey and a predator birth, and a predator death respectively. Let $X_t = (X_{1,t}, X_{2,t})$ denote the number of both species at time $t$. Each reaction $\mathbf{R}_i$ is assumed to have an associated constant $\theta_i$ and hazard function $h_i(X_t, \theta_i)$ which gives the propensity for a reaction event of type $i$ at time $t$ to occur. Reaction events are dependent on the current state of the system as well as the reaction rate parameters. Hence the  evolution of species counts corresponds to a Markov process on a discrete state space. This reaction network is summarised by its stoichiometry matrix, $S$ and hazard function $h(\mathbf{X},\theta)$:
\begin{align}
  S &= \left(
    \begin{array}{rrr}
      1 & -1 & 0 \\
      0 & 1 & -1
    \end{array} \right), &
  h(\mathbf{X},\theta) &= (\theta_{1}X_{1}, \theta_{2}X_{1}X_{2}, \theta_{3}X_{2}).
\end{align}

\subsection{Inference setup}\label{sec:lotka-volterra-setup}
To simulate the exact dynamics, we use the \texttt{smfsb} R package \citep{smfsb}, implementing the ``Gillespie algorithm'' \citep{gillespie1977}. Whilst the Gillespie algorithm allows exact simulation of the time and type of each reaction event that occurs, observed data $\mathbf{Y}= (y_0, y_1, \dotsc, y_T )$ are typically noisy observations at discrete time intervals, where each $y_t$ is 2-dimensional with $y_t^j \sim \pi(\dot|X_{j,t}, \sigma)$ {\it i.i.d.} for $j=1,2$. We corrupt
each $X_{j,t}$ with {\it i.i.d.}  Gaussian error with mean 0 and variance $\sigma^2$, $\pi(y_t^j|X_{j,t}, \sigma^2) \sim \mathcal{N} (X_{j,t}, \sigma^2)$. The true parameter values generating the observed data is set to $\btheta=(1, 0.005, 0.6)$ and the starting values $X_0 = (50, 100)$ are used throughout.  When generating the observed data we let $\sigma=30$ and later infer $\sigma$ along with the model parameters $\btheta$, resulting in a total of 4 parameters.

When computing model realisation with the Gillespie algorithm, we let $T=30$ and observe the Markov jump process at uniformly spaced intervals with step length $0.2$, resulting in a time series of length $150$ for each species. After corrupting the time series data with the previously described Gaussian error we compute a total of 9 summary statistics that is used when performing the inference. These summary statistics, first suggested in \cite{summariesABC}, consist of the mean, (log) variance, and the lag 1 and lag 2 two auto-correlations of the time series of each species. Additionally the cross-correlation between the two time series is included in the summary statistic.

Following the setup in \cite{owen2015}, we place uniform prior information on $\log(\theta)$,
\begin{equation}
  \label{eq:lvprior}
  \log(\theta_{i}) \sim \mathcal{U}(-6,2), \quad i = 1,2,3.
\end{equation}
and $\log(\sigma)$
\begin{equation}
  \label{eq:lvprior2}
  \log(\sigma) \sim \mathcal{U}(\log(0.5),\log(50)).
\end{equation}

Prior to running the inference, for each of the nine summary statistics we computed trimmed means, where    $1.25\%$ of the values were trimmed from each end and corresponding trimmed standard deviations, obtained from 10,000 prior predictive simulations, and we used these to standardise both the observed summaries and the summaries simulated during the inference.  
For each SeMPLE run we set a budget of $30\,000$ model simulations, uniformly distributed across 3 algorithm rounds. The number of mixture components is set to $K=15$ (a number which was picked without any specific tuning) and we set the covariance matrices in GLLiM to be full unconstrained matrices, all different across the $K$ components. Moreover, we do not inflate the covariance in the Metropolis-Hastings step (ie we set $\gamma=1$). The mixture probability threshold to remove negligible mixture components during the SeMPLE run was set to $0.005$.

For a MJP Lotka-Volterra model, we did not find a ready implementation in \texttt{SBIBM} (there, an ODE-based Lotka-Volterra model is provided, not a MJP version), and this is likely because with such a model it is impossible to perfectly control the computational effort, as for a given value of $\theta$ the time spent to complete a trajectory is random. 
For this reason, we found more practical to compare SeMPLE with approximate Bayesian computation (ABC) experiments using the code from \cite{picchini2022guided}. For SMC-ABC, we used the same observed data and the same standardised summary statistics as with SeMPLE. 
We consider sequential Monte Carlo ABC (SMC-ABC), and also use the posterior from SMC-ABC as ``reference posterior''. SMC-ABC is the state-of-art approach to ABC inference and is the algorithm of choice in recent inference platforms such as \texttt{ABCpy} (\citealp{dutta2017abcpy}) and \texttt{pyABC} (\citealp{pyabc}). We consider two versions of SMC-ABC that differ in terms of proposal function: the first one is the ``classic'' SMC-ABC, in that it uses as proposal distribution a Gaussian random walk of the type $\theta^{**}\sim N(\theta^*,2\Sigma)$ (\citealp{beaumont2009}, \citealp{filippi2013optimality}), where $\theta^*$ is a parameter ``particle'' randomly sampled from the most recent set of ``parameter particles'' (these providing an ABC approximation to the posterior), and $\Sigma$ is the empirical weighted covariance matrix of these ``parameter particles''. This first sampler is named \texttt{standard} in \cite{picchini2022guided}, as it is the most common SMC-ABC proposal sampler, as found in most literature and in available software packages. A second SMC-ABC sampler is the one denoted \texttt{blockedopt} in \cite{picchini2022guided}, which is also a Gaussian proposal sampler but not of random-walk type, and is much more resource-effective, as it more rapidly ``guide'' the parameters towards the bulk of the posterior region (see \citealp{picchini2022guided} for details). We ran both SMC-ABC versions with 1,000 particles.

\subsection{Results}\label{sec:lotka-volterra-results}

\begin{figure}[htbp]
    \centering
    \includegraphics[width=0.8\textwidth]{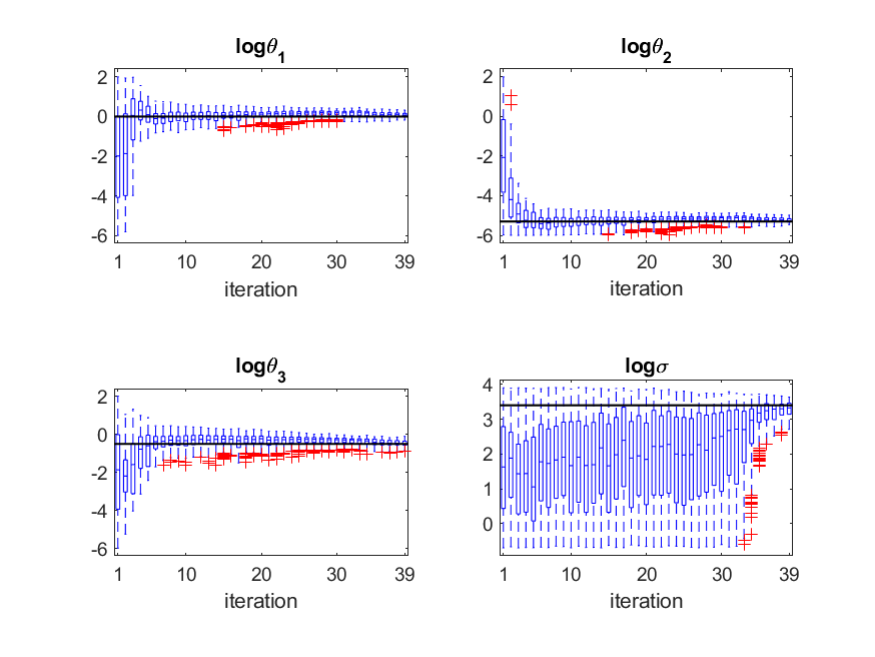}
    \caption{Lotka-Volterra: evolution of ABC posteriors using the ``guided'' \texttt{blockedopt} SMC-ABC across 39 rounds and 3,337,640 model simulations. Ground-truth parameters are marked with horizontal lines.}
    \label{fig:lv_abc_blocked_long}
\end{figure}

\begin{figure}[htbp]
    \centering
    \includegraphics[width=0.8\textwidth]{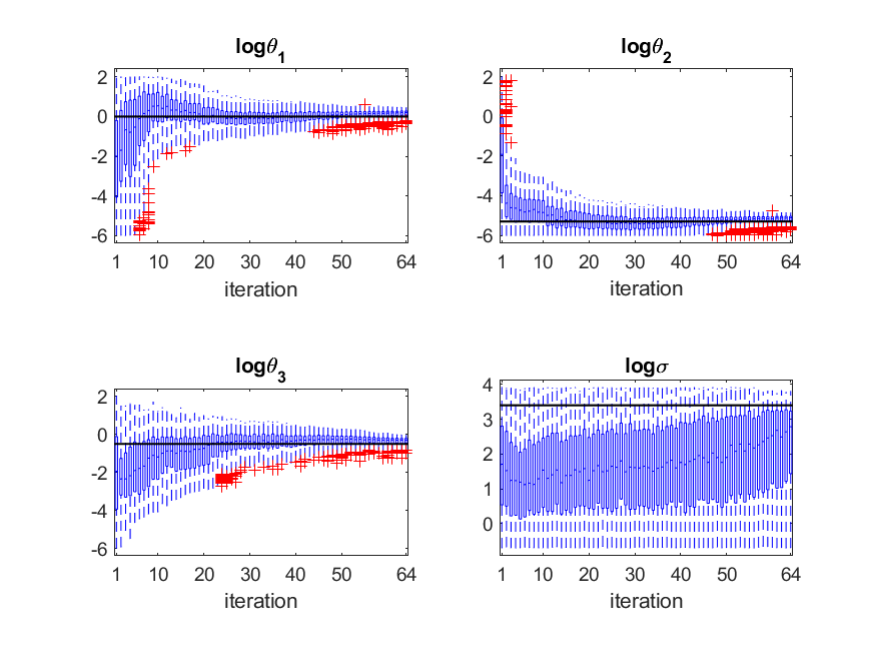}
    \caption{Lotka-Volterra: evolution of ABC posteriors using \texttt{standard} SMC-ABC across 64 rounds and 3,110,349 model simulations. Ground-truth parameters are marked with horizontal lines.}
    \label{fig:lv_abc_standard_long}
\end{figure}

 For this case study, we consider the posterior returned by the \texttt{blockedopt} SMC-ABC as ``reference posterior'', as it is obtained after more than three million model simulations (specifically 3,337,640 simulations), that is a computational budget more than 100 times larger than the budget used for SeMPLE. 
In Figure \ref{fig:lv_abc_blocked_long} we show that, while the ``guided'' \texttt{blockedopt} SMC-ABC can infer $(\theta_1,\theta_2,\theta_3)$ for  a smaller budget than millions of simulations, however inferring also $\sigma$ does requires many additional simulations. When we use the \texttt{standard} SMC-ABC, results after 3,110,349 model simulations are in Figure \ref{fig:lv_abc_standard_long}, and it is clear that the variability in the inference is much larger, and in particular \texttt{standard} struggles in inferring $\sigma$.

\begin{figure}[htbp]
    \centering
    \includegraphics[width=0.5\textwidth]{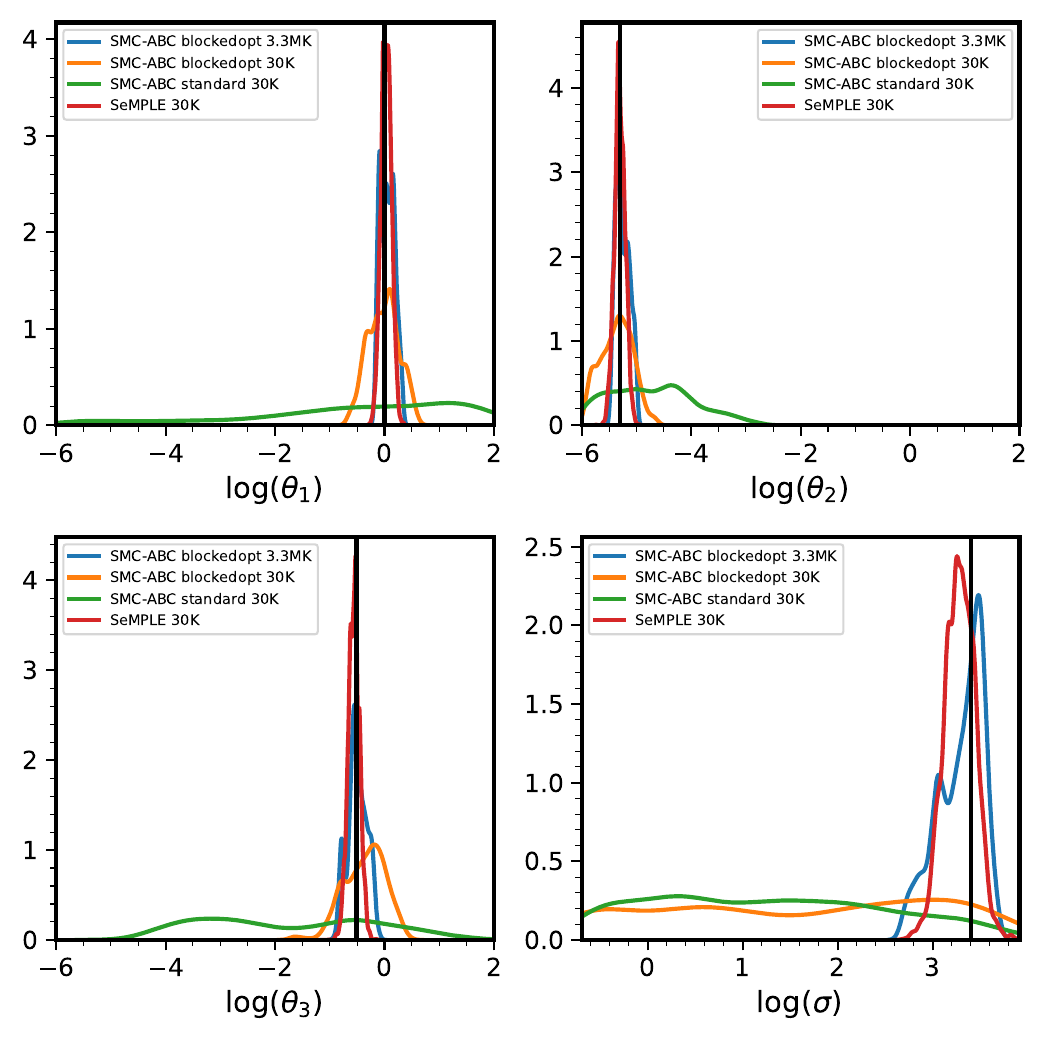}
    \caption{Lotka-Volterra: marginal posterior plots from SMC-ABC and SeMPLE. The SeMPLE posterior sample is based on 10,000 posterior draws, and was returned after a total of 30,000 model simulations. The \texttt{blockedopt} SMC-ABC posterior samples from \textcolor{black}{round} 10 and 39 used a total 32,865 and 3,337,640 model simulations respectively. The posterior samples from \texttt{standard} SMC-ABC used a total of 30,436 model simulations.}
    \label{fig_supp:lv-kde-obs2}
\end{figure}

It is of interest to compare results for SeMPLE and SMC-ABC for a similar number of model simulations. We anticipate to observe something that is well known already from \cite{greenberg2019,sbibm}, namely the SMC-ABC inefficiency in terms of model simulations. What we mostly wish to verify is the ability for SeMPLE to return accurate inference. 
In Figure \ref{fig_supp:lv-kde-obs2} we compare results when using around 30,000 model simulations, however notice that it is impossible to compare the methods for the exact same budget, as SMC-ABC employs while-loops that keep repeating model simulations until a prefixed number of parameter proposals is accepted. It is clear that when the number of model simulations is limited to around 30,000, neither of the SMC-ABC versions can produce accurate inference. The SMC-ABC methods struggle especially with estimating the noise parameter $\sigma$ when the number of model simulations is limited. The SeMPLE posterior after 30,000 model simulations is more similar to the \texttt{blockedopt} SMC-ABC posterior obtained after approximately 3.3 million model simulations, showing the computational efficiency of SeMPLE with a much smaller simulation budget. 

\begin{figure}[htbp]
    \centering
    \includegraphics[width=0.5\textwidth]{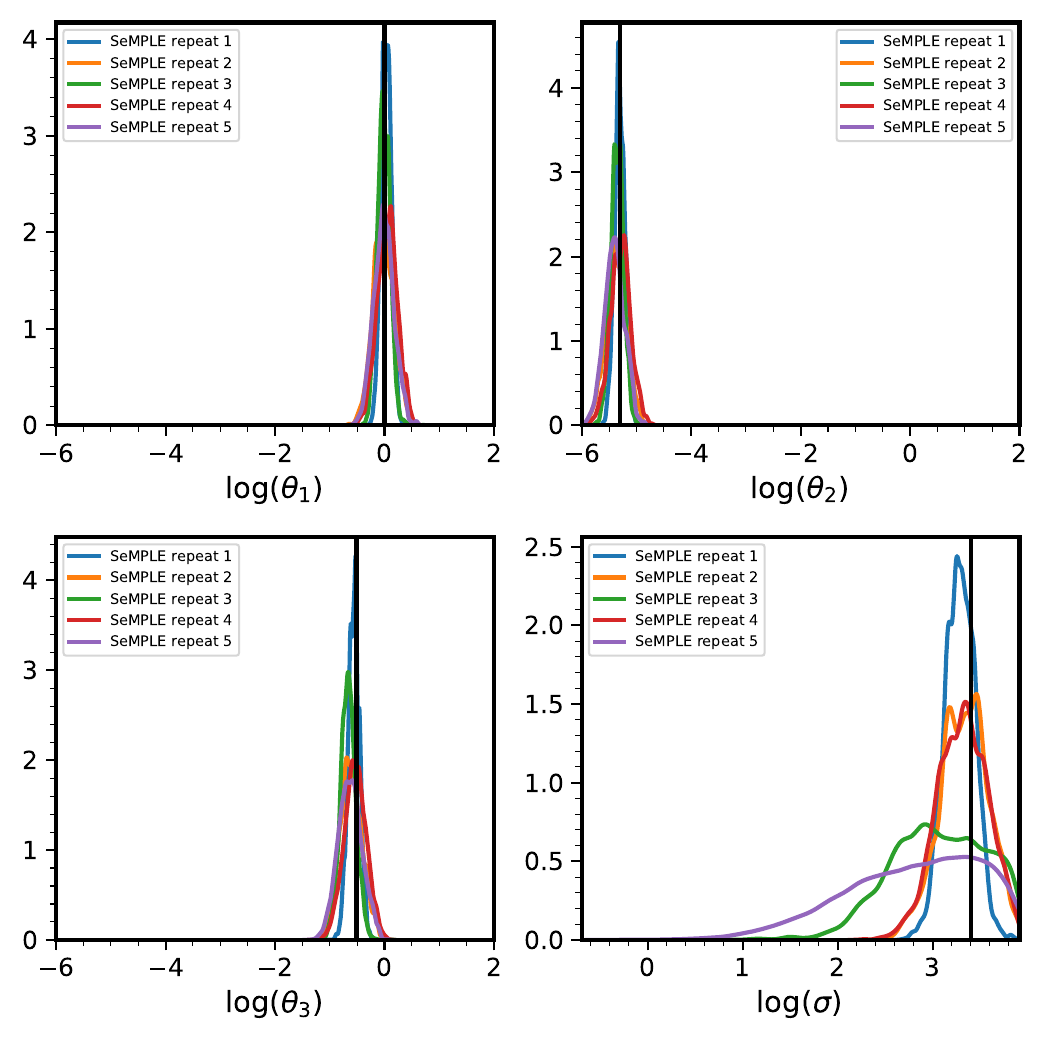}
    \caption{Lotka-Volterra: marginal posterior plots from 5 repeated runs of SeMPLE using the same observed data. The posterior sample is based on 10,000 posterior draws, and was returned after a total of 30,000 model simulations. The x-axes are adjusted to the support of the uniform prior of each parameter.}
    \label{fig:lv_kde_5repeat}
\end{figure}

To give an idea of the variation in the SeMPLE posterior sample output, in Figure \ref{fig:lv_kde_5repeat} we report 5 independent SeMPLE runs with the same observed data as in previous analyses. We see that SeMPLE consistently provides accurate inference of the model parameters $\btheta$ in multiple repetitions. There is a larger variation in terms of the noise parameter $\sigma$, suggesting that this is more difficult to infer with the limited budget of model simulations. However, when comparing with the SMC-ABC posterior distributions for $\sigma$, see  Figure \ref{fig_supp:lv-kde-obs2}, the SeMPLE results are much more informative for the given model simulation budget of 30,000.

\vfill

\clearpage

\clearpage
\section{Twisted prior model with 20 parameters}\label{sec:twisted}

This case study is merely illustrative of the ability of SeMPLE to rapidly return sensible inference for a posterior with highly correlated components. Unlike with other examples, here we did not perform a full comparison with other methods, as we found it difficult to specify the non-standard prior used here to work within the framework provided by \texttt{SBIBM}. 

\subsection{Model definition}\label{sec:twisted-model}

We now consider a model with 20 parameters and a challenging posterior characterised by a strong correlation between some of the parameters. This case study is particularly interesting, as the likelihood only provides location information about the unknown parameter while the dependence structure in the posterior comes mostly from the prior. For this reason, the posterior dependence
changes direction depending on whether the likelihood locates the posterior in the left or right tail of
the prior (see \citealp{nott} for a graphical illustration). This case study was analysed in an ABC context in \cite{li2017extending} and \cite{picchini2022guided}.  The model assumes $\by=(y_1,...,y_{d_\theta})$ drawn from a $d_\theta$-dimensional Gaussian $\by\sim \mathcal{N}(\btheta,\bm{\Psi})$, with $\btheta=(\theta_1,...,\theta_{d_\theta})$ and diagonal covariance matrix $\Psi=\mathrm{diag}(\sigma_0,...,\sigma_0)$. The prior is the ``twisted-normal'' prior of \cite{haario1999adaptive},  with density function proportional to
$
p(\btheta)\propto \exp\biggl\{-\theta_1^2/200-(\theta_2-b\theta_1^2+100b)^2/2-\mathbbm{1}_{\{d_\theta>2\}}\sum_{j=3}^{d_\theta}\theta_j^2/2\biggr\},
$
where $\mathbbm{1}_B$ denotes the indicator function of the set $B$.  This prior is essentially a product of independent Gaussian distributions with the exception that the component for $(\theta_1,\theta_2)$ is modified to produce a ``banana shape'', with the strength of the bivariate dependence determined by the parameter $b$.  Simulation from  $p(\btheta)$ is achieved by first drawing $\btheta$ from a $d_\theta$-dimensional multivariate Gaussian as $\btheta\sim \mathcal{N}(\bm{0},\bm{A})$, where $\bm{A}=\mathrm{diag}(100,1,...,1)$, and then placing the value $\theta_2+b\theta_1^2-100b$ in the slot for $\theta_2$. We set $b=0.1$, a value inducing a strong correlation in the prior between the first two components of $\btheta$.  We consider $d_\theta=20$, i.e. both $\by$ and $\btheta$ have length 20, and (same as \citealp{li2017extending,picchini2022guided}) we set $\sigma_0=1$ as fixed and known, with observations given by the 20-dimensional vector $\by_o=(10,0,...,0)$, where only the first entry is non-zero.

\subsection{Results}\label{sec:twisted-results}

\begin{figure}[h]
\centering
\begin{subfigure}[b]{0.49\textwidth}\includegraphics[width=1\textwidth]{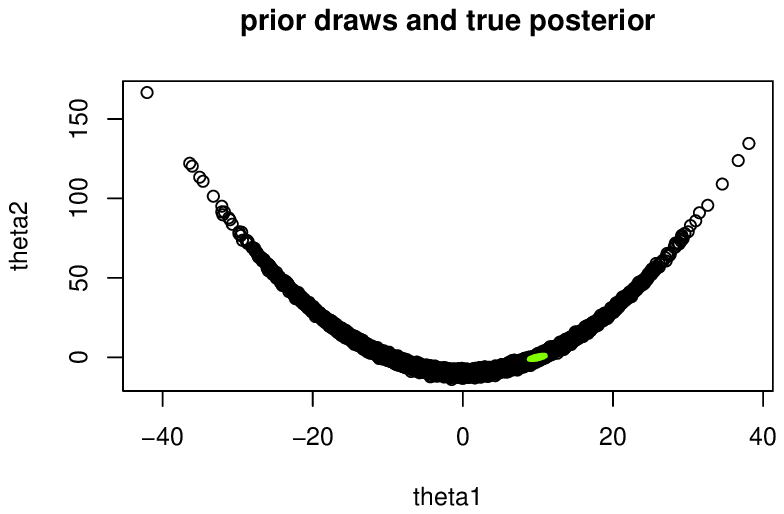}
\caption{Draws from prior $p(\theta_1,\theta_2)$ (black) and exact posterior (green)}\label{fig:twisted_prior_exact}
\end{subfigure}
    \begin{subfigure}[b]{0.49\textwidth}
\includegraphics[width=1\textwidth]{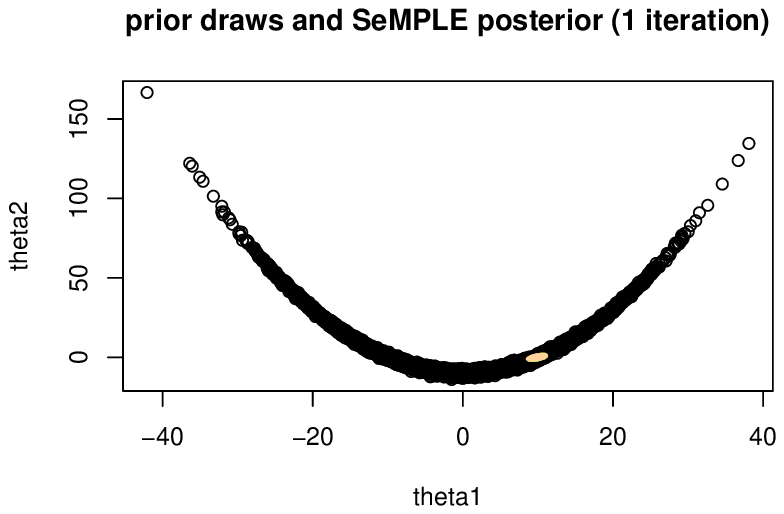}
    \caption{Draws from prior $p(\theta_1,\theta_2)$ (black) and from SeMPLE (brown) }\label{fig:twisted_prior_semple1}
\end{subfigure}
\caption{Twisted prior example: prior, true posterior and draws from SeMPLE at round $r=1.$}
\label{fig:twisted}
\end{figure}

\begin{figure}[h]
\centering
\begin{subfigure}[b]{0.49\textwidth}\includegraphics[width=0.9\textwidth]{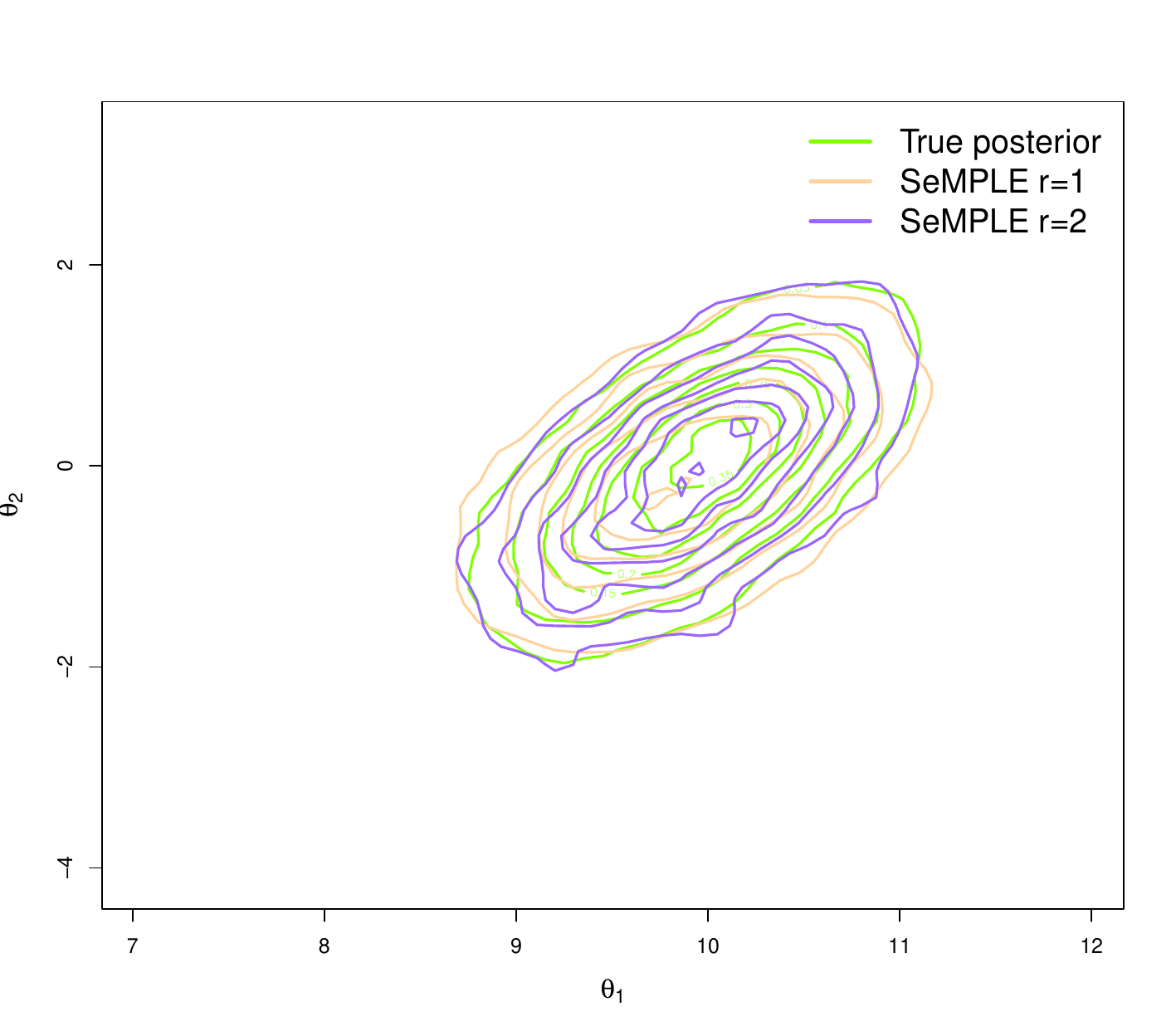}
\caption{Posteriors for $(\theta_1,\theta_2)$}
\end{subfigure}
    \begin{subfigure}[b]{0.49\textwidth}
\includegraphics[width=0.9\textwidth]{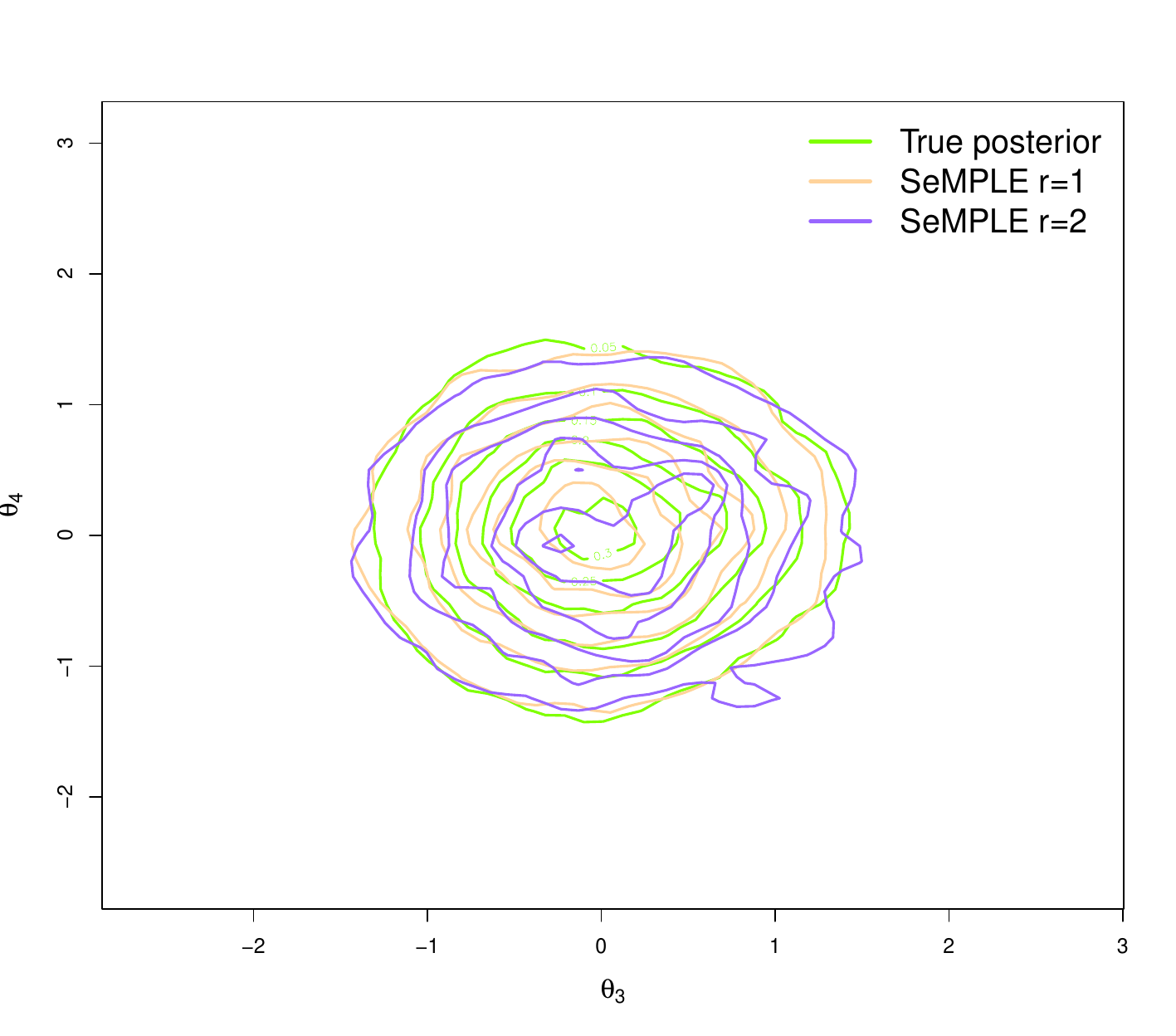}
    \caption{Posteriors for $(\theta_3,\theta_4)$}
\end{subfigure}
\caption{Twisted prior example: (a) Contour plots of the exact posterior for $(\theta_1,\theta_2)$ (green) and SeMPLE posterior at $r=1$ (brown) and $r=2$ (blue). (b) Same as (a) but for $(\theta_3,\theta_4)$.}
\label{fig:twisted_zoomed}
\end{figure}

We now show how a single round of SeMPLE ($r=1$), hence without any MCMC,  can in this case return excellent inference (notice, generally we recommend to run at least 2 rounds of SeMPLE, as MCMC is typically needed to ensure that samples from $r=1$ do not ``leak'' outside the prior's support). 
Figure \ref{fig:twisted} compares the banana-shaped prior for the first two components of $\btheta$  with draws  from the true posterior obtained via MCMC (panel (a)), and then with draws from a single \textcolor{black}{round} ($r=1$) of SeMPLE initialised with $K=10$ components (panel (b)). Here SeMPLE has been trained on 10,000 prior predictive simulations. Despite the true posterior occupies a tiny region within the prior's support, SeMPLE immediately identifies the bulk of the posterior in one \textcolor{black}{round}, that is Figure \ref{fig:twisted}(b) shows samples via the learned GLLiM posterior model. Then, in Figure \ref{fig:twisted_zoomed}(a) we zoom on the posterior region to better show the quality of the SeMPLE approximation for both $r=1$ and $r=2$: see Figure \ref{fig:twisted_zoomed}(a) for $(\theta_1,\theta_2)$ and Figure \ref{fig:twisted_zoomed}(b) for $(\theta_3,\theta_4)$, the latter being an illustrative example of the non-correlated components (inference behaves the same for all the remaining components, hence further marginals are not reported). It appears that, for this model, the correction brought by MCMC at $r=2$ is not necessary. 

\clearpage

\section{Biological model of the translation kinetics after mRNA transfection}\label{sec:mRNA}
This model serves as an example of a multidimensional SDE model with a higher number of model parameters, compared to the Ornstein-Uhlenbeck model. Moreover here the model is challenging as it is partially observed (only one coordinate is observed), and observations incorporate measurement measurement error. Reference posterior samples cannot be obtained exactly, as the transition densities are not known for this model.

\subsection{Model definition} \label{sec:mrna-model}
We consider the SDE model from \cite{pieschner2022identifiability} in a simulation study. In \cite{pieschner2022identifiability} the model was used to describe translation kinetics after mRNA transfection. Specifically, in a real data scenario,  time lapse microscopy images of the cells are taken for several hours, and for the first hour, the cells would be incubated with mRNA lipoplexes. Released mRNA gets translated into a green fluorescent protein (GFP) which causes the cells to fluoresce. The production of GFPs is the \textit{translation process}. The SDE below is used to study the translation kinetics of one cell, based on one observed fluorescence trajectory, where the two-dimensional stochastic process $(m(t),p(t))$ has $m(t)$ to represent the amount of mRNA molecules at time $t$, and $p(t)$ is the amount of GFP molecules at time $t$.
The SDE model is given by
\begin{equation}
    \label{bio_sde}
    d
    \begin{pmatrix}
        m \\
        p
    \end{pmatrix}
    (t)
    = 
    \begin{pmatrix}
        -\delta \cdot m(t) \\
        k \cdot m(t) - \gamma \cdot p(t)
    \end{pmatrix}
    dt
    + 
    \begin{pmatrix}
        \sqrt{\delta \cdot m(t)} & 0 \\
        0 & \sqrt{k \cdot m(t) + \gamma \cdot p(t)}
    \end{pmatrix}
    d B_t,
\end{equation}
where $dB_t$ is a two-dimensional standard Brownian motion. It is assumed that all mRNA molecules (within one cell) are released at once from the lipoplexes and denote this initial time point by $t_0$. Before $t_0$, there are neither mRNA nor GFP molecules, and at $t_0$, an amount $m_0$ of mRNA molecules is released, i.e
\[
\begin{pmatrix}
        m(t) \\
        p(t)
    \end{pmatrix}
    = 
    \begin{pmatrix}
        0 \\
        0
    \end{pmatrix}\qquad
    \text{for $t<t_0$}
\]
while $m(t_0)=m_0$ and $p(0)=0$.

We take as observable mapping 
\[
    {y}(t_i) = \log(\textrm{scale} \cdot p(t_i) + \textrm{offset})+\varepsilon(t_i) \qquad i=1,...,n,
\]
where $\varepsilon(t_i)$ is iid Gaussian measurement error  $\varepsilon(t_i) \sim \mathcal{N}(0, \sigma^2)$, and $t_1,...,t_n$ are observational time instants (see further below). Note that the observations depend only indirectly on process $\{m(t)\}$, and only $\{p(t)\}$ is observed. The parameter vector we wish to infer is $\btheta = (\delta, \gamma, k, m_0, \textrm{scale}, t_0, \textrm{offset}, \sigma)$.

It has been shown by \cite{pieschner2022identifiability} that SDE modelling improves identifiability of parameters, compared to using a corresponding ODE model. Nevertheless, some parameters are still unidentifiable, see in particular supplementary section A.4.1 in \cite{pieschner2022identifiability}. 

Solutions to the SDE \eqref{bio_sde} are simulated using an Euler-Maruyama scheme implemented in \texttt{Rcpp} \citep{rcpp} from $t=t_0$ to $t=30$, with step size 0.01.
The observed time series is interpolated from the Euler-Maruyama approximation at $n=60$ equidistant time points from $t_1=0.5$ to $t_n=30$. Note that this implies that observations can be made prior to $t=t_0$ where $p(t_i)=0$, $t_i<t_0$.

Inference is performed on the parameters log-scale, and the prior distributions are all Gaussian as follows
\begin{align*}
    \log \delta & \sim \mathcal{N}(-0.694, 0.5), \qquad
    \log \gamma \sim \mathcal{N}(-3, 0.5) \\
    \log k & \sim \mathcal{N}(0.027, 0.5),\qquad
    \log m_0  \sim \mathcal{N}(5.704, 0.5) \\
    \log \textrm{scale} & \sim \mathcal{N}(0.751, 0.5),\qquad
    \log t_0  \sim \mathcal{N}(-0.164, 0.5),\\
    \log \textrm{offset} & \sim \mathcal{N}(2.079, 0.5), \qquad
    \log \sigma \sim \mathcal{N}(-2, 0.5).
\end{align*}    

SeMPLE, SNL and SNPE-C were run for five different observed data sets, each generated with a different set of parameters given in Table \ref{tab:mrna-true-val}. For each method, the total simulation budget was set to 30k model simulations that were uniformly distributed across the number of algorithm rounds. For SeMPLE, the number of algorithm rounds was set to $R=3$, the number of starting mixture components was set to $K=10$, and the covariance matrices in GLLiM were set to be full unconstrained matrices. We did not inflate the covariance in the Metropolis-Hastings step ($\gamma = 1$). The mixture probability threshold to remove negligible mixture components was set to 0.005. For SNPE-C and SNL the number of algorithm rounds was set to 10.

\begin{table}[h]
\centering
\begin{tabular}{|l|l|l|l|l|l|l|l|l|}
\hline
$\,$ & $\log \delta$ & $\log \gamma$ & $\log k$ & $\log m_0$ & $\log \textrm{scale}$ & $\log t_0$ & $\log \textrm{offset}$ & $\log \sigma$ \\ \hline
Data set 1 & -0.694        & -3            & 0.027    & 5.704      & 0.751        & -0.164     & 2.079         & -2            \\ \hline
Data set 2 & -1.827        & -3.138        & -0.426   & 6.869      & 0.697        & 0.465      & 2.787         & -1.993        \\ \hline
Data set 3 & -1.153        & -3.501        & -0.181   & 4.601      & 0.386        & -2.121     & 2.185         & -2.578        \\ \hline
Data set 4 & -1.297        & -2.134        & 1.118    & 5.577      & 0.938        & 0.12       & 1.122         & -2.854        \\ \hline
Data set 5 & -0.243        & -4.596        & 0.168    & 5.573      & -0.086       & 1.234      & 1.392         & -1.655        \\ \hline
\end{tabular}
\caption{Translation kinetics model: parameter values used to generate observations for the five data sets.}
\label{tab:mrna-true-val}
\end{table}

\subsection{Results} \label{sec:mrna-results}
\begin{figure}[h]
    \centering
    \includegraphics[scale=0.7]{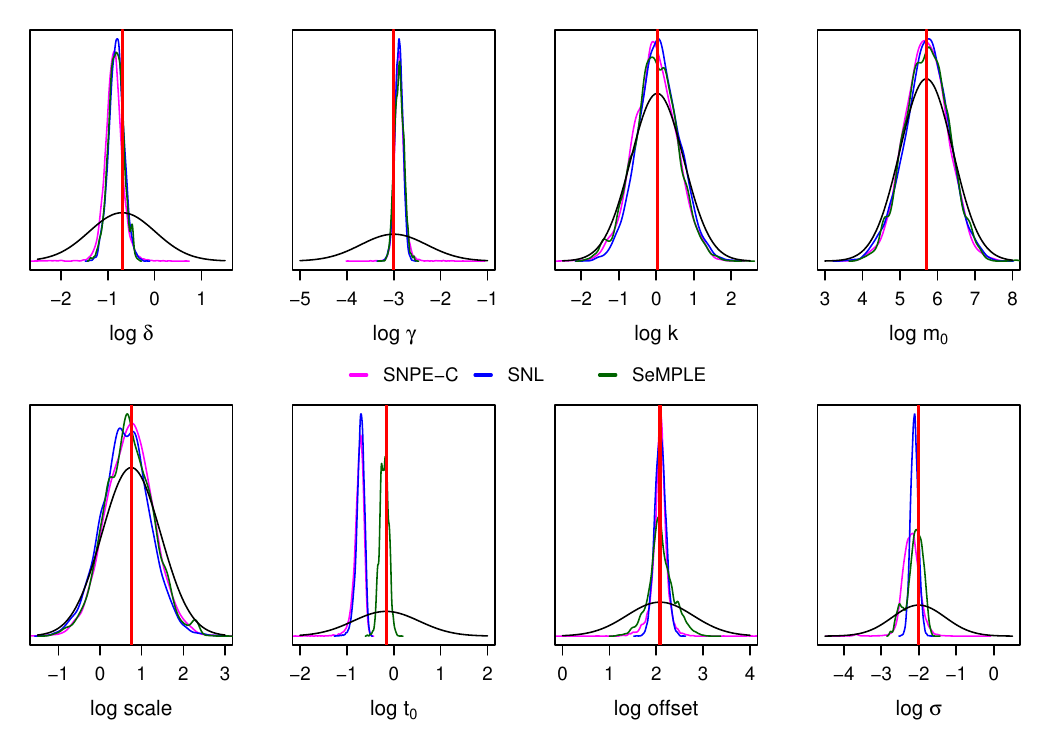}
    \caption{Translation kinetics model, observed data set \#1 out of 5: marginal posteriors from SeMPLE, SNPE-C and SNL (after 30k model simulations), priors (black lines) and ground truth parameters (vertical red lines). For SNPE-C and SNL we show results after 10 algorithm rounds, and after 3 rounds for SeMPLE.}
    \label{fig:mrna-all-obs1}
\end{figure}

\begin{figure}[h]
    \centering
    \includegraphics[scale=0.7]{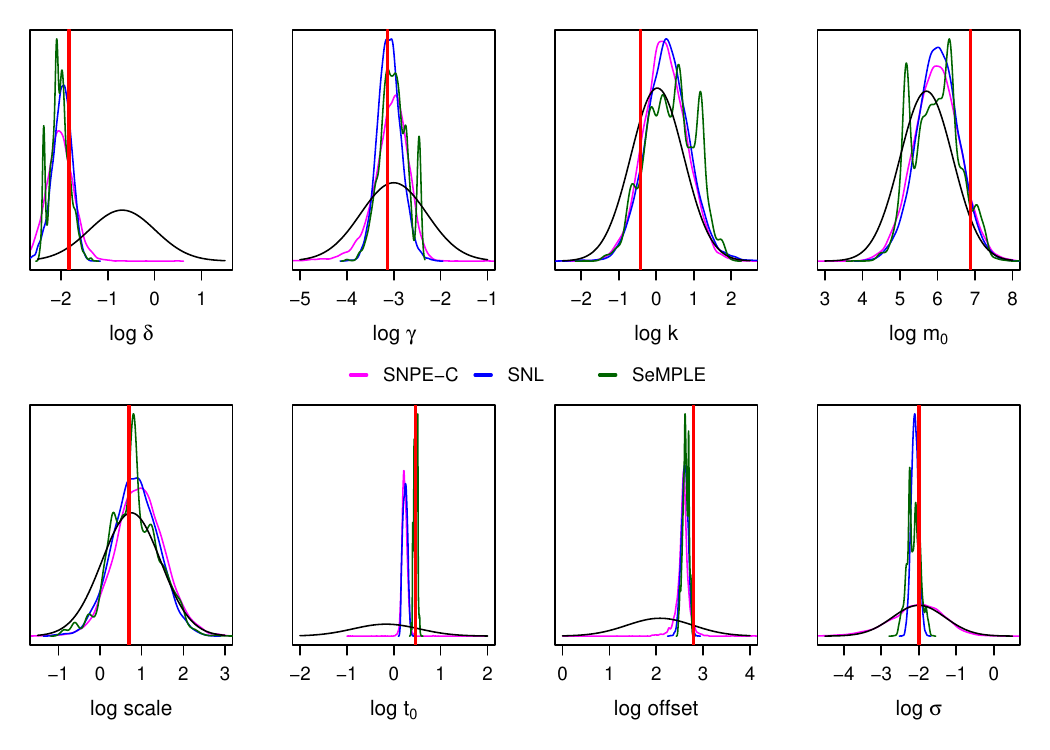}
    \caption{Translation kinetics model, observed data set \#2 out of 5: marginal posteriors from SeMPLE, SNPE-C and SNL (after 30k model simulations), priors (black lines) and ground truth parameters (vertical red lines). For SNPE-C and SNL we show results after 10 algorithm rounds, and after 3 rounds for SeMPLE.}
    \label{fig:mrna-all-obs2}
\end{figure}

\begin{figure}[h]
    \centering
    \includegraphics[scale=0.7]{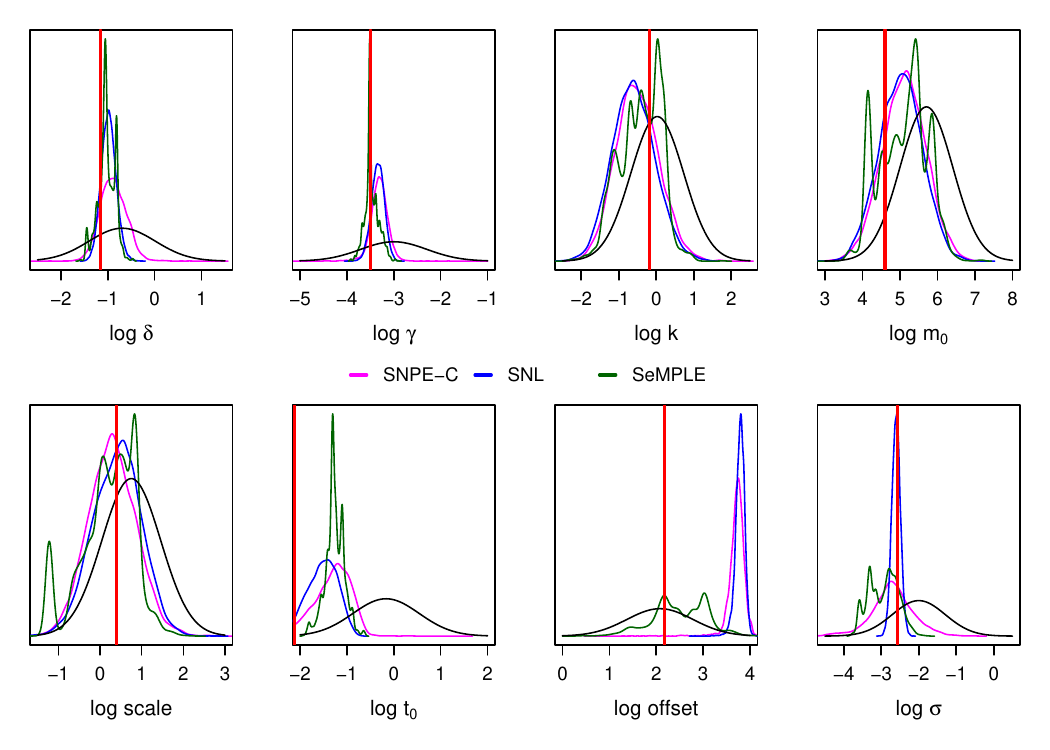}
    \caption{Translation kinetics model, observed data set \#3 out of 5: marginal posteriors from SeMPLE, SNPE-C and SNL (after 30k model simulations), priors (black lines) and ground truth parameters (vertical red lines). For SNPE-C and SNL we show results after 10 algorithm rounds, and after 3 rounds for SeMPLE.}
    \label{fig:mrna-all-obs3}
\end{figure}

\begin{figure}[h]
    \centering
    \includegraphics[scale=0.7]{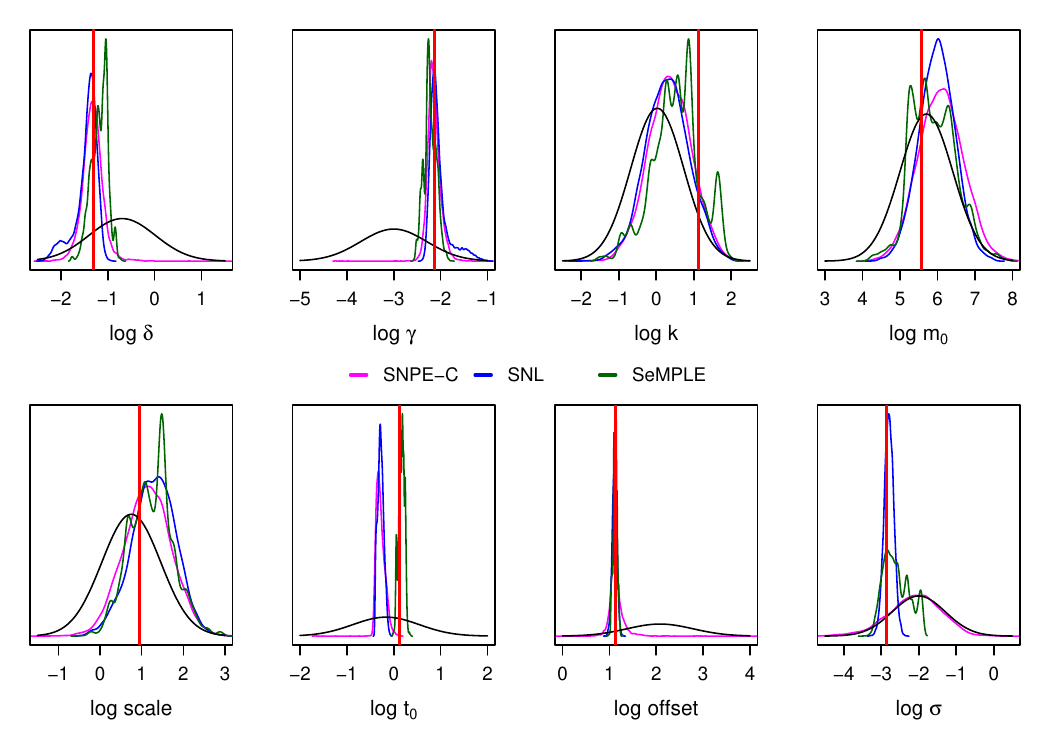}
    \caption{Translation kinetics model, observed data set \#4 out of 5: marginal posteriors from SeMPLE, SNPE-C and SNL (after 30k model simulations), priors (black lines) and ground truth parameters (vertical red lines). For SNPE-C and SNL we show results after 10 algorithm rounds, and after 3 rounds for SeMPLE.}
    \label{fig:mrna-all-obs4}
\end{figure}

\begin{figure}[h]
    \centering
    \includegraphics[scale=0.7]{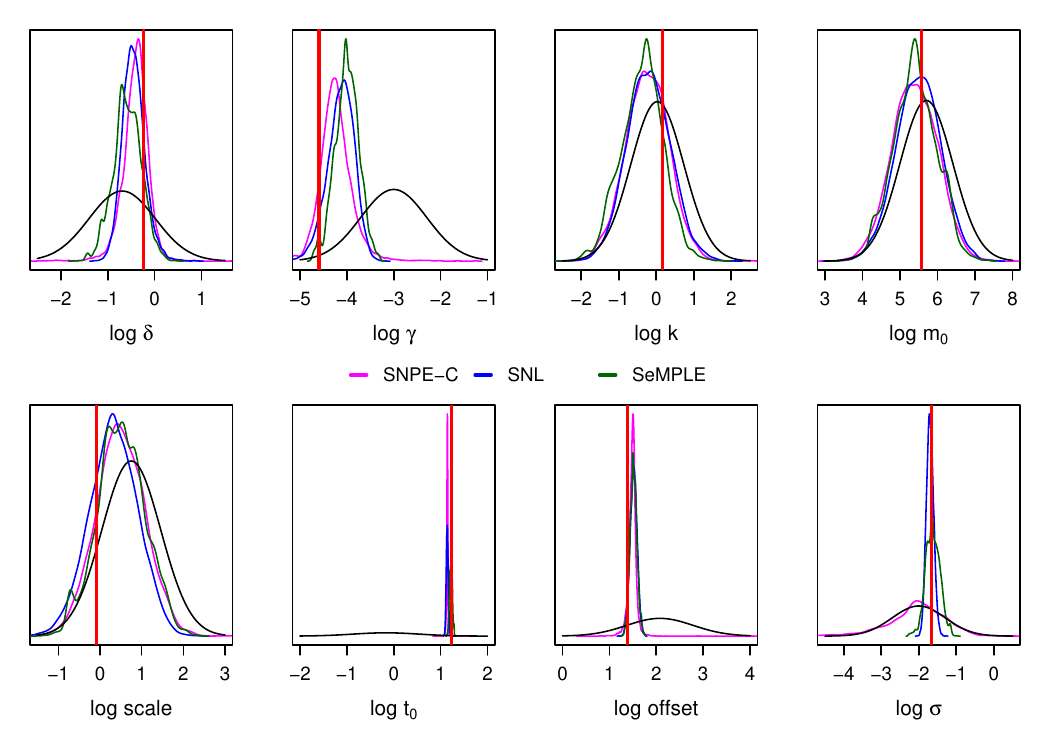}
    \caption{Translation kinetics model, observed data set \#5 out of 5: marginal posteriors from SeMPLE, SNPE-C and SNL (after 30k model simulations), priors (black lines) and ground truth parameters (vertical red lines). For SNPE-C and SNL we show results after 10 algorithm rounds, and after 3 rounds for SeMPLE.}
    \label{fig:mrna-all-obs5}
\end{figure}

\begin{figure}[h]
    \centering
    \includegraphics[scale=0.7]{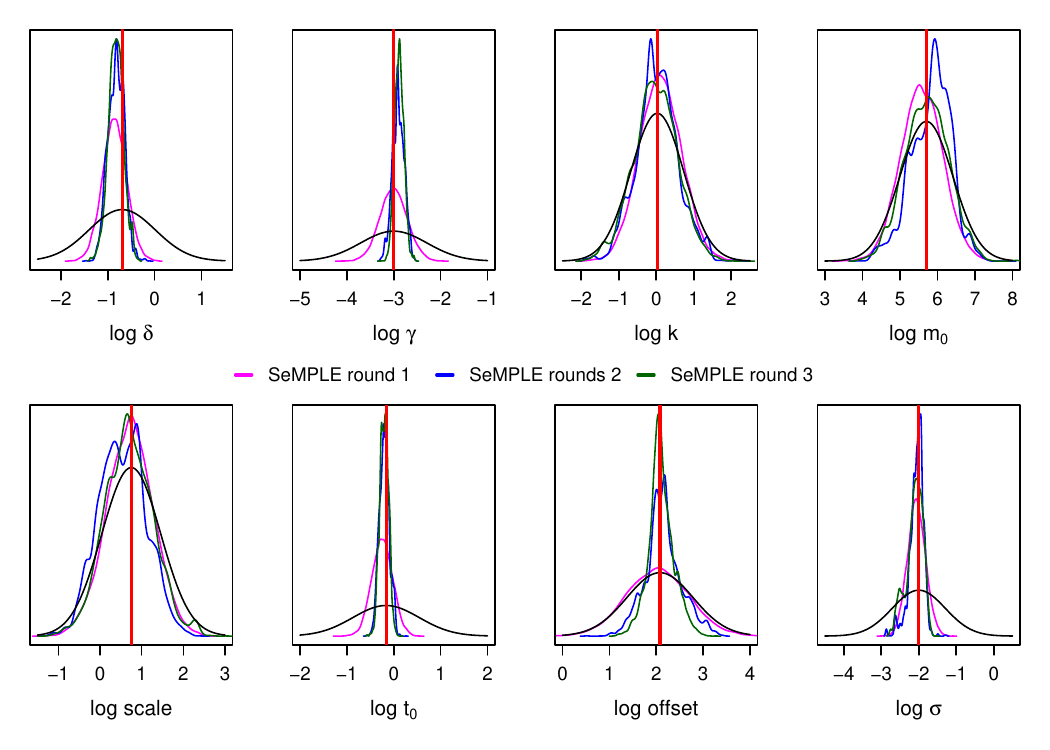}
    \caption{Translation kinetics model, observed data set \#1 out of 5: marginal posteriors from SeMPLE for every algorithm round (corresponding to 10k, 20k and 30k model simulations), priors (black lines) and ground truth parameters (vertical red lines).}
    \label{fig:mrna-semple-obs1}
\end{figure}

\begin{figure}[h]
    \centering
    \includegraphics[scale=0.7]{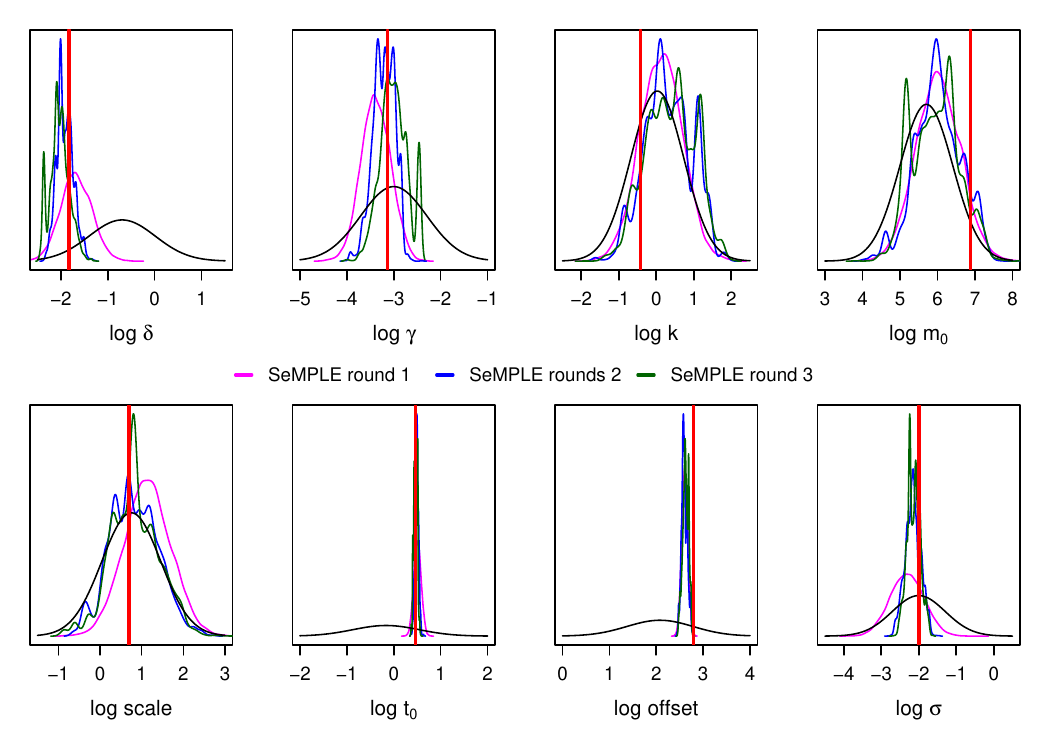}
    \caption{Translation kinetics model, observed data set \#2 out of 5: marginal posteriors from SeMPLE for every algorithm round (corresponding to 10k, 20k and 30k model simulations), priors (black lines) and ground truth parameters (vertical red lines). }
    \label{fig:mrna-semple-obs2}
\end{figure}

\begin{figure}[h]
    \centering
    \includegraphics[scale=0.7]{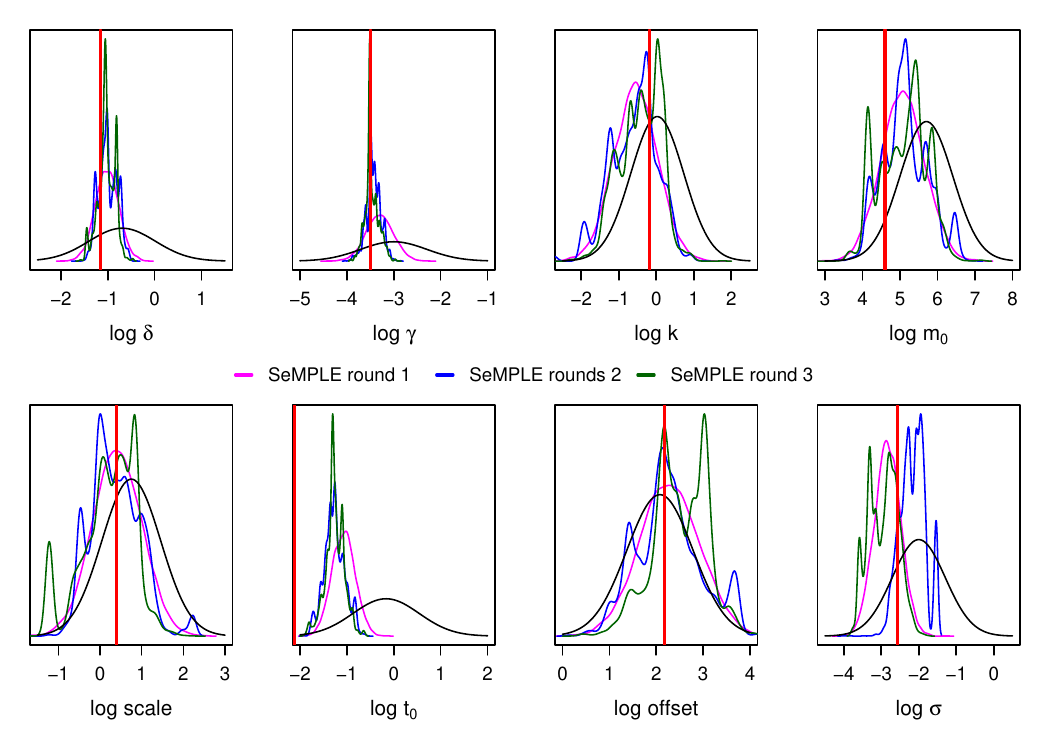}
    \caption{Translation kinetics model, observed data set \#3 out of 5: marginal posteriors from SeMPLE for every algorithm round (corresponding to 10k, 20k and 30k model simulations), priors (black lines) and ground truth parameters (vertical red lines).}
    \label{fig:mrna-semple-obs3}
\end{figure}

\begin{figure}[h]
    \centering
    \includegraphics[scale=0.7]{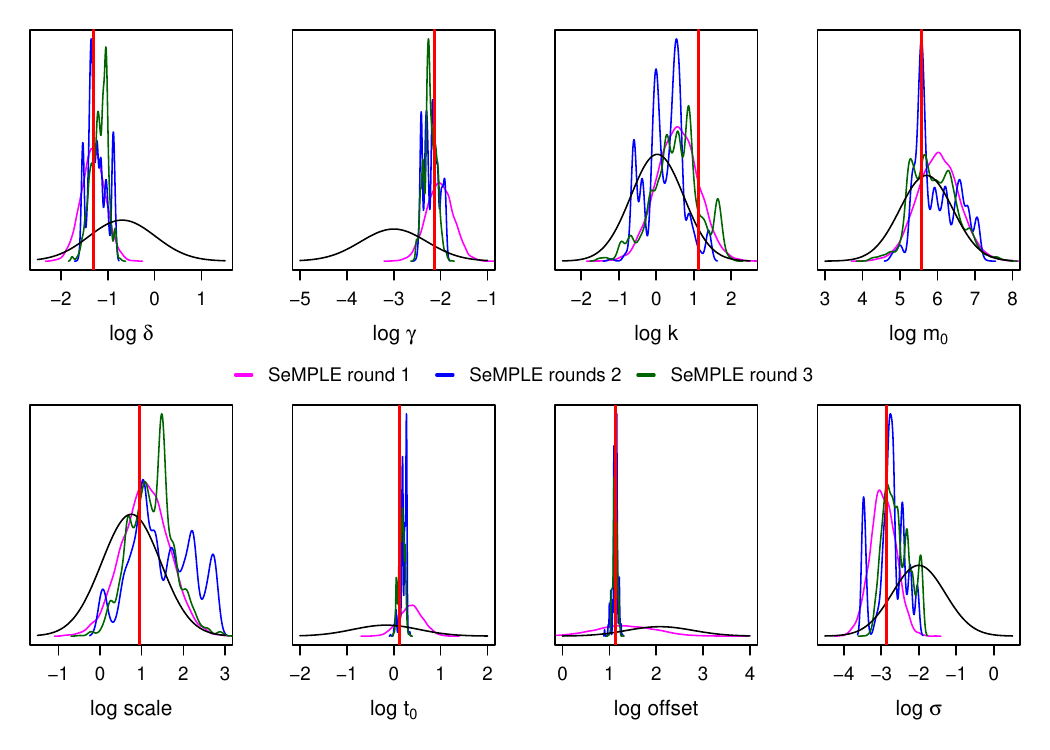}
    \caption{Translation kinetics model, observed data set \#4 out of 5: marginal posteriors from SeMPLE for every algorithm round (corresponding to 10k, 20k and 30k model simulations), priors (black lines) and ground truth parameters (vertical red lines).}
    \label{fig:mrna-semple-obs4}
\end{figure}

\begin{figure}[h]
    \centering
    \includegraphics[scale=0.7]{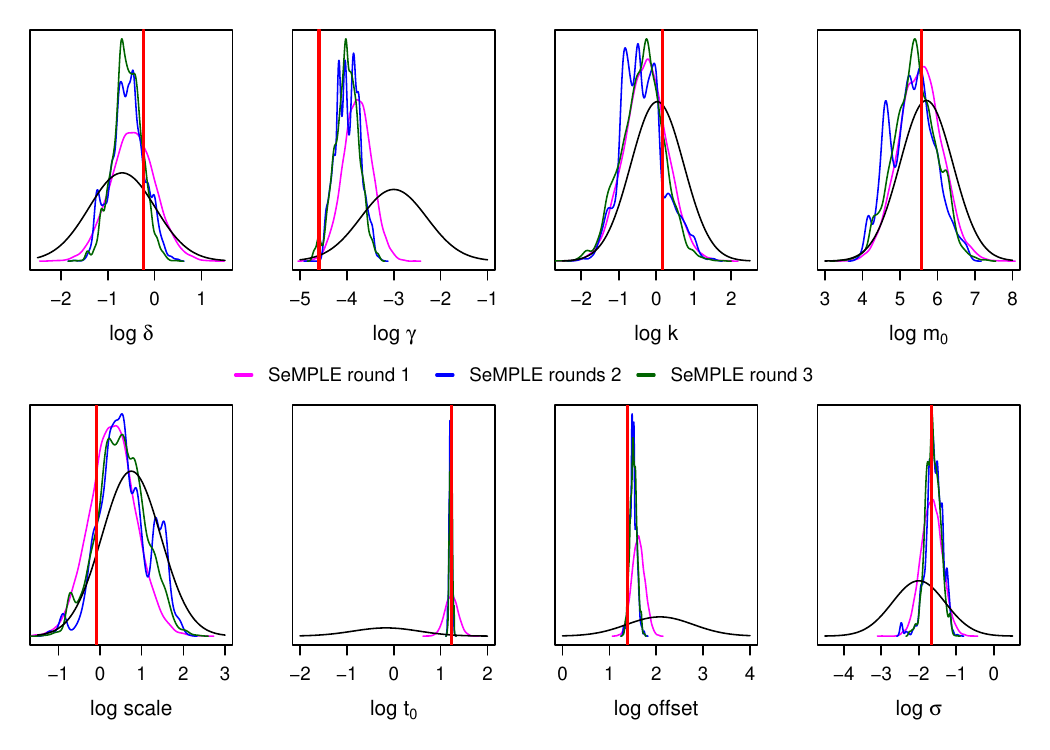}
    \caption{Translation kinetics model, observed data set \#5 out of 5: marginal posteriors from SeMPLE for every algorithm round (corresponding to 10k, 20k and 30k model simulations), priors (black lines) and ground truth parameters (vertical red lines).}
    \label{fig:mrna-semple-obs5}
\end{figure}

Figures \ref{fig:mrna-all-obs1}-\ref{fig:mrna-all-obs5} compare the marginal posteriors obtained by SeMPLE, SNPE-C and SNL, where each figure corresponds to a different dataset. All results are obtained after 30k model simulations but the number of algorithm rounds is 10 for both SNPE-C and SNL, while we ran 3 rounds for SeMPLE. Inference is very similar across the three methods, however it is clear that only SeMPLE can identify $t_0$ accurately, while SNL and SNPE-C provide a biased estimation of $t_0$ (an exception being the inference for dataset 3 in Figure \ref{fig:mrna-all-obs3}, where we deliberately challenged the inference by setting the ground truth value for $t_0$ far in the left tail of the prior). Moreover, SNPE-C is often unable to infer the measurement error variability $\sigma$ (datasets 2, 4, 5). As previously found in \cite{pieschner2022identifiability}, the lack of identifiability of parameters $(k,m_0,\textrm{scale})$ is confirmed in our analyses. However the other parameters are correctly inferred by SeMPLE. Finally, in Figures \ref{fig:mrna-semple-obs1}-\ref{fig:mrna-semple-obs5} we focus on displaying the performance of SeMPLE for each round.

\end{document}